\pdfoutput=1

\documentclass[11pt]{article}

\usepackage[final]{ACL2023}

\usepackage{times}
\usepackage{latexsym}
\usepackage{graphicx} %
\usepackage[T1]{fontenc}
\usepackage{booktabs}
\usepackage{multirow}
\usepackage{color, colortbl}
\usepackage{graphicx}
\usepackage{placeins}

\usepackage{tipa}
\usepackage[utf8]{inputenc}
\usepackage{caption}
\usepackage{subcaption}
\usepackage{booktabs}
\usepackage{bm}
\usepackage{hyperref}
\usepackage{amsmath}
\usepackage{cleveref}
\usepackage{tcolorbox}
\usepackage{amsmath}
\usepackage{amssymb}
\usepackage{xcolor} 
\usepackage{arydshln} 
\usepackage{xspace}
\usepackage{array}
\usepackage{lingmacros}
\usepackage{pifont}
\usepackage{colortbl}
\usepackage{listings}
\usepackage{enumitem} %

\newcommand{\cmark}{\ding{51}}%
\newcommand{\xmark}{\ding{55}}%
\newcommand\fnsep{\textsuperscript{,}}

\newcommand*{\afridoct}{\textsc{AfriDoc{-}MT}\xspace}
\newcommand*{\yoruba}{Yor\`ub\'a\xspace}
\newcommand*{\tech}{\textit{tech}\xspace}
\newcommand*{\health}{\textit{health}\xspace}

\usepackage[utf8]{inputenc}

\usepackage{microtype}

\usepackage{inconsolata}

\title{\afridoct: Document-level MT Corpus for African Languages}

\author{
Jesujoba O. Alabi$^{1,\dagger}$, Israel Abebe Azime$^{1,\dagger}$, Miaoran Zhang$^{1}$, Cristina España-Bonet$^{2,3}$, \\
 \textbf{Rachel Bawden$^{4}$,  Dawei Zhu$^{1}$,  David Ifeoluwa Adelani$^{5,\dagger}$, Clement Oyeleke Odoje$^{6}$,}\\ 
 \textbf{Idris Akinade$^{6,\dagger}$, Iffat Maab$^{7}$, Davis David$^{8,\dagger}$, Shamsuddeen Hassan Muhammad$^{9,\dagger}$,} \\
 \textbf{Neo Putini$^{10,\dagger}$, David O. Ademuyiwa$^{11}$, Andrew Caines$^{12}$, Dietrich Klakow$^{1}$} \\\\
 \footnotesize
$^{\dagger}$Masakhane NLP, 
$^1$Saarland University, Saarland Informatic Campus,
$^2$DFKI GmbH,
$^3$Barcelona Supercomputing Center,\\ 
 \footnotesize
$^4$Inria, Paris, France,
$^5$Mila, McGill University \& Canada CIFAR AI Chair,
$^6$University of Ibadan, Nigeria, \\
 \footnotesize
$^7$National Institute of Informatics, Japan,
$^8$Selcom Tanzania,
$^9$Imperial College London,
$^{10}$University of KwaZulu-Natal,\\ 
 \footnotesize
$^{11}$Loughborough University, U.K,
$^{12}$University of Cambridge, U.K.~~~\\
    \small{{\tt  jalabi@lsv.uni-saarland.de}}
}

\begin{document}
\maketitle
\begin{abstract}
This paper introduces \afridoct, a document-level multi-parallel translation dataset covering English and five African languages: Amharic, Hausa, Swahili, \yoruba, and Zulu. The dataset comprises 334 health and 271 information technology news documents, all human-translated from English to these languages. We conduct document-level translation benchmark experiments by evaluating the ability of neural machine translation (NMT) models and large language models (LLMs) to translate between English and these languages, at both the sentence and pseudo-document levels, the outputs being realigned to form complete documents for evaluation. Our results indicate that NLLB-200 achieves the best average performance among the standard NMT models, while GPT-4o outperforms general-purpose LLMs. Fine-tuning selected models leads to substantial performance gains, but models trained on sentences struggle to generalize effectively to longer documents. Furthermore, our analysis reveals that some LLMs exhibit issues such as under-generation, over-generation, repetition of words and phrases, and off-target translations, specifically for translation into African languages.\footnote{The data can be accessed on \href{https://huggingface.co/datasets/masakhane/AfriDocMT}{HuggingFace} and \href{https://github.com/masakhane-io/afridoc-mt}{GitHub}.}

\end{abstract}

\section{Introduction}
\label{sec:intro}

The field of machine translation (MT) has seen notable progress in the past years,  with neural machine translation (NMT) models achieving close to human performance in many high-resource language directions~\citep{NIPS2017_3f5ee243,akhbardeh-etal-2021-findings,mohammadshahi-etal-2022-small,yuan-etal-2023-lego,kocmi-etal-2023-findings,nllb2024scaling}. However, efforts have primarily been concentrated  on sentence-level translation,  without the use of inter-sentential context.  

In recent years, there has been interest in document-level translation (i.e.~the holistic translation of multiple sentences), where sentences are translated with their context rather than in isolation. Document-level translation is important in order to capture discourse relations~\citep{bawden-etal-2018-evaluating,voita-etal-2018-context,maruf2021survey}, maintain consistency and coherence across sentences~\citep{herold-ney-2023-improving}, particularly for technical domains, but poses unique challenges, such as how to handle longer documents~\citep{wang-etal-2024-benchmarking} given the limited context size of translation models. 
Current efforts have primarily focused on high-resource language directions, where document-level datasets are readily available~\citep{lopes-etal-2020-document,feng-etal-2022-learn,wu-etal-2023-document,wang-etal-2023-document-level,wu2024adaptinglargelanguagemodels}, and so far there has been no work on low-resource African languages.
Developing and evaluating document-level MT systems for low-resource languages is a useful and under-studied direction, which requires the creation of datasets.  

\begin{table*}[t]
\small\centering
  \begin{tabular}{lrcllll}
  \toprule
  \textbf{Dataset} & \textbf{\#Langs.} & \textbf{Multiway}  & \textbf{Domain} & \textbf{Type} & \textbf{\#Sents.} & \textbf{\#Docs.}  \\
  \midrule
  TICO-19~\citep{anastasopoulos-etal-2020-tico} & 12 & \cmark & health & document-level & 4k & 30 \\
  MAFAND-MT~\citep{adelani-etal-2022-thousand} & 16 & \xmark & news & sentence-level & 4k-35k & - \\
  FLORES-200~\citep{nllb2022}& $42$ & \cmark & general & sentence-level & 3k & - \\
  NTREX-128~\cite{federmann-etal-2022-ntrex} & 24 & \cmark & news & sentence-level & 1.9k & - \\
  \afridoct(Ours) & 5 & \cmark & tech, health & document-level & 10k & 271-334\\
  \bottomrule
  \end{tabular}
  \vspace{-2mm}
  \caption{Overview of highly related work, including for each dataset the number of African languages, the domain, the kind of MT task they can be used for and the range of the sentence numbers for each language direction. }
  \label{tab:works}
\end{table*}

To fill this gap, we present \afridoct, a  document-level translation dataset for English from and into five African languages: Amharic, Hausa, Swahili, \yoruba, and Zulu, created through the manual translation of English documents. It consists of 334 \textit{health} documents and 271 \textit{tech} documents. In addition, \afridoct supports multi-way translation, allowing translations not only between English and the African languages but also between any two of the languages covered. %

We conduct a comprehensive set of document translation benchmark experiments on \afridoct, using sentence-level and pseudo-document translation due to most models' limited context length, and then realigning them to form complete documents. We evaluate performance using automatic metrics and compare the results of encoder-decoder models with decoder-only LLMs across both domains. Our results demonstrate that NLLB-200, both before and after fine-tuning on \afridoct, excels in sentence translation, surpassing all other models. GPT-4o performs equally well for sentences and pseudo-documents, while other decoder-only models lag behind. In addition to automatic metrics, we use GPT-4o as a judge, human evaluation, and qualitative assessment to compare documents translation carried out sentence-by-sentence and as pseudo-documents for selected models. The evaluation shows that GPT-4o is generally unreliable for assessing document translations into African languages. However, we observe agreement between other evaluation methods, all indicating that sentence-by-sentence translation results in better document-level translation into African languages. We conduct additional analyses of the models' outputs to better understand their behavior and why they under-perform when translating pseudo-documents. They show that LLMs often under-generate, contain repetitions, and produce off-target translations, especially when translating into African languages.

\section{Related Work}
\label{sec:related}
\paragraph{MT Datasets for African Languages} 
Several MT datasets exist for African languages, including web-mined datasets such as WikiMatrix~\citep{schwenk-etal-2021-wikimatrix} and CCMatrix~\citep{schwenk-etal-2021-ccmatrix}. However, they have been adjudged to be of poor quality for certain low-resource subsets, including African languages~\citep{kreutzer-etal-2022-quality}. There are also well curated datasets for African languages including the Bible~\citep{mccarthy-etal-2020-johns}, JW300~\citep{agic-vulic-2019-jw300}\footnote{The dataset is no longer available for use.} and MAFAND-MT~\citep{adelani-etal-2022-thousand}, which are from religious and news domains. 

There exist several MT evaluation benchmark datasets for African languages. They can be categorized into two kinds. First, evaluation datasets specifically designed for translating into or from African languages~\citep[\textit{inter alia}]{ezeani-et-al-2020,azunre-et-al-2021,adelani-etal-2021-effect,adelani-etal-2022-thousand}. Second, benchmark datasets covering many languages, including African languages. For example, TICO-19~\citep{anastasopoulos-etal-2020-tico}, NTREX-128~\citep{federmann-etal-2022-ntrex}, FLORES-101~\citep{goyal-etal-2022-flores} and FLORES-200~\citep{nllb2024scaling} are a few such datasets. However, most of these datasets are designed for sentence-level MT, primarily drawn from religious or news domains, although some consist of translated sentences originating from the same document.
To the best of our knowledge, only TICO-19, a health domain translation benchmark, has the potential to be used for document-level MT, while it is restricted to topics related to COVID-19. \Cref{tab:works} gives a comparison of the most relevant existing benchmarks.

\paragraph{Document-level Neural Machine Translation} Document-level NMT aims to overcome the limitations of sentence-level systems by translating an entire document as a whole. Similar to context-aware NMT, which involves translating segments with additional, localized context, it differs in that it involves in principle translating an entire document holistically. Both document-level and context-aware MT allow for the possibility of improving translation quality for context-dependent phenomena such as coreference resolution~\citep{muller-etal-2018-large,bawden-etal-2018-evaluating,voita-etal-2018-context,herold-ney-2023-improving}, lexical disambiguation~\citep{rios-gonzales-etal-2017-improving, martinez-garcia-etal-2019-context}, and lexical cohesion~\citep{wong-kit-2012-extending,garcia2014document,garcia2017using,bawden-etal-2018-evaluating,voita-etal-2019-good}. Various methods have been proposed to extend sentence-level models to capture document-level context~\citep{tiedemann-scherrer-2017-neural,libovicky-helcl-2017-attention,bawden-etal-2018-evaluating,miculicich-etal-2018-document,sun-etal-2022-rethinking}. The emergence of LLMs, such as GPT-3~\citep{NEURIPS2020_1457c0d6}, Llama~\citep{dubey2024llama3herdmodels} and Gemma~\citep{gemmateam2024gemma2improvingopen}, has transformed NLP, including for MT~\citep{zhu-etal-2024-fine, zhu-etal-2024-multilingual, lu2024llamax}. Pre-trained on vast amounts of text, LLMs can effectively manage long-range dependencies, making them in principle well-suited for document-level translation. While these models have shown promising results for high-resource languages~\citep{wu-etal-2023-document,wang-etal-2023-document-level,wu2024adaptinglargelanguagemodels}, research remains limited for low-resource languages~\citep{ul-haq-etal-2020-document}.

\section{\afridoct Corpus}
\label{sec:domain}
\paragraph{Languages and their characteristics} We cover five languages from the two most common African language families: Afro-Asiatic and Niger-Congo. Three languages belong to the Niger-Congo family: Swahili (North-East Bantu), \yoruba (Volta-Niger) and isiZulu (Southern Bantu). The other two languages belong to the Afro-Asiatic family: Amharic (Semitic) and Hausa (Chadic). The choice of languages was based on geographical representation, speaking population, and web coverage (which we consider as a proxy for the potential performance of existing models on these languages).
Furthermore, each of these languages has over $20$ million speakers. All of them use the Latin script except for Amharic, which uses the Ge'ez script. The Latin-script languages use the Latin alphabet with the omission of some letters and the addition of new ones, and the use of diacritics (e.g.,~\yoruba). The languages are tonal, except for Amharic and Swahili. Just like English, all languages follow the subject-verb-object word order. Refer to \citet{Adelani_2022} for a comprehensive overview of the characteristics of these languages. \Cref{tab:languages} shows summarized details.  %

\begin{table}[t]
    \centering\small
    \scalebox{0.95}{ %
    \begin{tabular}{llr}
        \toprule
        \textbf{Language} & \textbf{Classification}  & \textbf{Spkrs.} (M) \\
        \midrule
    Amharic [amh] & Afro-Asiatic/Semitic  & $57.6$ \\
    Hausa [hau] & Afro-Asiatic/Chadic &   $78.5$ \\
    Swahili [swa] & Niger-Congo/Bantu &  $71.6$ \\
    \yoruba [yor] & Niger-Congo/Volta-Niger & $45.9$ \\
    isiZulu [zul] & Niger-Congo/Bantu & $27.8$ \\
        \bottomrule
    \end{tabular}
    }
    \caption{Languages in the \afridoct corpus, their classification and number of speakers (in millions).}
    \label{tab:languages}
\end{table}

\paragraph{Data Collection and Preprocessing} We scraped English articles from the websites of Techpoint Africa\footnote{\url{https://techpoint.africa/}} and the World Health Organization (WHO).\footnote{\url{https://www.who.int/health-topics}}\fnsep\footnote{\url{https://www.who.int/news-room/}} The articles cover different topics of different lengths with an average length of 30 and 37 sentences for \health and \tech respectively. While our corpus is initially structured at the article level, we aim to make it suitable for sentence-level translation tasks as well. To achieve this, we segmented the raw articles into sentences using NLTK~\citep{bird2009natural}. To ensure high segmentation quality, we recruited a linguist and a professional translator to verify the correctness of the segmentation and make corrections as needed. Finally, we selected 334 and 271 English articles/documents from the \health and \tech domains respectively, which represents 10k sentences each per domain.

\paragraph{Translation} We translated the extracted 10k English sentences to the 5 African languages through 4 expert translators per language.\footnote{Each translator was paid $\$1,250$ for $2,500$ sentences.} The translators were recruited through a language coordinator who is also a native speaker of the language. The 10k sentences were distributed equally among the translators and the translations were done in-context (i.e.~the translators translated on the sentence level but had access to the  whole document). Due to the domain-specific nature of the task, before starting the translation process, we conducted a translation workshop, during which three translation experts shared their experiences in creating terminologies and they also shared existing resources with the translators including short translation guidelines (\Cref{sec:app_translation_guide}).

\paragraph{Quality Checks}
Quality control was conducted using automated quality estimation, followed by manual inspections by our language coordinators. We also used Quality Estimation (QE), specifically AfriCOMET~\citep{wang2024afrimte},\footnote{\url{https://huggingface.co/masakhane/africomet-qe-stl}} Given a translated sentence in any African language and its corresponding source English sentence, AfriCOMET generates a score between 0 and 1, where 0 indicates poor quality and higher values signify better quality. The translators, in collaboration with the language coordinators, were tasked with reviewing instances that had quality estimation scores below 0.65. This step was essential to identify and correct low-quality translations.

\Cref{fig:distribution1} shows the distribution of final quality scores for five languages across both domains. Manual inspection indicates that QE scores below 0.65 do not necessarily reflect poor translations, consistent with \citet{adelani2024irokobenchnewbenchmarkafrican}, likely due to domain shift, translation length, and other factors, which warrant further investigation.

\paragraph{\afridoct data split} We created train, development (dev), and test splits for each domain. To prevent data leakage, documents sharing sentences with the same English translation were assigned to the training set. The dev set contains 25–33 documents, and the test set 59–61 documents, drawn from the remaining data.
\Cref{tab:app_lang_toks} shows the average number of whitespace-separated tokens per sentence across domains and languages, including English. The \health domain has more tokens than \tech. Hausa and \yoruba are longer than English, likely due to their descriptive nature, while Swahili is similar in length. Amharic and Zulu are relatively shorter, reflecting interesting linguistic properties. \Cref{tab:data_stats} provides additional data statistics.
\Cref{tab:data_stats} shows some data statistics. 

The \health data is licensed under CC BY-NC-SA 3.0, while the \tech data is licensed under CC BY-NC-SA 4.0. %

\begin{table}[t]
    \centering\small
    \scalebox{1}{ %
    \begin{tabular}{lcccc}
        \toprule
        \textbf{Domain} & \textbf{Train}  & \textbf{Dev.} & \textbf{Test}  & \textbf{Min/Max/Avg} \\
        \midrule
    
    \multicolumn{4}{l}{\textbf{Number of documents}} \\
    \health & $240$ & $33$ & $61$ & $2$/$151$/$29.9$ \\
    \tech & $187$ & $25$ & $59$ &  $8$/$247$/$36.9$ \\
    \midrule
    \multicolumn{4}{l}{\textbf{Number of sentences}} \\
    \health & $7041$ & $977$ & $1982$ & - \\
    \tech & $7048$ & $970$ & $1982$ &  - \\
    
        \bottomrule
    \end{tabular}
   }
    \caption{The number of documents and sentences in  \afridoct, and (at the document level) minimum, maximum and average sentences per document.}
    \label{tab:data_stats}
\end{table}

\section{Benchmark Experiments}
Given the \afridoct data, we conducted both sentence- and document-level translation, evaluating two types of models: encoder-decoder and decoder-only models. While the majority of these models are open-source, we also evaluated two proprietary models of the same type. Our evaluation primarily focuses on document-level translation, reflecting the availability of our document-level translation corpus. For completeness, we also conduct a series of sentence-level experiments, with the results presented in \Cref{sec:app_extra_result}.

\begin{table}[t]
    \centering\small
    \scalebox{0.85}{ %
    \begin{tabular}{lcccccc}
        \toprule
        \textbf{Domain} & \textbf{eng} & \textbf{amh}  & \textbf{hau} & \textbf{swa} & \textbf{yor} & \textbf{zul}\\
        \midrule
    
    \multicolumn{7}{l}{\textbf{Sentence}} \\
    \health & $21.6$ & $19.3$ & $28.1$ & $23.2$ & $27.9$ & $16.7$ \\
    \tech & $17.8$ & $15.6$ & $22.2$ & $18.0$ & $23.7$ & $13.4$ \\
    \midrule
    \multicolumn{7}{l}{\textbf{Document}} \\
    \health & $647.3$ & $576.7$ & $841.7$ & $695.4$ & $834.8$ & $500.1$ \\
    \tech & $658.2$ & $575.0$ & $821.6$ & $665.4$ & $873.4$ & $495.9$ \\

        \bottomrule
    \end{tabular}
   }
    \caption{The average number of tokens in \afridoct, both at sentence and document level.}
    \label{tab:app_lang_toks}
\end{table}
 
\subsection{Models}
\paragraph{Encoder-Decoder Models}
We evaluate five kinds of open encoder-decoder model including Toucan~\citep{elmadany-etal-2024-toucan, adebara-etal-2024-cheetah}, M2M-100~\citep{fan2020englishcentric}, NLLB-200~\citep{nllb2024scaling}, MADLAD-400~\citep{kudugunta2023madlad400}, and Aya-101~\citep{ustun2024aya}. Toucan is an Afro-centric multilingual MT model supporting 150 African language pairs. In comparison, M2M-100, NLLB-200, and MADLAD-400 cover between 100 and 450 language pairs. Aya-101, an instruction-tuned mT5 model~\citep{xue-etal-2021-mt5}, supports 100 languages and can translate between various languages, including those considered in \afridoct.

\paragraph{Decoder-only Models}
We also evaluate open and closed decoder-only models. Open models include LLama3.1~\citep{dubey2024llama3herdmodels}, Gemma2~\citep{gemmateam2024gemma2improvingopen}, their instruction-tuned variants, and LLaMAX3~\citep{lu2024llamax}—a LLama3-based model further pre-trained on 100+ languages, including several African ones. %
The closed models include OpenAI GPT models (GPT-3.5 Turbo and GPT-4o)~\citep{chatgpt}, which have been shown to have document-level translation ability~\citep{wang-etal-2023-document-level}.  While their language coverage is not well documented, they show some ability to handle African languages~\citep{adelani2024irokobenchnewbenchmarkafrican,bayes2024uhurabenchmarkevaluatingscientific}, though far below their performance in English, their primary training language.

We present the result of $12$ models in total, including the 1.2B version of Toucan, 1.3B and 3.3B versions of NLLB-200, 3B and 13B versions of MADLAD-400 and Aya-101 respectively. We also have the 8B instruction tuned version of LLama3.1 (LLama3.1-IT), 9B version of Gemma-2 (Gemma2-IT), and LLaMAX3-Alpaca.\footnote{We refer to it as LLaMAX3-Alp in the results tables.} We provide more description of the models in \Cref{sec:app_eval_methods}.

\paragraph{Supervised fine-tuning of the models}
For sentence-level evaluation, we jointly fine-tune NLLB-200 with 1.3B parameters on the 30 language directions and on the two domains to make the models more specialized. Similarly, we did supervised fine-tuning (SFT) on LLaMAX3 and LLama3.1 using the prompt augmentation approach from \citep{zhu-etal-2024-preference}, as shown in \Cref{sec:app_eval_prompts}. We chose these two models because LLaMAX3 is already adapted to several languages including our languages of interest, and LLama3.1 because of its long context window.  We perform SFT on LLaMAX3 and LLama3.1 for document-level translation, using pseudo-documents with $k$=10. We refer to each system as \{model\_name\}-SFT$_k$.\footnote{We denote models finetuned on sentences as \{model\_name\}-SFT or \{model\_name\}-SFT$_1$ }

\subsection{Experimental Setup} 

\paragraph{Sentence-level Evaluation} Given that our created dataset can be used for sentence-level translation and as a baseline for document-level translation, we evaluate all models on the test splits for each domain. 
We evaluate the translation models (M2M-100, NLLB-200, and MADLAD-400) using the Fairseq~\citep{ott2019fairseq} codebase for (M2M-100 and NLLB-200), and the Transformers~\citep{wolf-etal-2020-transformers} codebase for MADLAD-400. However, for other models including Aya-101, we use the EleutherAI LM Evaluation Harness (\texttt{lm-eval}) tool~\citep{biderman2024lessons} using the three templates listed in \Cref{tab:prompt_examples} of~\Cref{sec:app_eval_prompts}.

\begin{table*}[t]
 \footnotesize
 \begin{center}
 \resizebox{\textwidth}{!}{%
  \begin{tabular}{p{2.1cm}l|cccccc|cccccc||c}
    \toprule
    \rowcolor{gray!30}
    \textbf{Model} & \textbf{Size} & \multicolumn{6}{c}{\textit{eng {$\rightarrow$} X}} & \multicolumn{6}{c}{\textit{X {$\rightarrow$} eng}} & \textbf{AVG} \\
     \rowcolor{gray!30}
     &  & \textbf{amh} & \textbf{hau} & \textbf{swa} & \textbf{yor} & \textbf{zul} & \textbf{Avg.} & \textbf{amh} & \textbf{hau} & \textbf{swa} & \textbf{yor} & \textbf{zul} & \textbf{Avg.} &   \\
\midrule
\multicolumn{15}{l}{\textbf{Encoder-Decoder}}\\
Toucan & 1.2B & $33.8_{1.2}$ & $57.6_{1.4}$ & $70.3_{0.8}$ & $36.0_{1.5}$ & $58.0_{1.0}$ & $51.2$ & $54.7_{1.0}$ & $57.7_{1.3}$ & $65.2_{0.9}$ & $54.0_{1.2}$ & $59.9_{0.8}$ & $58.3$ & 54.7 \\
NLLB-200 & 1.3B & $49.8_{1.5}$ & $64.7_{2.2}$ & $75.5_{0.8}$ & $45.1_{1.0}$ & $69.0_{1.3}$ & $60.8$ & $69.4_{1.3}$ & $65.3_{1.7}$ & $75.3_{0.8}$ & $66.3_{1.1}$ & $73.2_{0.9}$ & $69.9$ & 65.4 \\
MADLAD-400 & 3B & $36.5_{0.9}$ & $54.4_{2.0}$ & $74.2_{0.9}$ & $19.1_{0.9}$ & $57.1_{1.4}$ & $48.3$ & $68.9_{1.1}$ & $63.8_{1.6}$ & $76.1_{0.6}$ & $51.4_{1.8}$ & $68.9_{0.9}$ & $65.8$ & 57.0 \\
NLLB-200 & 3.3B & $53.0_{1.9}$ & $65.2_{2.2}$ & $76.7_{0.7}$ & $43.8_{1.1}$ & $70.7_{1.3}$ & $61.9$ & $70.9_{1.3}$ & $66.5_{1.7}$ & $77.0_{0.7}$ & $67.6_{1.1}$ & $74.7_{1.0}$ & $71.3$ & 66.6 \\
Aya-101 & 13B & $36.6_{0.9}$ & $56.4_{1.5}$ & $44.7_{2.4}$ & $31.2_{1.4}$ & $58.6_{0.8}$ & $45.5$ & $64.6_{1.1}$ & $61.5_{1.4}$ & $70.8_{0.8}$ & $57.9_{1.3}$ & $67.4_{0.8}$ & $64.4$ & 55.0 \\
\multicolumn{15}{l}{\textbf{SFT on \afridoct}}\\
\rowcolor{blue!20}
NLLB-SFT & 1.3B & $\textbf{55.9}_{1.6}$ & $\textbf{67.4}_{1.9}$ & $\textbf{81.3}_{0.7}$ & $\textbf{61.5}_{1.0}$ & $\textbf{73.7}_{1.6}$ & $\textbf{68.0}$ & $\textbf{72.4}_{1.2}$ & $\textbf{67.5}_{1.6}$ & $\textbf{79.2}_{0.7}$ & $\textbf{71.8}_{1.1}$ & $\textbf{76.5}_{0.9}$ & $\textbf{73.5}$ & \textbf{70.7} \\
\midrule
\multicolumn{15}{l}{\textbf{Decoder-only}}\\
Gemma2-IT & 9B & $20.1_{0.7}$ & $56.4_{1.4}$ & $71.2_{0.7}$ & $21.0_{0.6}$ & $41.6_{1.1}$ & $42.1$ & $61.6_{0.9}$ & $62.5_{1.3}$ & $74.2_{0.7}$ & $54.7_{1.3}$ & $63.9_{0.9}$ & $63.4$ & 52.7 \\
LLama3.1-IT & 8B & $19.6_{0.5}$ & $45.9_{1.4}$ & $63.7_{0.9}$ & $19.7_{0.6}$ & $28.5_{0.7}$ & $35.5$ & $53.9_{0.9}$ & $59.8_{1.3}$ & $69.1_{0.9}$ & $53.4_{1.3}$ & $54.0_{1.1}$ & $58.0$ & 46.8 \\
LLaMAX3-Alp & 8B & $30.5_{0.8}$ & $56.3_{1.5}$ & $67.8_{0.8}$ & $19.3_{0.8}$ & $56.1_{0.9}$ & $46.0$ & $63.3_{1.0}$ & $62.4_{1.3}$ & $71.7_{0.8}$ & $56.1_{1.1}$ & $65.3_{0.9}$ & $63.8$ & 54.9 \\
GPT-3.5 & -- & $20.4_{0.6}$ & $44.3_{0.9}$ & $76.7_{0.6}$ & $21.3_{0.9}$ & $51.1_{0.9}$ & $42.8$ & $48.3_{0.9}$ & $52.4_{1.2}$ & $75.0_{0.6}$ & $52.1_{1.2}$ & $59.5_{0.9}$ & $57.4$ & 50.1 \\
GPT-4o & -- & $36.7_{0.8}$ & $64.2_{1.9}$ & $79.8_{0.6}$ & $29.3_{1.6}$ & $69.0_{1.3}$ & $55.8$ & $67.2_{1.0}$ & $66.5_{1.5}$ & $78.1_{0.6}$ & $69.1_{1.1}$ & $75.1_{1.0}$ & $71.2$ & 63.5 \\
\multicolumn{15}{l}{\textbf{SFT on \afridoct}}\\
\rowcolor{blue!20}
LLaMAX3-SFT & 8B & $46.8_{1.2}$ & $62.5_{1.4}$ & $73.1_{0.9}$ & $57.5_{1.0}$ & $67.5_{1.0}$ & $61.5$ & $66.6_{1.2}$ & $58.9_{1.6}$ & $73.1_{1.1}$ & $64.7_{1.5}$ & $70.5_{1.0}$ & $66.8$ & 64.1 \\
\rowcolor{blue!20}
LLama3.1-SFT & 8B & $45.6_{1.1}$ & $61.8_{1.5}$ & $71.5_{1.0}$ & $57.0_{1.1}$ & $66.8_{0.9}$ & $60.6$ & $64.3_{1.2}$ & $59.5_{1.5}$ & $72.1_{0.8}$ & $64.8_{1.5}$ & $69.0_{1.0}$ & $65.9$ & 63.2 \\
\bottomrule
    
  \end{tabular}
  }
  \vspace{-3mm}
  \caption{Performance of the models in the Health domain, measured by d-chrF at the sentence-level, realigned to the document-level. For each model and language, the best result from three prompt variations is reported.}
  \vspace{-4mm}
  \label{tab:main_result2}
  \end{center}
\end{table*}

\begin{table*}[t]
 \footnotesize
 \begin{center}
 \resizebox{\textwidth}{!}{%
  \begin{tabular}{p{2.1cm}l|cccccc|cccccc||c}
    \toprule
    \rowcolor{gray!30}
    \textbf{Model} & \textbf{Size} & \multicolumn{6}{c}{\textit{eng {$\rightarrow$} X}} & \multicolumn{6}{c}{\textit{X {$\rightarrow$} eng}} & \textbf{AVG} \\
     \rowcolor{gray!30}
     &  & \textbf{amh} & \textbf{hau} & \textbf{swa} & \textbf{yor} & \textbf{zul} & \textbf{Avg.} & \textbf{amh} & \textbf{hau} & \textbf{swa} & \textbf{yor} & \textbf{zul} & \textbf{Avg.} &   \\
\midrule
\multicolumn{15}{l}{\textbf{Encoder-Decoder}}\\
Toucan & 1.2B & $32.0_{1.6}$ & $59.5_{1.7}$ & $66.1_{1.7}$ & $37.1_{2.0}$ & $58.5_{1.4}$ & $50.7$ & $54.0_{1.6}$ & $59.9_{1.5}$ & $64.1_{1.4}$ & $54.3_{1.3}$ & $59.6_{1.2}$ & $58.4$ & 54.5 \\
NLLB-200 & 1.3B & $49.3_{2.0}$ & $65.7_{2.2}$ & $72.3_{1.6}$ & $43.0_{1.3}$ & $70.3_{1.3}$ & $60.1$ & $69.5_{1.0}$ & $66.8_{1.5}$ & $72.0_{1.4}$ & $63.0_{1.2}$ & $71.5_{1.2}$ & $68.5$ & 64.3 \\
MADLAD-400 & 3B & $37.3_{1.3}$ & $57.0_{2.8}$ & $62.1_{2.9}$ & $21.3_{1.0}$ & $58.5_{1.8}$ & $47.3$ & $68.6_{1.1}$ & $66.0_{1.4}$ & $72.1_{1.4}$ & $53.1_{1.4}$ & $67.6_{1.2}$ & $65.5$ & 56.4 \\
NLLB-200 & 3.3B & $52.2_{2.4}$ & $65.4_{2.3}$ & $72.8_{1.5}$ & $40.1_{1.8}$ & $71.6_{1.3}$ & $60.4$ & $70.9_{1.0}$ & $67.7_{1.5}$ & $73.2_{1.4}$ & $63.9_{1.1}$ & $72.5_{1.2}$ & $69.6$ & 65.0 \\
Aya-101 & 13B & $37.3_{1.1}$ & $58.9_{2.3}$ & $42.4_{2.6}$ & $31.4_{1.4}$ & $58.9_{1.5}$ & $45.8$ & $65.2_{1.2}$ & $64.8_{1.2}$ & $69.1_{1.1}$ & $58.5_{1.3}$ & $67.1_{1.1}$ & $64.9$ & 55.4 \\
\multicolumn{15}{l}{\textbf{SFT on \afridoct}}\\
\rowcolor{blue!20}
NLLB-SFT & 1.3B & $\textbf{53.4}_{2.4}$ & $\textbf{67.9}_{2.2}$ & $\textbf{76.5}_{1.6}$ & $\textbf{59.5}_{1.3}$ & $\textbf{74.0}_{1.5}$ & $\textbf{66.2}$ & $\textbf{72.1}_{1.0}$ & $69.0_{1.3}$ & $74.1_{1.4}$ & $\textbf{67.5}_{1.1}$ & $\textbf{74.3}_{1.1}$ & $\textbf{71.4}$ & \textbf{68.8} \\
\midrule
\multicolumn{15}{l}{\textbf{Decoder-only}}\\
Gemma2-IT & 9B & $20.6_{0.6}$ & $58.3_{1.5}$ & $68.7_{1.6}$ & $23.9_{1.3}$ & $46.5_{1.8}$ & $43.6$ & $61.1_{1.3}$ & $65.4_{1.4}$ & $71.5_{1.2}$ & $56.7_{1.3}$ & $63.8_{1.1}$ & $63.7$ & 53.7 \\
LLama3.1-IT & 8B & $19.5_{0.9}$ & $47.8_{1.3}$ & $63.4_{1.5}$ & $20.8_{1.2}$ & $30.4_{1.3}$ & $36.4$ & $51.0_{1.3}$ & $61.0_{1.4}$ & $66.0_{1.3}$ & $53.5_{1.2}$ & $52.4_{1.3}$ & $56.8$ & 46.6 \\
LLaMAX3-Alp & 8B & $30.3_{1.1}$ & $58.9_{1.9}$ & $64.9_{1.7}$ & $22.0_{0.8}$ & $58.6_{1.7}$ & $46.9$ & $63.4_{1.4}$ & $64.9_{1.5}$ & $69.1_{1.1}$ & $56.5_{1.3}$ & $65.7_{1.2}$ & $63.9$ & 55.4 \\
GPT-3.5 & -- & $22.6_{0.8}$ & $49.2_{1.5}$ & $72.6_{1.6}$ & $23.0_{1.0}$ & $53.6_{1.5}$ & $44.2$ & $47.4_{1.5}$ & $56.5_{1.3}$ & $71.5_{1.4}$ & $54.0_{1.3}$ & $59.9_{1.1}$ & $57.9$ & 51.0 \\
GPT-4o & -- & $36.9_{1.2}$ & $65.2_{2.3}$ & $75.3_{1.6}$ & $29.4_{1.5}$ & $71.1_{1.4}$ & $55.6$ & $67.2_{1.0}$ & $\textbf{69.1}_{1.4}$ & $\textbf{74.4}_{1.4}$ & $66.4_{1.1}$ & $73.4_{1.1}$ & $70.1$ & 62.8 \\
\multicolumn{15}{l}{\textbf{SFT on \afridoct}}\\
\rowcolor{blue!20}
LLaMAX3-SFT & 8B & $42.8_{1.5}$ & $62.4_{1.9}$ & $67.6_{1.4}$ & $55.2_{1.5}$ & $66.0_{1.2}$ & $58.8$ & $63.0_{1.2}$ & $53.5_{1.9}$ & $67.5_{1.2}$ & $57.3_{1.3}$ & $66.8_{1.3}$ & $61.6$ & 60.2 \\
\rowcolor{blue!20}
LLama3.1-SFT & 8B & $41.6_{1.7}$ & $61.8_{2.0}$ & $66.4_{1.3}$ & $54.9_{1.4}$ & $64.6_{1.6}$ & $57.9$ & $62.0_{1.2}$ & $58.6_{1.5}$ & $67.1_{1.2}$ & $61.3_{1.3}$ & $65.6_{1.3}$ & $62.9$ & 60.4 \\
\bottomrule
    
  \end{tabular}
  }
  \vspace{-3mm}
  \caption{Performance of the models in the Tech domain, measured by d-chrF at the sentence-level, realigned to the document-level. For each model and language, the best result from three prompt variations is reported.}
  \label{tab:main_result1}
  \end{center}
\end{table*}

\paragraph{Document-level Evaluation} We also perform document-level translation using a setup similar to the sentence-level experiment, but only with models that meet context length requirements.  An initial analysis showed that some models were unable to process entire documents due to input length limits, which were exceeded by token counts in some languages (Amharic and \yoruba). To address this, we adopted a similar approach to~\citet{lee-etal-2022-docmt5}, splitting documents into fixed-size chunks of $k$ sentences to fit within token limits; the final chunk may contain fewer than $k$ sentences. To select an appropriate chunk size, we conducted initial tests with $k = 1$ (sentence-level), 5, 10, and 25, choosing $k=10$ for our experiments. We provide results from this analysis in~\Cref{tab:app_doc_stats}.

\subsection{Evaluation Metrics} 
Evaluating document-level translation remains challenging, as existing automatic metrics struggle to capture improvements and account for discourse phenomena~\citep{jiang-etal-2022-blonde,dahan2024survey}, and embedding-based metrics have not been explored in this context for African languages due to the lack of data. Hence, we realigned sentence-level or pseudo-translation outputs into full documents, then computed BLEU~\citep{papineni-etal-2002-bleu} and chrF~\citep{popovic-2015-CHRF} to create document BLEU (d-BLEU) and document chrF (d-chrF). Metrics were computed using SacreBLEU\footnote{\texttt{case:mixed|eff:no|tok:13a|smooth:exp|v:2.3.1}}~\citep{post-2018-call} with bootstrap resampling ($n=1000$) to report 95\% confidence intervals. We report d-chrF scores for the best prompt per model and language direction in the main text, as chrF better captures the morphological richness of African languages~\citep{adelani-etal-2022-thousand}, with full results provided in~\Cref{sec:app_extra_result}.

\begin{table*}[t]
 \footnotesize
 \begin{center}
 \resizebox{\textwidth}{!}{%
  \begin{tabular}{p{2.4cm}l|cccccc|cccccc||c}
    \toprule
    \rowcolor{gray!30}
    \textbf{Model} & \textbf{Size} & \multicolumn{6}{c}{\textit{eng {$\rightarrow$} X}} & \multicolumn{6}{c}{\textit{X {$\rightarrow$} eng}} & \textbf{AVG} \\
     \rowcolor{gray!30}
     &  & \textbf{amh} & \textbf{hau} & \textbf{swa} & \textbf{yor} & \textbf{zul} & \textbf{Avg.} & \textbf{amh} & \textbf{hau} & \textbf{swa} & \textbf{yor} & \textbf{zul} & \textbf{Avg.} &   \\
\midrule
\multicolumn{15}{l}{\textbf{Encoder-Decoder}}\\
MADLAD-400 & 3B & $27.5_{1.8}$ & $40.2_{2.3}$ & $46.6_{3.4}$ & $15.1_{0.8}$ & $43.6_{2.6}$ & $34.6$ & $63.3_{1.6}$ & $62.5_{2.0}$ & $74.4_{0.9}$ & $44.2_{1.6}$ & $66.6_{1.5}$ & $62.2$ & 48.4 \\
Aya-101 & 13B & $28.7_{1.6}$ & $48.5_{2.3}$ & $34.7_{3.4}$ & $18.7_{1.3}$ & $54.9_{1.4}$ & $37.1$ & $61.6_{1.7}$ & $62.3_{1.8}$ & $71.2_{0.9}$ & $56.1_{2.1}$ & $69.0_{1.0}$ & $64.0$ & 50.6 \\
\midrule
\multicolumn{15}{l}{\textbf{Decoder-only}}\\
Gemma2-IT & 9B & $6.5_{0.6}$ & $37.0_{3.4}$ & $52.9_{3.6}$ & $6.4_{0.5}$ & $12.0_{1.0}$ & $23.0$ & $36.5_{3.0}$ & $51.8_{3.4}$ & $65.0_{3.0}$ & $44.8_{2.9}$ & $56.1_{3.3}$ & $50.8$ & 36.9 \\
LLama3.1-IT & 8B & $7.5_{0.5}$ & $14.0_{1.2}$ & $43.2_{3.9}$ & $6.4_{0.7}$ & $8.7_{0.6}$ & $16.0$ & $23.8_{2.3}$ & $49.3_{4.1}$ & $62.8_{3.3}$ & $31.7_{3.9}$ & $34.0_{3.7}$ & $40.3$ & 28.1 \\
LLaMAX3-Alp & 8B & $11.4_{0.9}$ & $28.9_{2.9}$ & $40.4_{3.2}$ & $9.2_{0.8}$ & $23.6_{1.8}$ & $22.7$ & $29.2_{2.1}$ & $41.7_{3.8}$ & $55.4_{4.9}$ & $23.5_{3.0}$ & $40.5_{4.7}$ & $38.1$ & 30.4 \\
GPT-3.5 & -- & $11.6_{0.5}$ & $23.1_{2.0}$ & $76.1_{0.6}$ & $10.1_{0.9}$ & $29.2_{2.1}$ & $30.0$ & $41.6_{2.3}$ & $52.7_{1.5}$ & $77.7_{0.6}$ & $51.7_{1.6}$ & $61.1_{1.1}$ & $56.9$ & 43.5 \\
GPT-4o & -- & $29.6_{1.7}$ & $\textbf{63.8}_{1.9}$ & $\textbf{80.2}_{0.6}$ & $29.6_{2.1}$ & $\textbf{69.5}_{1.6}$ & $\textbf{54.5}$ & $\textbf{69.5}_{1.1}$ & $\textbf{69.3}_{1.7}$ & $\textbf{81.0}_{0.6}$ & $\textbf{73.8}_{1.0}$ & $\textbf{78.2}_{1.1}$ & $\textbf{74.4}$ & \textbf{64.4} \\
\multicolumn{13}{l}{\textbf{SFT on \afridoct}}\\
\rowcolor{blue!20}
LLaMAX3-SFT & 8B & $24.1_{1.6}$ & $29.0_{3.2}$ & $42.2_{4.2}$ & $33.8_{2.8}$ & $33.7_{3.1}$ & $32.6$ & $22.6_{1.8}$ & $22.9_{2.6}$ & $33.1_{4.4}$ & $27.2_{3.6}$ & $31.5_{6.7}$ & $27.5$ & 30.0 \\
\rowcolor{blue!20}
LLama3.1-SFT & 8B & $25.2_{1.8}$ & $31.9_{4.0}$ & $50.2_{6.4}$ & $33.8_{2.8}$ & $38.6_{4.1}$ & $35.9$ & $24.2_{3.7}$ & $24.1_{4.1}$ & $33.7_{5.4}$ & $30.2_{4.7}$ & $29.3_{6.2}$ & $28.3$ & 32.1 \\
\rowcolor{blue!20}
LLaMAX3-SFT$_{10}$ & 8B & $\textbf{37.8}_{2.2}$ & $51.9_{5.0}$ & $74.4_{3.5}$ & $\textbf{52.2}_{3.3}$ & $55.0_{5.5}$ & $54.2$ & $64.0_{3.4}$ & $66.7_{2.8}$ & $77.8_{0.7}$ & $71.8_{1.0}$ & $74.1_{0.9}$ & $70.9$ & 62.6 \\
\rowcolor{blue!20}
LLama3.1-SFT$_{10}$ & 8B & $27.6_{2.4}$ & $49.7_{5.2}$ & $64.1_{5.6}$ & $50.3_{2.8}$ & $47.0_{4.8}$ & $47.8$ & $63.8_{1.1}$ & $61.7_{3.5}$ & $74.4_{3.5}$ & $68.9_{3.4}$ & $71.4_{1.0}$ & $68.0$ & 57.9 \\
\bottomrule
    
  \end{tabular}
  }
  \vspace{-3mm}
  \caption{Performance results of various models on the pseudo-documents ($k=$10) translation task (Health domain), measured using d-chrF. The best prompt was selected for each language after evaluating three different prompts.}
  \vspace{-4mm}
  \label{tab:main_result4}
  \end{center}
\end{table*}

\begin{table*}[t]
 \footnotesize
 \begin{center}
 \resizebox{\textwidth}{!}{%
  \begin{tabular}{p{2.4cm}l|cccccc|cccccc||c}
    \toprule
    \rowcolor{gray!30}
    \textbf{Model} & \textbf{Size} & \multicolumn{6}{c}{\textit{eng {$\rightarrow$} X}} & \multicolumn{6}{c}{\textit{X {$\rightarrow$} eng}} & \textbf{AVG} \\
     \rowcolor{gray!30}
     &  & \textbf{amh} & \textbf{hau} & \textbf{swa} & \textbf{yor} & \textbf{zul} & \textbf{Avg.} & \textbf{amh} & \textbf{hau} & \textbf{swa} & \textbf{yor} & \textbf{zul} & \textbf{Avg.} &   \\
\midrule
\multicolumn{15}{l}{\textbf{Encoder-Decoder}}\\
MADLAD-400 & 3B & $29.5_{2.1}$ & $38.3_{4.3}$ & $31.7_{4.6}$ & $15.1_{1.1}$ & $44.1_{3.6}$ & $31.8$ & $62.6_{2.0}$ & $63.5_{2.2}$ & $66.4_{3.2}$ & $45.9_{2.4}$ & $63.4_{2.2}$ & $60.3$ & 46.0 \\
Aya-101 & 13B & $30.1_{1.5}$ & $55.0_{3.2}$ & $51.7_{3.5}$ & $22.3_{1.7}$ & $55.0_{1.9}$ & $42.8$ & $62.5_{1.4}$ & $65.5_{1.3}$ & $68.8_{1.8}$ & $55.7_{2.4}$ & $68.4_{1.0}$ & $64.2$ & 53.5 \\
\midrule
\multicolumn{15}{l}{\textbf{Decoder-only}}\\
Gemma2-IT & 9B & $6.2_{0.7}$ & $42.1_{3.9}$ & $51.0_{5.3}$ & $6.6_{0.8}$ & $15.4_{1.7}$ & $24.3$ & $35.9_{4.8}$ & $50.1_{4.6}$ & $57.7_{3.7}$ & $48.2_{3.4}$ & $51.7_{3.7}$ & $48.7$ & 36.5 \\
LLama3.1-IT & 8B & $7.4_{0.9}$ & $15.3_{1.9}$ & $43.3_{4.4}$ & $6.2_{1.1}$ & $8.8_{0.7}$ & $16.2$ & $26.1_{2.0}$ & $48.7_{3.4}$ & $59.0_{2.7}$ & $34.4_{3.2}$ & $34.7_{3.1}$ & $40.6$ & 28.4 \\
LLaMAX3-Alp & 8B & $11.4_{1.2}$ & $32.5_{4.4}$ & $38.1_{4.1}$ & $12.0_{1.4}$ & $26.1_{2.2}$ & $24.0$ & $29.4_{2.9}$ & $51.4_{4.3}$ & $62.4_{2.5}$ & $24.7_{3.6}$ & $48.8_{5.3}$ & $43.3$ & 33.7 \\
GPT-3.5 & -- & $13.5_{1.1}$ & $29.7_{2.5}$ & $72.1_{1.6}$ & $12.7_{1.2}$ & $35.1_{2.9}$ & $32.6$ & $38.5_{4.0}$ & $56.3_{1.5}$ & $73.5_{1.4}$ & $53.0_{1.6}$ & $61.2_{1.3}$ & $56.5$ & 44.6 \\
GPT-4o & -- & $31.3_{1.9}$ & $\textbf{65.1}_{2.5}$ & $\textbf{75.1}_{1.6}$ & $28.1_{1.8}$ & $\textbf{70.7}_{1.5}$ & $54.0$ & $\textbf{68.6}_{1.1}$ & $\textbf{71.6}_{1.4}$ & $\textbf{76.5}_{1.6}$ & $\textbf{70.1}_{1.1}$ & $\textbf{76.5}_{1.1}$ & $\textbf{72.7}$ & \textbf{63.3} \\
\multicolumn{13}{l}{\textbf{SFT on \afridoct}}\\
\rowcolor{blue!20}
LLaMAX3-SFT & 8B & $21.7_{2.0}$ & $29.9_{3.2}$ & $37.0_{3.4}$ & $30.5_{2.7}$ & $31.7_{3.5}$ & $30.2$ & $24.2_{2.6}$ & $27.6_{4.2}$ & $32.3_{4.5}$ & $28.5_{3.3}$ & $29.8_{5.4}$ & $28.5$ & 29.3 \\
\rowcolor{blue!20}
LLama3.1-SFT & 8B & $21.0_{2.0}$ & $30.8_{3.2}$ & $40.0_{4.1}$ & $33.4_{3.8}$ & $29.3_{3.1}$ & $30.9$ & $23.9_{2.5}$ & $28.9_{4.3}$ & $36.9_{5.8}$ & $32.2_{4.3}$ & $32.3_{5.2}$ & $30.8$ & 30.9 \\
\rowcolor{blue!20}
LLaMAX3-SFT$_{10}$ & 8B & $\textbf{37.7}_{2.1}$ & $58.6_{5.1}$ & $68.3_{3.9}$ & $49.3_{4.1}$ & $60.9_{3.9}$ & $\textbf{55.0}$ & $65.4_{1.4}$ & $68.5_{1.3}$ & $73.1_{1.2}$ & $67.7_{1.2}$ & $71.6_{1.2}$ & $69.3$ & 62.1 \\
\rowcolor{blue!20}
LLama3.1-SFT$_{10}$ & 8B & $23.7_{1.9}$ & $47.0_{5.2}$ & $58.6_{5.6}$ & $\textbf{49.7}_{3.8}$ & $43.8_{4.5}$ & $44.5$ & $60.9_{2.7}$ & $65.4_{2.5}$ & $71.1_{1.2}$ & $66.3_{1.2}$ & $66.4_{4.0}$ & $66.0$ & 55.3 \\
\bottomrule
    
  \end{tabular}
  }
  \vspace{-3mm}
  \caption{Performance results of various models on the pseudo-documents ($k=$10) translation task (Tech domain), measured using d-chrF. The best prompt was selected for each language after evaluating three different prompts.}
  \label{tab:main_result3}
  \end{center}
\end{table*}

We use GPT-4o as a judge to evaluate translation outputs, following recent work showing LLMs' effectiveness in assessing translation quality and analyzing errors~\citep{wu2024adaptinglargelanguagemodels,sun-etal-2025-fine}. Following~\citet{sun-etal-2025-fine}, we assess each translated document's fluency, content errors (CE), and cohesion errors—specifically lexical (LE) and grammatical (GE) errors—using GPT-4o, with evaluation limited to a few model outputs due to cost constraints (\Cref{sec:llm_judge}). We also complement this with human evaluation for direct assessment scores (\Cref{app:human_eval}) and qualitative analysis through manual inspection (\Cref{app:qualitat_eval}).

\section{Results}
\subsection{Sentence-level Evaluation}
In \Cref{tab:main_result1,tab:main_result2} we present d-chrF scores based on the realigned documents, created by merging the translated sentences into their corresponding documents. We highlight our main findings below, and sentence-level evaluation results using sentence-level metrics are reported in \Cref{sec:app_extra_result}.

\begin{table*}[t]
 \footnotesize
 \begin{center}
 \resizebox{\textwidth}{!}{%
  \begin{tabular}{p{2.5cm}l|cccccc|cccccc}
    \toprule
    \rowcolor{gray!30}
    \textbf{Model} & \textbf{Setup} & \multicolumn{6}{c}{\textit{eng {$\rightarrow$} X}} & \multicolumn{6}{c}{\textit{X {$\rightarrow$} eng}} \\
     \rowcolor{gray!30}
 & &\textbf{amh} & \textbf{hau} &\textbf{swh} &\textbf{yor} & \textbf{zul} & \textbf{avg} &\textbf{amh} & \textbf{hau} &\textbf{swh} &\textbf{yor} & \textbf{zul} & \textbf{avg} \\
\midrule
\multicolumn{14}{l}{\textbf{Fluency}} \\
Aya-101& Sent & 2.8 & 2.9 & 1.3 & 1.2 & 2.7 & 2.2 & 3.0 & 2.9 & 3.3 & 2.3 & 2.9 & 2.9 \\
& Doc10 & 2.5 & 2.9 & 1.8 & 1.4 & 3.1 & 2.3 & 3.6 & 3.2 & 3.8 & 2.9 & 3.3 & 3.4 \\
\addlinespace
GPT-3.5& Sent & 1.1 & 1.4 & 4.9 & 1.1 & 1.5 & 2.0 & 3.0 & 2.5 & 3.6 & 2.6 & 2.6 & 2.9 \\
& Doc10 & 1.0 & 1.1 & 4.9 & 1.0 & 1.3 & 1.9 & 4.2 & 4.1 & 4.7 & 3.9 & 4.0 & 4.2 \\
\addlinespace
LLaMAX3-SFT$_{1}$& Sent & 3.5 & 3.2 & 3.6 & 4.0 & 3.3 & 3.5 & 3.5 & 2.9 & 3.9 & 3.2 & 3.6 & 3.4 \\
& Doc10 & 1.7 & 2.3 & 3.2 & 2.7 & 2.4 & 2.5 & 3.0 & 2.6 & 3.1 & 3.2 & 3.1 & 3.0 \\
\addlinespace
LLaMAX3-SFT$_{10}$& Doc10 & 2.6 & 3.9 & 4.4 & 4.2 & 3.6 & 3.8 & 4.1 & 4.2 & 4.6 & 4.4 & 4.4 & 4.3 \\
\addlinespace
\midrule
\multicolumn{14}{l}{\textbf{Content Error}} \\
Aya-101& Sent & 9.0 & 9.1 & 9.5 & 8.2 & 13.7 & 9.9 & 15.8 & 17.1 & 17.5 & 23.8 & 19.1 & 18.7 \\
& Doc10 & 8.9 & 8.3 & 8.2 & 6.7 & 16.1 & 9.6 & 12.6 & 14.6 & 14.1 & 18.7 & 15.7 & 15.1 \\
\addlinespace
GPT-3.5& Sent & 7.2 & 11.3 & 4.2 & 7.8 & 20.9 & 10.3 & 9.5 & 13.2 & 12.4 & 12.9 & 15.9 & 12.8 \\
& Doc10 & 3.3 & 7.4 & 3.9 & 4.1 & 10.1 & 5.8 & 6.6 & 9.2 & 7.1 & 9.0 & 10.8 & 8.5 \\
\addlinespace
LLaMAX3-SFT$_{1}$& Sent & 10.0 & 9.5 & 12.3 & 12.3 & 12.4 & 11.3 & 11.5 & 9.6 & 11.8 & 12.2 & 12.5 & 11.5 \\
& Doc10 & 10.4 & 8.8 & 9.5 & 8.1 & 8.4 & 9.0 & 9.0 & 9.0 & 8.9 & 9.1 & 8.6 & 8.9 \\
\addlinespace
LLaMAX3-SFT$_{10}$& Doc10 & 7.3 & 14.6 & 10.5 & 10.3 & 12.1 & 11.0 & 8.7 & 9.6 & 8.9 & 10.2 & 9.4 & 9.4 \\
\bottomrule
    
  \end{tabular}
  }
  \vspace{-3mm}
  \caption{Document-level evaluation in the \health domain, judged by GPT-4o. Compares sentence- vs. document-level outputs on Fluency (1–5 scale) and Content Errors (CE).}
  \vspace{-4mm}
  \label{tab:gpt_main}
  \end{center}
\end{table*}

\paragraph{NLLB-200 %
outperforms all other encoder-decoder models across languages and domains}
On average the NLLB models obtain scores of 65.4/66.6 and 64.3/65.0 on \health and \tech domains respectively, with 3.3B outperforming 1.3B except when translating into \yoruba. When translating to English, the worst performing model across the two domains is Toucan. However, %
it gives better results than MADLAD-400 and Aya-101 when translating to African languages. Furthermore, translating to African languages is significantly worse compared to translating to English for all the models.

\paragraph{GPT-4o outperforms other decoder-only counterparts} GPT-4o on average outperforms other decoder-only LMs, with average d-chrF scores of 63.5 and 62.8 for health and tech respectively. The next best performing decoder-only model is LLaMAX3-Alpaca, with d-chrF scores of 54.9 and 55.4. Unlike other open decoder-based LLMs, LLaMAX3-Alpaca was trained on African languages through continued pretraining and adapted via instruction tuning. It outperforms Gemma2-IT by +2.2 in the health domain and +1.7 in the \tech domain, particularly when translating into African languages. In contrast, GPT-3.5 and LLama3.1-IT are the worst performing models. \raggedbottom

\paragraph{Fine-tuning models significantly improves translation quality} We obtain improved performance after fine-tuning NLLB-1.3B on \afridoct, and the resulting model outperforms the 3.3B version without fine-tuning. Similarly, the SFT-based LLMs (LLaMAX3 and LLama3.1) become the best performing open LLMs and outperform their baselines (LLaMAX3-Alpaca and LLama3.1-IT) but below GPT-4o. Overall, our fine-tuned NLLB-200 model is the state-of-the-art model, and our fine-tuned LLaMAX3 is competitive to GPT-4o.

\subsection{Document-level Evaluation}
In \Cref{tab:main_result3,tab:main_result4} we present d-chrF scores based on the best prompt per language for the translation output of the models when evaluated on the realigned documents from pseudo-documents with $k=$10 sentences per pseudo-document.

\paragraph{Pseudo-document translation is worse than sentence-level translation when translating into African languages} Our results from pseudo-document translation show a performance drop across different models compared to sentence-level translation, especially when translating into African languages. However, GPT-4o demonstrates similar and consistent performance in both setups and domains. Additionally, we observe that GPT-3.5 is the next best performing decoder-only LLM, which contrasts with its performance in sentence-level translation. Gemma2-IT outperforms LLaMAX3-Alpaca especially when translating into English, which also differs from the trends observed in the sentence-level setup.

\paragraph{LLMs trained on longer documents are better for long document translation} Both LLama models trained via SFT on sentences (LLama3.1-SFT, and LLaMAX3-SFT) show a decline in performance in the pseudo-document setting compared to sentence-level translation. However, the same models trained via SFT on pseudo-documents with $k$=10 demonstrate significant improvements on pseudo-documents. Interestingly, the LLaMAX3-SFT$_{10}$ model performs consistently well, achieving results comparable to its sentence-level counterpart on sentence-level tasks, and also outperforming LLama3.1-SFT$_{10}$, particularly when translating into African languages.

\begin{figure*}[t]
\setlength{\belowcaptionskip}{-4pt}
  \centering
    \begin{subfigure}{0.235\textwidth}
        \includegraphics[width=\textwidth]{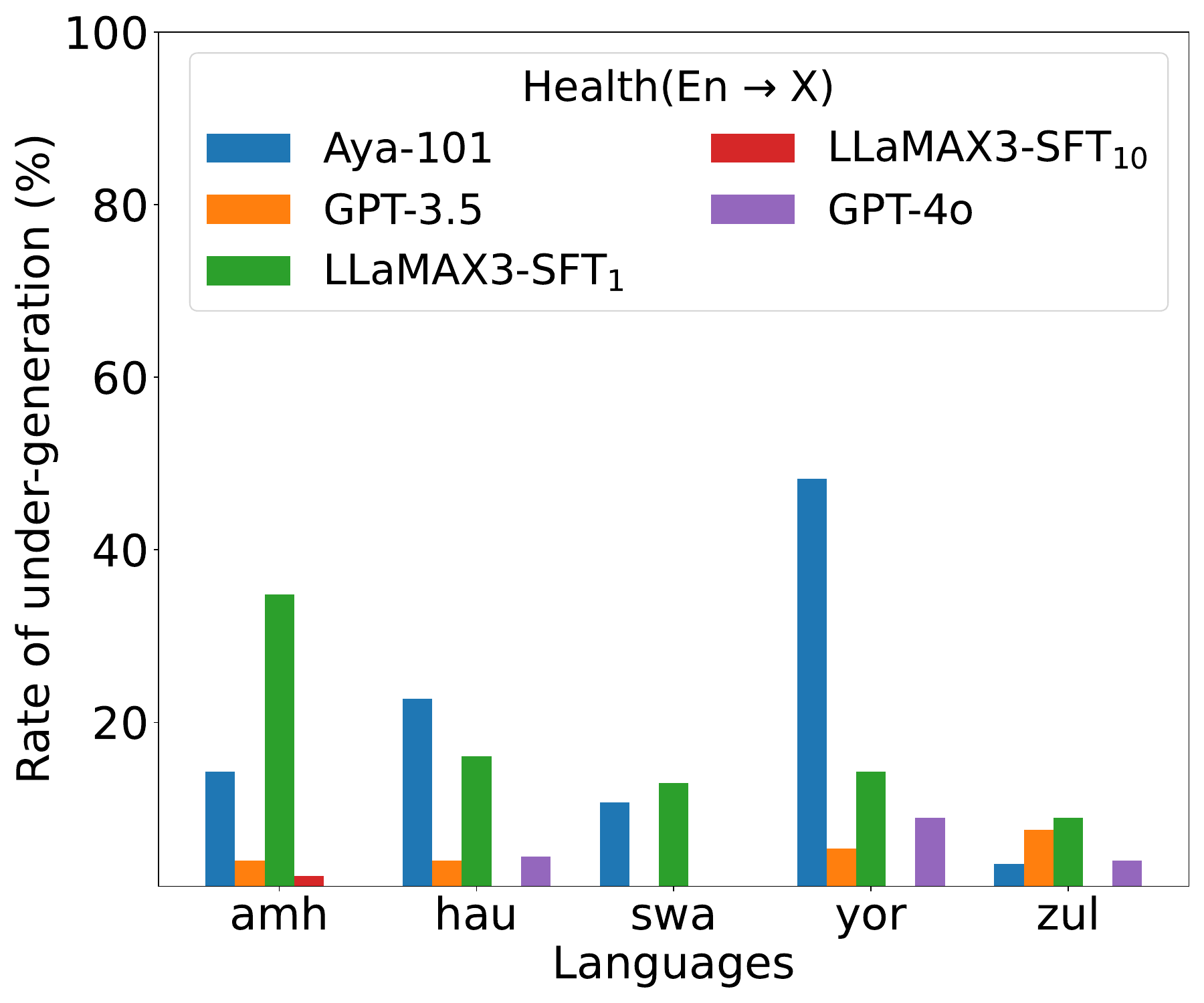}
        \caption{}
        \label{fig:undergen2x}
    \end{subfigure}
     ~
    \begin{subfigure}{0.235\textwidth}
        \includegraphics[width=\textwidth]{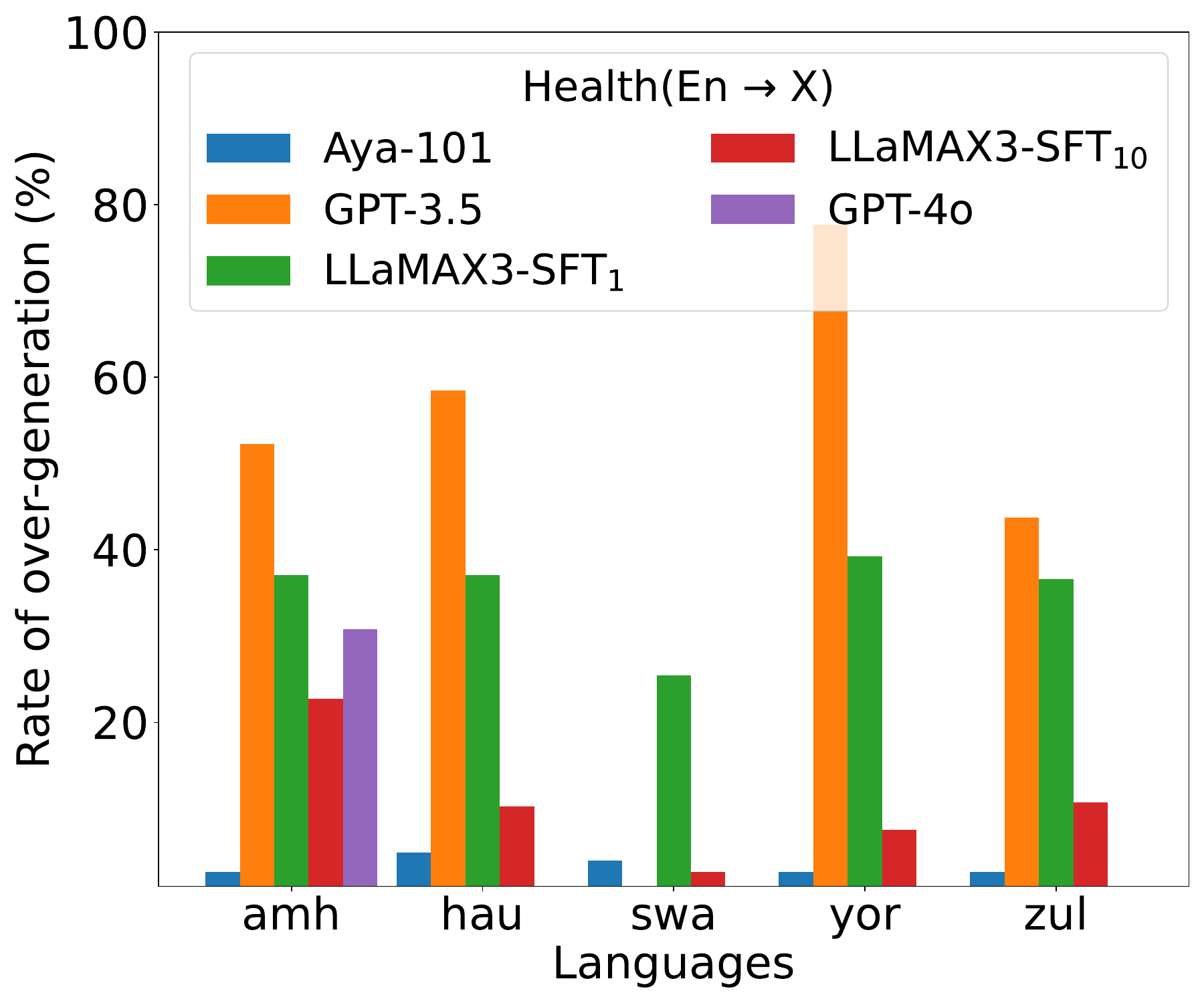}
        \caption{}
        \label{fig:undergen2en}
    \end{subfigure}
     ~     
     \begin{subfigure}{0.235\textwidth}
         \includegraphics[width=\textwidth]{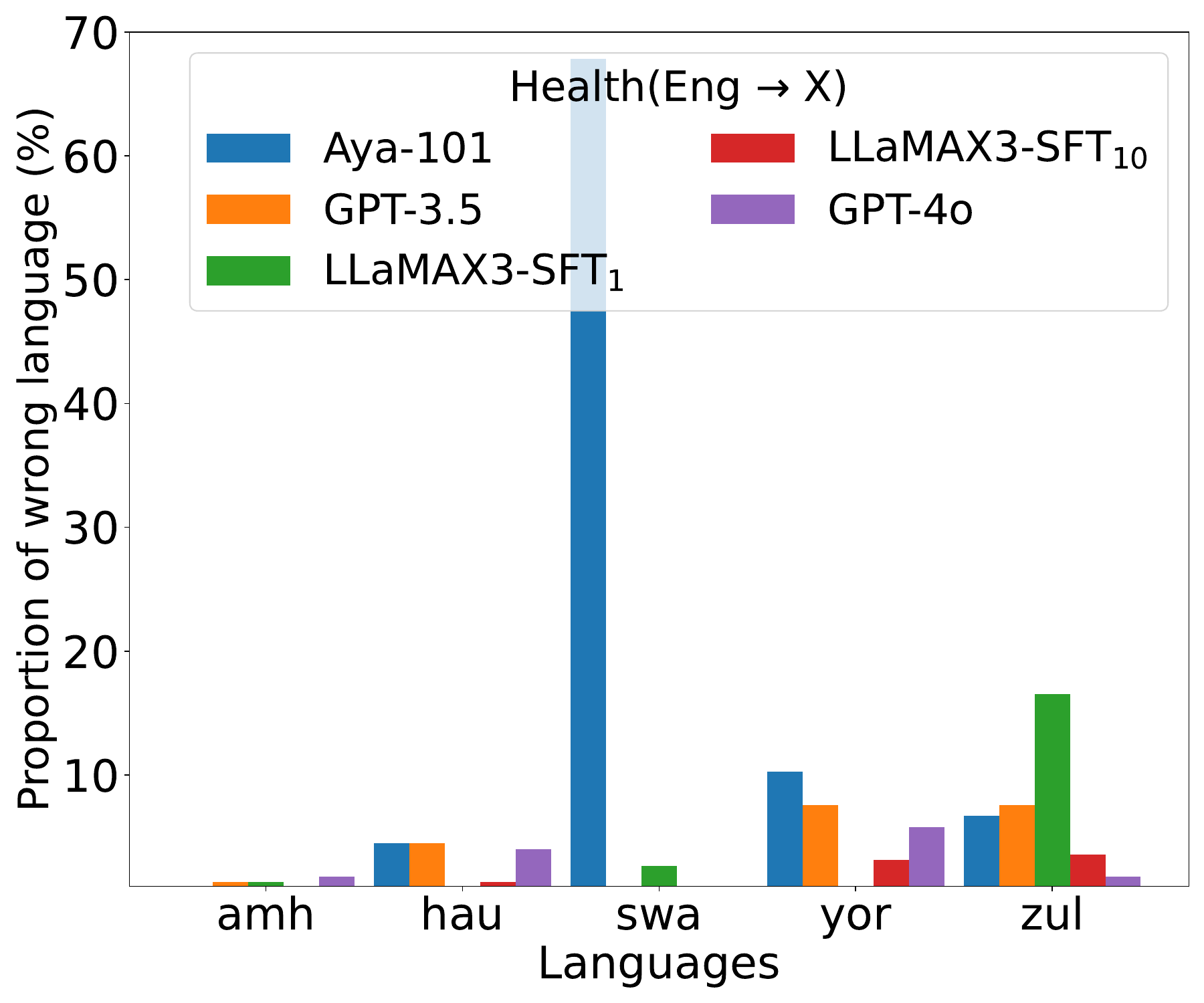}
         \caption{}
         \label{fig:mistranhealthen2x}
     \end{subfigure}
     ~
    \begin{subfigure} {0.235\textwidth} 
        \includegraphics[width=\textwidth]{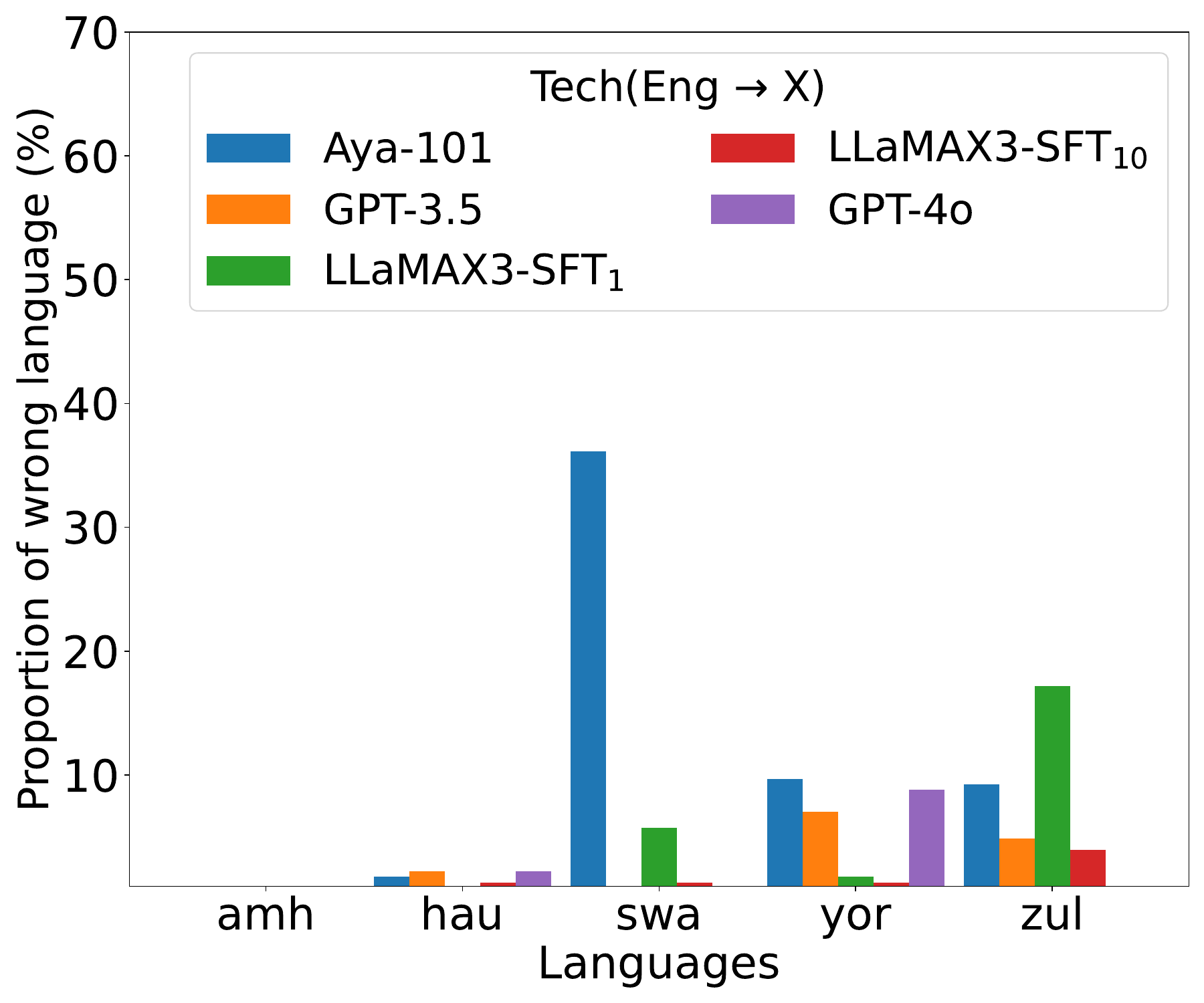}
        \caption{}
    \label{fig:mistrantechen2x}
    \end{subfigure}
\caption{Rate of under-generation, over-generation, and off-target translation in pseudo-document translation ($k=10$).}
\label{fig:issues}
\end{figure*}

\subsection{GPT-4o based evaluation} 
\Cref{tab:gpt_main} presents average GPT-4o evaluations for fluency and content errors (CE) of realigned outputs from sentence-level and pseudo-document-level tasks (k=10) across four models in the \health domain. When translating into English, pseudo-document outputs are generally rated as more fluent and show fewer content errors, except for LLaMAX3-SFT$_1$, which, when evaluated on pseudo-documents, shows lower fluency but still fewer content errors—an outcome that is counterintuitive. However, when translating into African languages, the results are less consistent. Notably, GPT-3.5 achieves a fluency score of 4.9 for both fine-tuned versions of LLaMAX3-SFT—a score no model achieved when translating into English. Additionally, \yoruba, a language that had some of the lowest d-chrF scores across models, achieved fluency scores of 4.0 and 4.2 with the two LLaMAX3-SFT versions. These inconsistencies raise concerns about GPT-4o’s reliability. Consequently, we focus on human evaluation going forward. Full GPT-4o results are provided in \Cref{sec:app_gpt_analysis_result}.

\subsection{Human evaluation} 
In \Cref{tab:human_eval_health} we report average direct assessment (DA) scores (on a scale from 0 to 100) from three annotators per language for the \health domain, when translating into four African languages. For each language, we used 30 documents across models and both domains to compute inter-annotator agreement. We obtained Krippendorff’s alpha values of $\geq$ 0.40, which are relatively low due to the fine granularity of the evaluation scale. \textbf{Human evaluation results align closely with d-chrF}, which favors sentence-level translations over pseudo-document translations when translating into African languages. Among the models, LLaMAX3-SFT$_1$ receives higher ratings at the sentence-level but is rated lower when translating pseudo-documents. In contrast, LLaMAX3-SFT$_{10}$ receives slightly lower ratings than LLaMAX3-SFT$_1$ at the sentence-level but is rated higher in the pseudo-document setting. GPT-3.5 is generally rated as the weakest model across languages, except for Swahili, where its performance is comparatively better  (see \Cref{app_human_eval_result} for details).

\begin{table}[t]
\small\centering
\scalebox{0.85}{
  \begin{tabular}{ll|ccccc}
  \toprule
  \textbf{Model} & \textbf{Setup} & \textbf{amh} & \textbf{hau} & \textbf{swh} & \textbf{yor} & \textbf{zul} \\
  \midrule
  \multirow{2}{*}{GPT-3.5} &Sent & 14.6 & 29.6 & 66.5 & 7.5 & 9.2 \\
& Doc10 & 1.7 & 16.4 & \textbf{68.3} & 4.2 & 3.1 \\
\addlinespace
\multirow{2}{*}{LLaMAX3-SFT$_{1}$} &Sent & \textbf{64.5} & \textbf{81.5} & 57.9 & \textbf{65.1} & \textbf{48.1} \\ 
&Doc10 & 27.4 & 45.7 & 50.6 & 44.3 & 22.5 \\
\addlinespace
LLaMAX3-SFT$_{10}$ & Doc10 & 38.5 & 76.7 & 62.4 & 64.9 & 46.7 \\
  \bottomrule
  \end{tabular}
  }
  \vspace{-3mm}
  \caption{Average DA score (scale 0–100) from the human evaluators per language in the \health domain.}
  \label{tab:human_eval_health}
\end{table}

\subsection{Qualitative evaluation} 
Our qualitative analysis, based on feedback from native speakers who are also authors, indicates that GPT-3.5 frequently over-generates in the pseudo-document setup by repeating words and phrases—except in Swahili, where it performs best.
However, for \yoruba, it often uses inconsistent or partial diacritics, resulting in inaccuracies. LLaMAX3-SFT$_1$ also exhibits repetition in pseudo-document translations, likely due to a length generalization problem~\citep{anil2022exploring}, and does so more than LLaMAX3-SFT$_{10}$. For the other four languages, LLaMAX3-SFT$_1$ with the sentence-level setup was rated higher than other models and configurations, owing to better context preservation and fewer repetitions. These observations are consistent with both d-chrF and DA scores, although d-chrF scores tend to be inflated.

\begin{figure}[t]
\setlength{\belowcaptionskip}{-10pt}
\setlength{\abovecaptionskip}{-5pt}   %
  \centering
    \begin{subfigure}{0.48\columnwidth}
        \includegraphics[width=\textwidth]{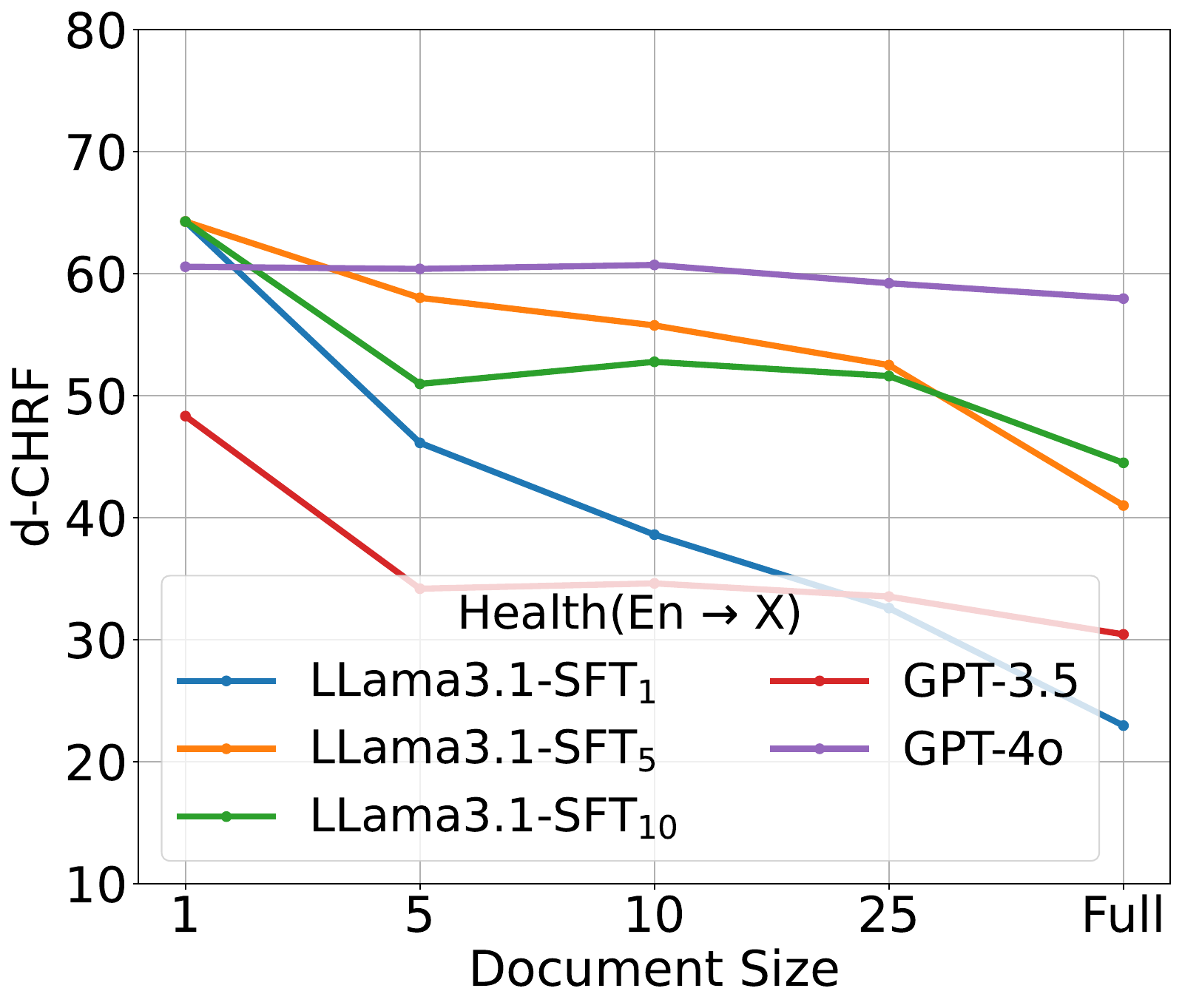}
        \label{fig:lengthperf2x}
    \end{subfigure}
    ~
    \begin{subfigure}{0.48\columnwidth}
        \includegraphics[width=\textwidth]{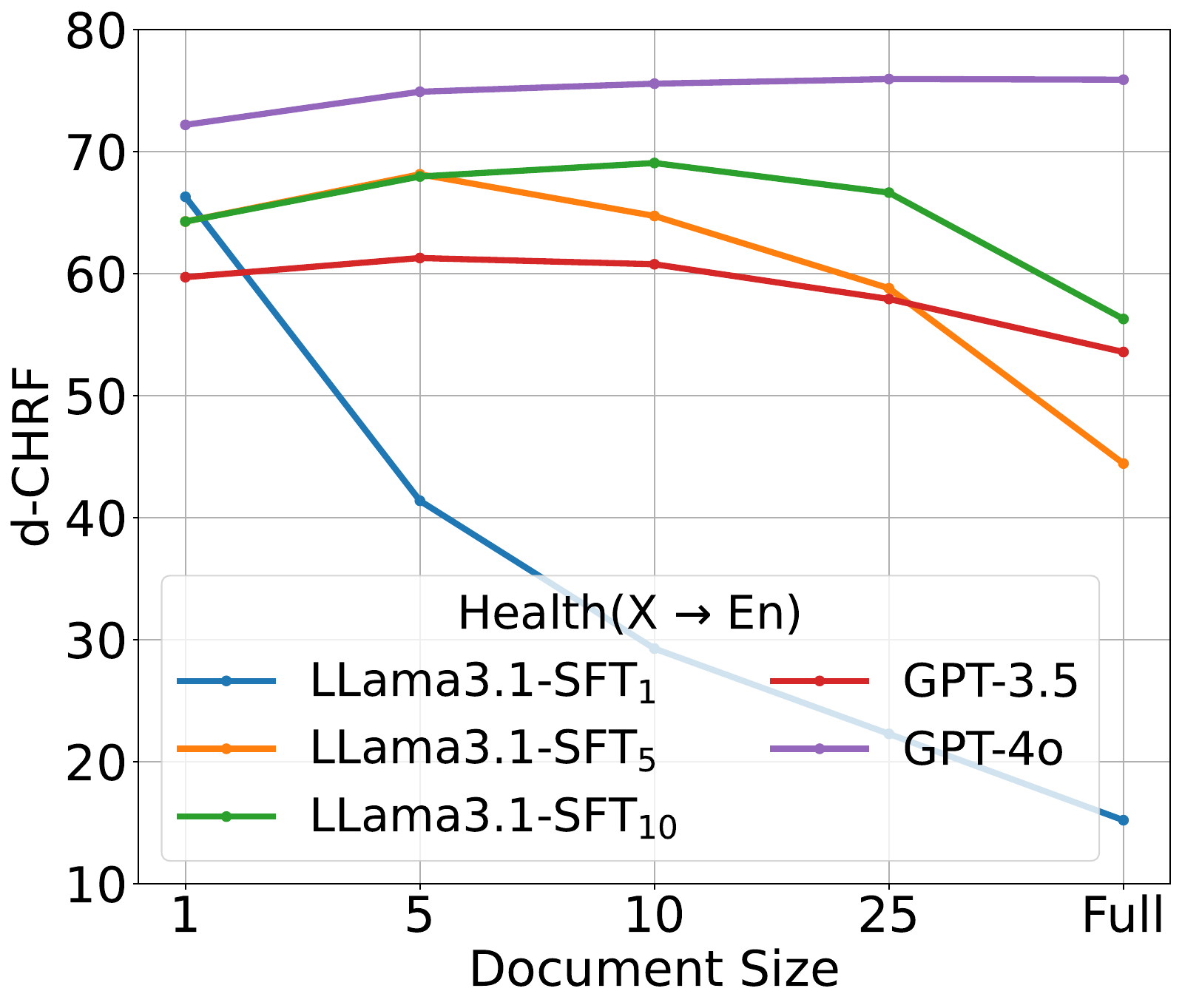}
        \label{fig:lengthperf2en}
    \end{subfigure}
\vspace{-2em}
\caption{Comparison of Average d-chrF scores across models and pseudo-document lengths.}
\label{fig:lengthperf}
\end{figure}

\section{Discussion and Analysis}
To better understand model behavior, we analyze their pseudo-document ($k=10$) translation outputs based on our findings and common issues in document-level MT with LLMs~\citep{wu2024adaptinglargelanguagemodels,wang-etal-2024-benchmarking}. Additional results are provided in \Cref{sec:app_extra_analysis}.

\paragraph{Are the outputs generated by translation models of appropriate length?} We compare model translations to the reference translations to identify empty outputs and cases of under- or over-generation. We found that all models rarely generate empty translations (refer to \Cref{sec:app_extra_analysis}), although GPT-3.5 and GPT-4o showed a slight tendency to do so for \yoruba and Zulu, occurring in under 10\% of cases. To  evaluate this we compute the compute model output to reference translation length ratio across five models and for the languages, for the evaluated models and took the lowest 10th percentile, and this returns 73\% and upper percentile of 550\%. However, we defined under-generation as outputs <70\% of reference length, and over-generation as >143\%. As shown in \Cref{fig:undergen2x,fig:undergen2en} and consistent with our qualitative findings, GPT-3.5 tends to over-generate more in African languages except for Swahili, while LLaMAX3-SFT$_{1}$ often under-generates as a sentence-level model. Moreover, all models over-generated by about 20\% for Amharic, likely due to its unique script.

\paragraph{Do LLMs generate translations in the correct target languages?} We evaluate whether these models understand the task by generating outputs in the target languages using the OpenLID~\citep{burchell-etal-2023-open} language identification model. Our results show that these models rarely generate outputs in the wrong language when translating to English. However, when translating to African languages, there is a higher likelihood of incorrect language translations, particularly with open models (see \Cref{fig:mistranhealthen2x,fig:mistrantechen2x}). These off-target languages often include English, and other languages including other African languages.

\paragraph{What is the effect of document length on translation quality?} We compare average d-chrF scores of selected models, including GPT-3.5/4 and LLama3.1-SFT$_k$ ($k$ = 1, 5, 10), evaluated across pseudo-document lengths of 1, 5, 10, 25, and full length. As shown in \Cref{fig:lengthperf}, d-chrF scores generally decline with increasing document length for African language translations. The reverse translation direction shows a similar trend, except for GPT-4o, which improves with length. Models trained on longer documents also generalize better to longer inputs than those trained on sentences.

\section{Conclusion}
\label{sec:conclusion}
We introduce \afridoct, a document-level translation dataset in the \health and \tech domains for five African languages. We benchmarked various models, fine-tuning selected ones. Due to context length limits, documents were translated either sentence by sentence or as pseudo-documents. Outputs were evaluated using standard MT metrics, GPT-4o as a judge, and human direct assessment. NLLB-200 was the strongest built-in MT model, while GPT-4o outperformed general-purpose LLMs. However, our DA and qualitative analysis found GPT-4o's judgments inconsistent for African languages, and sentence-by-sentence translation proved more effective for some languages.

\section{Limitations}
\paragraph{Choice of LLMs and Prompts} We evaluated only a small subset of the numerous multilingual LLMs available. Our experiments were also limited by the context length of the LLMs, particularly for open LLMs. Except for LLama3.1, all other open LLMs have a context length of 8192 tokens, while encoder-decoder models were primarily based on T5. This makes it difficult to use the context length beyond a certain limit, making full document translation infeasible. Additionally, LLMs are prone to variance in performance based on the prompt. We therefore evaluated them for translation using three different prompts. However, it is possible that our prompts were not optimal.

\paragraph{Language Coverage} Africa is home to thousands of indigenous languages, many of which exhibit unique  linguistic properties. However, due to the high cost of translation using human translators and limited available funding, it is currently impossible to cover all languages. As a result, we focused on just five languages. We hope that future work will expand this dataset to include more languages and inspire the creation of additional datasets with broader coverage for document-level translation. 
Similarly, \afridoct is a multi-way parallel dataset. However, due to the cost of running inference over three prompts and across all 30 translation directions for all the models evaluated, most of our analysis is limited to translation tasks between English and the five African languages. While we fine-tuned NLLB-200, LLama3.1 and LLaMAX3 on all 30 directions, we only provide results from NLLB-200 for all directions both before and after fine-tuning for sentence-level and pseudo-document tasks in the \Cref{sec:app_extra_analysis}. 

\paragraph{Evaluation Metrics} Quality evaluation in MT is an open and ongoing area of research, especially for document-level translation. Recent works have proposed embedding-based metrics for evaluation at both the sentence and document levels. While this has been well explored for high-resource language pairs, it remains underexplored for African languages, although there is a tool, AfriCOMET, that works for sentence-level evaluation in African languages. However, we evaluated the document-level translation outputs using \emph{ModernBERT-base-long-context-qe-v1}\footnote{\url{https://huggingface.co/ymoslem/ModernBERT-base-long-context-qe-v1}}, trained on the WMT human evaluation dataset across 41 language pairs, including over 20 languages and three African languages (Hausa, Xhosa, and Zulu), two of which are included to our work. However, the scores were nearly identical across all models, offering no meaningful differentiation. Hence, for our document-level evaluation, in addition to lexical-based metrics, we incorporated three other evaluation approaches: using GPT-4o as a judge, human evaluation, and qualitative analysis. GPT-4o was employed to assess and rate the translation outputs of four models. While its ratings were consistent for translations into English, the same was not observed for translations into African languages, likely due to the model’s limited understanding of these languages. Therefore, we conducted a human evaluation for translations from English to African languages, comparing only three models due to cost constraints. However, we were unable to recruit annotators for Zulu. %

\paragraph{Model Coverage and Evaluation Focus} While we fine-tuned both NLLB-1.3B and LLaMAX3 models across all 30 language directions, due to computational constraints and the high cost of qualitative evaluation, our detailed analysis focuses only on translation between English and the 5 African languages. Nevertheless, we report quantitative results across all 30 directions for NLLB-1.3B. We will make all fine-tuned models publicly available to support future work, and we hope that further research will explore the remaining translation directions in greater depth.

\paragraph{Translationese and English-Centric Bias} A potential limitation of our dataset is the influence of translationese~\citep{koppel-ordan-2011-translationese}. Since all source material translated originates in English, translated sentences in African languages may exhibit patterns such as unnatural syntax or overly literal phrasing. Although we have not conducted an analysis to quantify these effects, prior work suggests that they can affect MT model performance, generalization and evaluation including direct assessment~\citep{freitag-etal-2019-ape,edunov-etal-2020-evaluation}. Furthermore, \afridoct may reflect a bias toward English in terms of structure, semantics, and cultural framing. We leave a deeper investigation of these issues to future work. 

\section*{Ethics Statement}
\afridoct was created with the utmost consideration for ethical standards. The English texts translated were sourced from publicly available and ethically sourced materials. The data sources were selected to represent different cultural perspectives, with a focus on minimizing any potential bias. Efforts were made to ensure the dataset does not include harmful, biased, or offensive content via manual inspection. 

\section*{Acknowledgements}
This work was carried out with support from Lacuna Fund, an initiative co-founded by The Rockefeller Foundation, Google.org, and Canada’s International Development Research Centre. This research project has benefited from the Microsoft Accelerate Foundation Models Research (AFMR) grant program. We acknowledge the support of OpenAI for providing API credits through their Researcher Access API programme. 
Jesujoba Alabi acknowledges the support of the BMBF’s (German Federal Ministry of Education and Research) SLIK project under the grant 01IS22015C. The work has also been partially funded by BMBF through the TRAILS project (grant number 01IW24005). Miaoran Zhang received funding from the DFG (German Research Foundation) under project 232722074, SFB 1102.
CEB acknowledges her AI4S fellowship within the “Generación D” initiative by Red.es, Ministerio para la Transformación Digital y de la Función Pública, for talent attraction (C005/24-ED CV1), funded by NextGenerationEU through PRTR.
R. Bawden's participation was partly funded by her chair in the PRAIRIE institute, funded by the French national agency ANR, as part of the ``Investissements d’avenir'' programme under the reference ANR-19-P3IA-0001 and by the French \textit{Agence Nationale de la Recherche} (ANR) under the projects TraLaLaM (``ANR-23-IAS1-0006'') and MaTOS (``ANR-22-CE23-0033''). David Adelani acknowledges the funding of IVADO and the Canada First Research Excellence Fund. 

Also, we also appreciate the translators whose names we have listed below:
\begin{enumerate}
    \item \textbf{Amharic:} Bereket Tilahun, Hana M. Tamiru, Biniam Asmlash, Lidetewold Kebede
    \item \textbf{Hausa:} Junaid Garba, Umma Abubakar, Jibrin Adamu, Ruqayya Nasir Iro
    \item \textbf{Swahili:} Mohamed Mwinyimkuu, Laanyu koone, Baraka Karuli Mgasa, Said Athumani Said
    \item \textbf{\yoruba:} Ifeoluwa Akinsola, Simeon Onaolapo, Ganiyat Dasola Afolabi, Oluwatosin Koya
    \item \textbf{Zulu:} Busisiwe Pakade, Rooweither Mabuya, Tholakele Zungu, Zanele Thembani
\end{enumerate}

Lastly, we thank the human annotators who were not part of the translation team.

\bibliography{anthology,custom}
\bibliographystyle{acl_natbib}

\appendix

\section{More details about \afridoct}
\label{sec:more_afridoc}

\begin{figure*}[h]
    \centering
    \includegraphics[width=\textwidth]{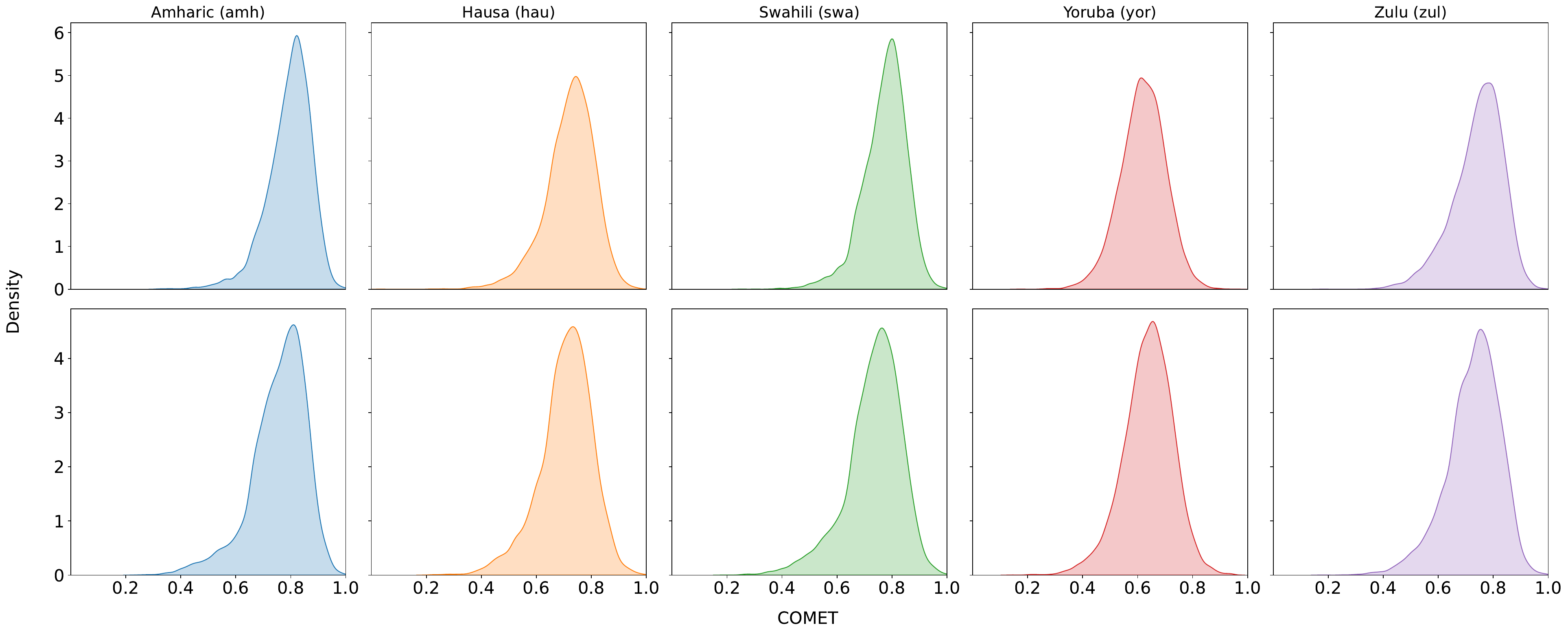}  %
    \caption{Distribution of the quality estimation of of the translated sentences using COMET scores for the \health (top), \tech (bottom).}
    \label{fig:distribution1}
\end{figure*}

\subsection{Translation guidelines}
\label{sec:app_translation_guide}
The translation guidelines, aside the details shared at the workshop on translation and terminology creation can be found below. 
\begin{itemize}
    \item Thank you for agreeing to work on this project. Below is the link to access the data for translation. The files are in .csv format, and you can open them using Google Sheets or Microsoft Excel (for offline work). 
    \item Each file contains 2500 sentences, and they are named in the format of a serial number followed by your first name.
    \item  Please do not  delete double empty rows, as they serve to separate paragraphs. Also, avoid deleting any rows, columns, or provided text. 
    \item Use the language field to input the translations. It is essential not to rely on translation engines, as our quality assurance process can detect this. Depending on such tools may result in potential issues that you would need to address, leading to additional work on your part.
    \item We will provide a list of extracted terminologies soon so that you can harmonize how terminologies are translated. 
    \item Thank you for your attention to these guidelines. Should you have any questions, concerns, or suggestions, feel free to contact us or reach out to your language coordinator.
\end{itemize}

\subsection{Creation of pseudo-documents for \afridoct} 
Given that the translated documents vary in length in terms of sentences and tokens, and considering the maximum token length limitations of the different LLMs used, we adopted a chunking approach for document-level evaluation. In this approach, documents were divided into smaller pseudo-documents that fit within the maximum length constraints of the models. To establish an appropriate chunk size, each document was divided into fixed-size chunks of $k$ sentences, with the possibility that the final chunk may contain fewer than $k$ sentences. These sentence groups, referred to as pseudo-documents, were used for document-level translation.

We conducted an initial analysis, testing different values for $k$ (5, 10, and 25), with $k$=1 serving as our sentence-level setup. \Cref{tab:app_doc_stats} presents the resulting number of parallel pseudo-documents, as well as the average number of tokens per pseudo-document per language for the various model tokenizers, including the 95th percentile token count. Our analysis revealed that Amharic and \yoruba—languages with unique characteristics such as non-Latin scripts and diacritics, respectively—had the largest average token counts across the tokenizers. Additionally, the domain with the highest number of average tokens for pseudo-document varies from language to language.

To accommodate both languages in our experiments, we chose pseudo-documents with $k$=10. However, for the SFT models described in \Cref{sec:app_sft}, we used both $k$=5 and $k$=10.
\begin{table*}[ht]
 \small
 \begin{center}
 \resizebox{\textwidth}{!}{%
  \begin{tabular}{ll|cc|cc|cc|cc}
    \toprule
    \rowcolor{gray!30}
    \textbf{Languages/Split} & \textbf{Models} & \multicolumn{2}{c}{\textbf{Full}} & \multicolumn{2}{c}{\textbf{25 sent.}} & \multicolumn{2}{c}{\textbf{10 sent.}} & \multicolumn{2}{c}{\textbf{5 sent.}} \\
    \rowcolor{gray!30}
    & & \textbf{Health} & \textbf{Tech} & \textbf{Health} & \textbf{Tech} & \textbf{Health} & \textbf{Tech} & \textbf{Health} & \textbf{Tech} \\
\midrule
\multicolumn{10}{l}{\textbf{Sizes of data splits in \afridoct pseudo-document}} \\
Train & {} & $240$ & $187$ & $402$ & $369$ & $812$ & $789$ & $1506$ & $1483$  \\
Dev & {} & $33$ & $25$ & $56$ & $48$ & $112$ & $106$ & $209$ & $204$  \\
Test & {} & $61$ & $59$ & $108$ & $106$ & $224$ & $227$ & $417$ & $418$  \\
\midrule
\midrule
\multicolumn{10}{l}{\textbf{Statistics of LLM tokens in \afridoct pseudo-document training splits}} \\
\multirow{5}{*}{ en } & NLLB-200 & $923.7$/$2017.6$ & $941.9$/$1982.1$ & $551.5$/$951.7$ & $477.4$/$758.8$ & $273.0$/$430.9$ & $223.2$/$343.6$ & $147.2$/$233.8$ & $118.8$/$184.9$ \\
& MADLAD-400 & $971.0$/$2095.2$ & $991.4$/$2100.1$ & $579.7$/$1017.1$ & $502.4$/$797.8$ & $287.0$/$449.3$ & $235.0$/$362.0$ & $154.7$/$245.0$ & $125.0$/$196.9$ \\
& Aya-101 & $1008.2$/$2183.5$ & $1020.5$/$2184.3$ & $601.9$/$1038.0$ & $517.2$/$820.2$ & $298.0$/$463.4$ & $241.9$/$372.6$ & $160.7$/$255.0$ & $128.7$/$199.0$ \\
& LLama3 & $801.4$/$1788.0$ & $842.5$/$1798.4$ & $478.5$/$833.8$ & $427.0$/$664.0$ & $236.9$/$372.9$ & $199.7$/$304.2$ & $127.8$/$203.0$ & $106.3$/$166.0$ \\
& Gemma-2 & $802.9$/$1820.1$ & $857.9$/$1857.6$ & $479.3$/$841.0$ & $434.8$/$689.6$ & $237.3$/$375.0$ & $203.4$/$314.0$ & $128.0$/$205.0$ & $108.2$/$169.0$ \\
& ModernBERT & $801.0$/$1800.7$ & $862.7$/$1819.3$ & $478.3$/$837.8$ & $437.3$/$685.6$ & $236.8$/$373.4$ & $204.5$/$311.0$ & $127.8$/$204.0$ & $108.9$/$171.0$ \\
\midrule

\multirow{5}{*}{ am } & NLLB-200 & $1304.4$/$2785.8$ & $1376.3$/$2888.7$ & $778.8$/$1329.9$ & $697.5$/$1130.8$ & $385.6$/$592.0$ & $326.2$/$520.0$ & $207.9$/$328.0$ & $173.5$/$282.9$ \\
& MADLAD-400 & $1624.8$/$3393.6$ & $1685.0$/$3487.4$ & $970.0$/$1684.2$ & $853.9$/$1380.4$ & $480.2$/$750.0$ & $399.4$/$640.2$ & $258.9$/$413.8$ & $212.5$/$342.9$ \\
& Aya-101 & $1887.4$/$3937.9$ & $1934.7$/$4126.9$ & $1126.8$/$1931.8$ & $980.5$/$1598.0$ & $557.9$/$855.4$ & $458.5$/$722.0$ & $300.8$/$477.8$ & $244.0$/$390.0$ \\
& LLama3 & $6798.0$/$13986.2$ & $6829.6$/$14750.9$ & $4058.5$/$6971.8$ & $3461.1$/$5584.8$ & $2009.3$/$3084.4$ & $1618.7$/$2560.8$ & $1083.3$/$1716.0$ & $861.2$/$1379.9$ \\
& Gemma-2 & $2817.9$/$5857.5$ & $2868.4$/$6227.4$ & $1682.1$/$2896.4$ & $1453.2$/$2342.4$ & $832.4$/$1267.8$ & $679.3$/$1071.6$ & $448.5$/$710.0$ & $361.0$/$575.0$ \\
& ModernBERT & $7347.8$/$15045.1$ & $7386.4$/$15952.3$ & $4386.4$/$7544.1$ & $3742.8$/$6035.8$ & $2171.1$/$3331.2$ & $1749.9$/$2760.4$ & $1170.2$/$1851.0$ & $930.6$/$1485.9$ \\
\midrule

\multirow{5}{*}{ ha } & NLLB-200 & $1204.4$/$2713.7$ & $1171.4$/$2463.0$ & $719.0$/$1252.8$ & $593.6$/$962.6$ & $356.0$/$554.0$ & $277.6$/$430.6$ & $191.9$/$306.8$ & $147.7$/$232.0$ \\
& MADLAD-400 & $1297.1$/$2849.4$ & $1260.5$/$2643.7$ & $774.4$/$1359.7$ & $638.8$/$1042.0$ & $383.4$/$606.4$ & $298.8$/$465.6$ & $206.7$/$329.0$ & $158.9$/$251.0$ \\
& Aya-101 & $1614.9$/$3497.4$ & $1535.3$/$3241.9$ & $964.1$/$1672.3$ & $778.0$/$1254.6$ & $477.3$/$742.6$ & $363.9$/$563.2$ & $257.4$/$410.8$ & $193.6$/$306.0$ \\
& LLama3 & $1916.7$/$4012.9$ & $1822.6$/$3917.9$ & $1144.3$/$1988.8$ & $923.7$/$1513.6$ & $566.6$/$882.4$ & $432.1$/$674.6$ & $305.5$/$488.8$ & $230.0$/$365.9$ \\
& Gemma-2 & $1642.4$/$3568.9$ & $1581.3$/$3373.4$ & $980.6$/$1716.7$ & $801.4$/$1297.8$ & $485.5$/$757.4$ & $374.8$/$584.0$ & $261.8$/$417.8$ & $199.4$/$317.8$ \\
& ModernBERT & $1998.5$/$4207.5$ & $1916.8$/$4139.7$ & $1193.1$/$2057.8$ & $971.5$/$1575.8$ & $590.8$/$916.9$ & $454.4$/$701.0$ & $318.6$/$510.8$ & $241.8$/$382.9$ \\
\midrule

\multirow{5}{*}{ sw } & NLLB-200 & $1100.8$/$2494.8$ & $1094.8$/$2187.5$ & $657.2$/$1145.9$ & $554.8$/$896.4$ & $325.4$/$517.0$ & $259.5$/$409.6$ & $175.4$/$280.0$ & $138.1$/$218.0$ \\
& MADLAD-400 & $1177.3$/$2629.9$ & $1155.3$/$2293.9$ & $702.8$/$1227.6$ & $585.5$/$938.6$ & $348.0$/$547.0$ & $273.8$/$436.0$ & $187.6$/$297.0$ & $145.7$/$231.9$ \\
& Aya-101 & $1345.3$/$2925.0$ & $1311.0$/$2667.8$ & $803.2$/$1390.9$ & $664.4$/$1076.2$ & $397.6$/$627.9$ & $310.7$/$487.4$ & $214.4$/$339.0$ & $165.3$/$261.0$ \\
& LLama3 & $1668.1$/$3605.0$ & $1619.4$/$3364.9$ & $995.9$/$1735.4$ & $820.7$/$1330.0$ & $493.1$/$771.4$ & $383.9$/$599.8$ & $266.0$/$418.0$ & $204.3$/$323.0$ \\
& Gemma-2 & $1413.3$/$3097.3$ & $1377.1$/$2770.0$ & $843.8$/$1467.7$ & $697.9$/$1126.2$ & $417.8$/$658.9$ & $326.4$/$513.0$ & $225.3$/$356.8$ & $173.7$/$277.9$ \\
& ModernBERT & $1757.9$/$3753.4$ & $1719.7$/$3594.1$ & $1049.5$/$1822.8$ & $871.6$/$1421.0$ & $519.7$/$810.0$ & $407.7$/$632.0$ & $280.2$/$441.0$ & $217.0$/$342.8$ \\
\midrule

\multirow{5}{*}{ yo } & NLLB-200 & $1702.6$/$3854.7$ & $1724.8$/$3577.1$ & $1016.5$/$1857.2$ & $874.1$/$1428.6$ & $503.2$/$814.7$ & $408.8$/$644.6$ & $271.3$/$443.8$ & $217.5$/$348.9$ \\
& MADLAD-400 & $1983.6$/$4470.9$ & $1990.4$/$4136.7$ & $1184.3$/$2137.5$ & $1008.7$/$1650.2$ & $586.3$/$939.4$ & $471.7$/$742.2$ & $316.1$/$512.0$ & $251.0$/$401.9$ \\
& Aya-101 & $2729.2$/$5832.3$ & $2659.8$/$5549.7$ & $1629.4$/$2956.4$ & $1347.9$/$2211.6$ & $806.7$/$1292.4$ & $630.4$/$988.0$ & $434.9$/$704.0$ & $335.4$/$544.0$ \\
& LLama3 & $2945.8$/$6322.4$ & $2880.0$/$5995.5$ & $1758.6$/$3203.9$ & $1459.4$/$2400.4$ & $870.5$/$1406.0$ & $682.5$/$1077.6$ & $469.3$/$767.8$ & $363.0$/$585.9$ \\
& Gemma-2 & $2620.4$/$5745.5$ & $2593.5$/$5406.9$ & $1564.3$/$2867.7$ & $1314.3$/$2143.8$ & $774.4$/$1245.4$ & $614.6$/$965.6$ & $417.4$/$678.0$ & $327.0$/$530.0$ \\
& ModernBERT & $3648.3$/$7780.9$ & $3595.2$/$7600.6$ & $2178.1$/$4002.0$ & $1822.0$/$3020.4$ & $1078.3$/$1761.4$ & $852.1$/$1339.8$ & $581.4$/$957.2$ & $453.3$/$733.9$ \\
\midrule

\multirow{5}{*}{ zu } & NLLB-200 & $1201.8$/$2513.3$ & $1230.4$/$2555.7$ & $717.5$/$1233.0$ & $623.5$/$1016.6$ & $355.2$/$554.3$ & $291.6$/$461.2$ & $191.5$/$300.0$ & $155.1$/$250.0$ \\
& MADLAD-400 & $1215.2$/$2524.0$ & $1230.7$/$2519.6$ & $725.5$/$1284.8$ & $623.7$/$1007.2$ & $359.2$/$557.8$ & $291.7$/$465.6$ & $193.7$/$305.5$ & $155.2$/$251.0$ \\
& Aya-101 & $1491.3$/$3012.2$ & $1485.2$/$3180.8$ & $890.3$/$1521.8$ & $752.7$/$1213.0$ & $440.8$/$688.9$ & $352.0$/$554.4$ & $237.7$/$372.8$ & $187.3$/$298.9$ \\
& LLama3 & $1921.7$/$3822.6$ & $1834.3$/$3933.4$ & $1147.3$/$1963.9$ & $929.7$/$1512.4$ & $568.1$/$885.4$ & $434.9$/$689.2$ & $306.4$/$475.8$ & $231.5$/$373.0$ \\
& Gemma-2 & $1787.5$/$3573.5$ & $1703.0$/$3666.1$ & $1067.2$/$1834.8$ & $863.0$/$1416.2$ & $528.3$/$819.4$ & $403.6$/$637.6$ & $284.9$/$447.8$ & $214.8$/$343.9$ \\
& MordernBERT & $2073.7$/$4134.2$ & $1965.8$/$4239.3$ & $1238.1$/$2138.4$ & $996.3$/$1625.6$ & $613.0$/$956.3$ & $466.1$/$737.0$ & $330.6$/$515.8$ & $248.0$/$399.0$ \\
\bottomrule
    
  \end{tabular}
  }
  \vspace{-3mm}
  \caption{\afridoct Pseudo-document statistics. The number of translation instances in the data \afridoct pseudo-document splits.  average and 95th percentile (average/95 percentile) of the \afridoct document train split tokenization statistics using the different LLM tokenizers.}
  \vspace{-4mm}
  \label{tab:app_doc_stats}
  \end{center}
\end{table*}

\section{Experimental details}
\label{sec:appendix}

\subsection{Evaluated Models}
\label{sec:app_eval_methods}
\subsubsection{Translation Models}
M2M-100~\citep{fan2020englishcentric} is a transformer-based multilingual neural translation model from Meta, trained to translate between 100 languages, including several African languages. It has three variants of different sizes: 400M parameters, 1.2B parameters, and 12B parameters. For our experiments, we evaluated the 400M and 1.2B variants.

\paragraph{NLLB}~\citep{nllb2024scaling} is a model similar to M2M-100, with broader coverage, trained to translate between just over 200 languages, including more than 50 African languages. It also has different sizes: 600M, 1.3B, 3.3B, and 54B parameters. For this work, we evaluated the first three variants.

\paragraph{MADLAD-400}~\citep{kudugunta2023madlad400} is a multilingual translation model based on the T5 architecture \citep{raffel-etal-2020-exploring}, covering 450 languages, including many African languages. It was trained on data collected from the Common Crawl dataset. The dataset underwent a thorough self-audit to filter out noisy content and ensure its quality for training MT models.

\paragraph{Toucan}~\citep{elmadany-etal-2024-toucan, adebara-etal-2024-cheetah} is another multilingual but Afro-centric translation model based on the T5 architecture, covering 150 language pairs of African languages. It was first pre-trained on large multilingual texts covering over 500 African languages and then finetuned on translation task covering over 100 language pairs. %

\subsubsection{Large Language Models}
\paragraph{Aya-101}~\citep{ustun2024aya} is an instruction-tuned mT5 model~\citep{xue-etal-2021-mt5} designed to handle both discriminative and generative multilingual tasks. With 13B parameters, it covers 100 languages and is capable of translating between a wide range of languages, including African languages.

\paragraph{Gemma2}~\citep{gemmateam2024gemma2improvingopen}  is a decoder-only LLM trained on billions of tokens sourced from the web. The training data primarily consists of English-language text, but it also includes code and mathematical content. While Gemma2 has an English-centric focus, it also possesses multilingual capabilities. We evaluate the base Gemma2 model with 9B parameters, as well as its instruction-tuned version.

\paragraph{LLama3.1}~\citep{dubey2024llama3herdmodels} is another decoder-only LLM trained on trillions of tokens across multiple languages. It was fine-tuned using existing instruction datasets as well as synthetically generated instruction data to create its instruction-tuned version. One advantage LLama3.1 has over other models is its context window of 128K tokens, the largest among all models considered in this work, making it particularly suitable for document-based tasks such as document-level translation. We evaluate the base LLama3.1 model with 8B parameters, as well as its instruction-tuned version.

\paragraph{LLaMAX3}~\citep{lu2024llamax} is a multilingual LLM built on the LLama3 with 8B parameters as its base. It was trained on 102 languages, including several African languages, through continued pretraining. Using an English instruction dataset (Alpaca), it was further fine-tuned to create LLaMAX3-Alpaca. We evaluated both models and compared their performance across various tasks.

\subsection{Supervised Finetuning}
\label{sec:app_sft}
We perform supervised fine-tuning to tailor LLMs for translation tasks. To train sentence-level MT systems, we use all parallel sentences from \afridoct to construct the training set, enabling the LLMs to translate across multiple directions and domains. Following \citet{zhu-etal-2024-preference}, we augment the parallel data with translation instructions, which are randomly sampled from a predefined set of 31 MT instructions for each training example.\footnote{We use the same instruction set as described in \citep{zhu-etal-2024-preference}.} To train document-level MT systems, we follow the same process, but train on longer segments formed by concatenating multiple sentences. When fine-tuning, we use a learning rate of $5e^{-6}$ and an effective batch size of 64. Models are trained for only one epoch, as further training does not result in improvements and may even lead to performance degradation.

Similarly, we fine-tuned the 1.3B version of NLLB-200 for sentence and pseudo-document (with 10 sentences) translation using the Fairseq~\citep{ott2019fairseq} codebase. We used all the training examples from 30 language directions across both domains. The model was fine-tuned for 50k steps using a learning rate of $5e^{-5}$, token batch size of 2048 and a gradient accumulation of 2. The checkpoint with the lowest validation loss was selected as the best model for evaluation.

\subsection{Evaluation setup}
The models were evaluated using different tools. For example, both the NLLB-200 and M2M-100 models were evaluated with the Fairseq codebase, while Toucan and MADLAD-400 were evaluated using the Hugging Face (HF) codebase. All other LLMs, including LLama3.1 (both instruction-tuned and SFT models), Gemma, and Aya-101, were evaluated using EleutherAI LM Evaluation Harness (\texttt{lm-eval}) tool~\citep{biderman2024lessons}. In all cases, greedy decoding was used.

The models evaluated have different context lengths. For encoder-decoder models, M2M-100 and NLLB have a maximum sequence length of 1024 and 512 respectively. Aya-101 and MADALAD, based on the T5 architecture, do not have a pre-specified maximum sequence length, so we fixed their maximum sequence length to 1024 for all experiments involving encoder-decoder models. However, for decoder-only models, Gemma and LLaMAX3 (based on LLama3) have a maximum sequence length of 8192, while LLama3.1 has a maximum sequence length of 128K. Since all the decoder-only models were evaluated using LM Evaluation Harness, we used a similar setup for them, selecting the maximum length based on the specific needs of each model.

\begin{table}[t]
    \centering\small
    \scalebox{1}{ %
    \begin{tabular}{lcc}
        \toprule
        \textbf{Setting} & \textbf{X $\rightarrow$ eng} & \textbf{eng $\rightarrow$ X} \\
        \midrule
    
    \multicolumn{3}{l}{\textbf{Sentence}} \\
    sentence & $512$ & $512$ \\
    \midrule
    \multicolumn{3}{l}{\textbf{Document}} \\
    5 & $4096$ & $4096$ \\
    10 & $4096$ & $4096$ \\
    25 & $1024$ & $8192$ ($11264$) \\
    Full & $2048$ & $16384$ ($32768$) \\
        \bottomrule
    \end{tabular}
   }
    \caption{The maximum number of tokens set for decoder-only LLMs when translating between English and African languages, and vice versa. Special cases for Amharic are indicated in brackets.}
    \label{tab:gener_toks}
\end{table}
 
\Cref{tab:gener_toks} shows the maximum number of generation tokens we set when translating between English and African languages. These numbers were chosen based on the statistics from \Cref{tab:app_doc_stats}. However, for Amharic, when translating pseudo-documents with 25 sentences and full documents, there were instances exceeding the 95th percentile derived from the training statistics. Therefore, we increased the token limit specifically for Amharic.

\subsection{Evaluation prompts}
\label{sec:app_eval_prompts}
While the translation models we evaluated do not require prompts, MADLAD-400, requires a prefix of the form \textbf{<2xx>} token, which is prepended to the source sentence. Here, xx indicates the target language using its language code (e.g., ``sw'' for Swahili). Similarly, Toucan uses just the target language ISO-693 code as prefix, which is prepended to the source sentence (e.g., ``swa'' for Swahili). For other models, including Aya-101, we used three different prompts for sentence-level translation and document translation experiments. The main difference between the prompts for these tasks is the explicit mention of ``text'' or ``document'' within the prompt, as shown in \Cref{tab:prompt_examples}. For the base models Gemma2, Llama3.1, LLaMAX3, and Aya-101, we prompted them directly using the respective prompts. However, for the instruction-tuned versions of Gemma2 and Llama3.1, we used their respective chat templates. For all Alpaca-based models, including our SFT models, we used the Alpaca template.

\subsection{Evaluation metrics}
We evaluate translation quality with BLEU~\citep{papineni-etal-2002-bleu} and ChrF~\citep{popovic-2015-CHRF} using SacreBLEU\footnote{\texttt{case:mixed|eff:no|} \texttt{tok:13a|smooth:exp|v:2.3.1}, }~\citep{post-2018-call}. We run significance tests using bootstrap resampling and report the $95\%$ confidence interval for the scores, based on a sample size of $1000$. We also use AfriCOMET\footnote{\url{https://huggingface.co/masakhane/africomet-stl-1.1}}~\citep{wang2024afrimte} to evaluate the quality of the translation outputs. We report the chrF scores of the best prompt for each model and language direction in the main paper, with all additional results provided in the \Cref{sec:app_extra_result}. For document-level experiments, we evaluated the LLMs using the same three prompts as in the sentence-level experiment. For evaluation, we used BLEU and chrF scores but excluded AfriCOMET due to its backbone model, AfroXLM-R-L~\citep{alabi-etal-2022-adapting, adelani-etal-2024-sib}, having a context length of 512 tokens. This made it impractical to compute COMET scores for document-level outputs. 

\subsection{GPT-4o as an evaluator for machine translation}
\label{sec:llm_judge}

We use GPT-4o to assess the quality of translation output, as demonstrated by~\citet{sun-etal-2025-fine}, which shows a correlation with human judgment. Due to the cost of this task, we limited our evaluation to a few selected models, including Aya-101, GPT-3.5, GPT-4o, and LLaMAX3 fine-tuned on \afridoct sentences and pseudo-documents of 10 sentences. We compared translations performed at the sentence-level and pseudo-document level in terms of fluency, content errors, and cohesion errors—specifically lexical (LE) and grammatical (GE) errors—using the same definitions as~\citet{sun-etal-2025-fine}.

Below are the prompts used to evaluate documents using GPT-4o for fluency, content errors, and cohesion errors—specifically lexical (LE) and grammatical (GE) errors.

\begin{itemize}
  \item \textbf{Fluency}: GPT-4o is prompted to rate the fluency of a document on a scale from 1 to 5, where 5 indicates high fluency and 1 represents low fluency. This evaluation is conducted without providing any reference document. For the final fluency score, we report the average rating across all documents. Below we provide the prompt used. 
    \begin{lstlisting}
Please evaluate the fluency of the following text in <<target>>.
    
------
    
### **Instructions:**

- **Task**: Evaluate the fluency of the text.

- Scoring: Provide a score from 1 to 5, where:

  - **5**: The text is **highly fluent**, with no grammatical errors, unnatural wording, or stiff syntax.
  - **4**: The text is **mostly fluent**, with minor errors that do not impede understanding.
  - **3**: The text is **moderately fluent**, with noticeable errors that may slightly affect comprehension.
  - **2**: The text has **low fluency**, with frequent errors that hinder understanding.
  - **1**: The text is **not fluent**, with severe errors that make it difficult to understand.
- **Explanation**: Support your score with specific examples to justify your evaluation.

------

### **Output Format:**

Provide your evaluation in the following JSON format:

```
{
  "Fluency": {
    "Score": "<the score>",
    "Explanation": "<your explanation on how you made the decision>"
  }
}
```

------

**Text to Evaluate:**

<<hypothesis>>

Answer:
    \end{lstlisting}

  \item \textbf{Accuracy}: GPT-4 is prompted to identify and list the mistakes, such as incorrect translations, omissions, additions, and any other errors, by comparing the model's output to the reference translation. After identifying these errors, we count all of them and compute the average across all documents, reporting that as the content error (CE). Below is the prompt used.
    \begin{lstlisting}
Please evaluate the accuracy of the following translated text in <<target>> by comparing it to the provided reference text.

------

### **Instructions:**

- **Task**: Compare the text to the reference text.

- Identify Mistakes: List all mistakes related to accuracy.

  - Mistake Types:

    - **Wrong Translation**: Incorrect meaning or misinterpretation leading to wrong information.
    - **Omission**: Missing words, phrases, or information present in the reference text.
    - **Addition**: Extra words, phrases, or information not present in the reference text.
    - **Others**: Mistakes that are hard to define or categorize.

- **Note**: If the text expresses the same information as the reference text but uses different words or phrasing, it is **not** considered a mistake.

- **Provide a List**: Summarize all mistakes without repeating the exact sentences. Provide an empty list if there are no mistakes.

------

### **Output Format:**

Provide your evaluation in the following JSON format:

```
{
  "Accuracy": {
    "Mistakes": [
      "<list of all mistakes in the text with format'Mistake Types: summarize the mistake', provide an empty list if there are no mistakes>"
    ]
  }
}
```

------

**Reference Text:**

<<reference>>

**Text to Evaluate:**

<<hypothesis>>
    \end{lstlisting}

  \item \textbf{Cohesion}: GPT-4 is prompted to rate cohesion-related mistakes, including lexical and grammatical errors, in the model's output, comparing it to the reference translation. We count each error individually, compute the average across the documents, and report them as lexical errors (LE) and grammatical rrrors (GE). Below is the prompt template we used. 
    \begin{lstlisting}
 Please evaluate the cohesion of the following translated text in <<target>> by comparing it to the provided reference text.

------

### **Instructions:**

- **Task**: Evaluate the cohesion of the text.

- **Definition**: Cohesion refers to how different parts of a text are connected using language structures like grammar and vocabulary. It ensures that sentences flow smoothly and the text makes sense as a whole.

- Identify Mistakes: List all mistakes related to cohesion.

  - Separate the mistakes into:

    - **Lexical Cohesion Mistakes**: Issues with vocabulary usage, incorrect or missing synonyms, or overuse of certain words that disrupt the flow.
    - **Grammatical Cohesion Mistakes**: Problems with pronouns, conjunctions, or grammatical structures that link sentences and clauses.

- **Provide Lists**: Provide separate lists for lexical cohesion mistakes and grammatical cohesion mistakes. Provide empty lists if there are no mistakes.

------

### **Output Format:**

Provide your evaluation in the following JSON format:

```
{
  "Cohesion": {
    "Lexical Cohesion Mistakes": [
      "<list of all mistakes in the text one by one, provide an empty list if there are no mistakes>"
    ],
    "Grammatical Cohesion Mistakes": [
      "<list of all mistakes in the text one by one, provide an empty list if there are no mistakes>"
    ]
  }
}
```

------

**Reference Text:**

<<reference>>

**Text to Evaluate:**

<<hypothesis>>   
\end{lstlisting}
\end{itemize}

Fluency can only have values between 1 and 5. However, the other metrics, including CE, GE, and LE, do not have a specific range and can take on any value because they are counts. Refer to~\citep{sun-etal-2025-fine} for more details about these metrics.

\subsection{Human Evaluation Setup}
\label{app:human_eval}
Beyond using GPT-4o as a judge, we also conduct human evaluation on a subset of outputs from GPT-3.5, LLaMAX3-SFT$_{1}$, and LLaMAX3-SFT$_{10}$ for two domains, focusing specifically on translation into five African languages due to cost constraints. Translation into English was excluded, as existing automatic metrics, including GPT-based evaluations, are already reliable for this direction.

For the human evaluation, three native speakers of the African languages—primarily translators involved in the dataset creation—were recruited. Each annotator was assigned 80 documents to evaluate, tasked with marking as many error spans as possible and rating the overall quality on a scale from 0 to 100. This annotation followed the error span annotation (ESA)~\citep{kocmi-etal-2024-error} protocol as implemented within the Appraise Evaluation Framework \citep{federmann-2018-appraise}. To assess consistency and inter-annotator agreement, 30 of the 80 documents were shared among all three annotators. \Cref{tab:app_human_eval_stat} shows statistics for 80 documents sampled from the models in both domains for each annotator. Each annotator was remunerated with \$55.\footnote{\href{https://github.com/masakhane-io/afridoc-mt/tree/main}{Annotation protocol.}}

\begin{table}[t]
    \centering\small
    \scalebox{0.85}{ %
        \begin{tabular}{lrrrrr}
        \toprule
        \multirow{2}{*}{\textbf{Model}} & \multirow{2}{*}{\textbf{Setup}} & \multicolumn{2}{c}{\textbf{Full}} & \multicolumn{2}{c}{\textbf{Shared}} \\
        \cmidrule(lr){3-4} \cmidrule(lr){5-6}
         &  & \textbf{health} & \textbf{tech} & \textbf{health} & \textbf{tech} \\
        \midrule
        \multirow{2}{*}{GPT-3.5} &Sent. &5 &5 &- &5 \\
        &Pseudo. &5 &5 &- &5 \\
        \multirow{2}{*}{LLaMAX3-SFT$_{1}$} &Sent. &5 &5 &5 &- \\
        &Pseudo. &5 &5 &5 &- \\
        LLaMAX3-SFT$_{10}$ &Pseudo &5 &5 &5 &5 \\
        \midrule
        Total & &25 &25 &15 &15 \\
        \bottomrule
        \end{tabular}   
   }
    \caption{The number of documents annotated by each annotator for human direct assessment.}
    \vspace{-3mm}
    \label{tab:app_human_eval_stat}
\end{table}

\subsection{Qualitative Analysis}
\label{app:qualitat_eval}
Alongside the human direct assessment of the translation outputs, we shared a subset of the outputs with one author per language, each a native speaker. They were tasked with analyzing the outputs to answer two key questions: (1) What common errors or flaws do the models exhibit across different setups? and (2) How fluent are the translation outputs produced by the models across the various settings?

\begin{figure}[t]
\setlength{\belowcaptionskip}{-2pt}
\setlength{\abovecaptionskip}{-1pt}   %
  \centering
    \begin{subfigure}{0.48\columnwidth}
        \includegraphics[width=\textwidth]{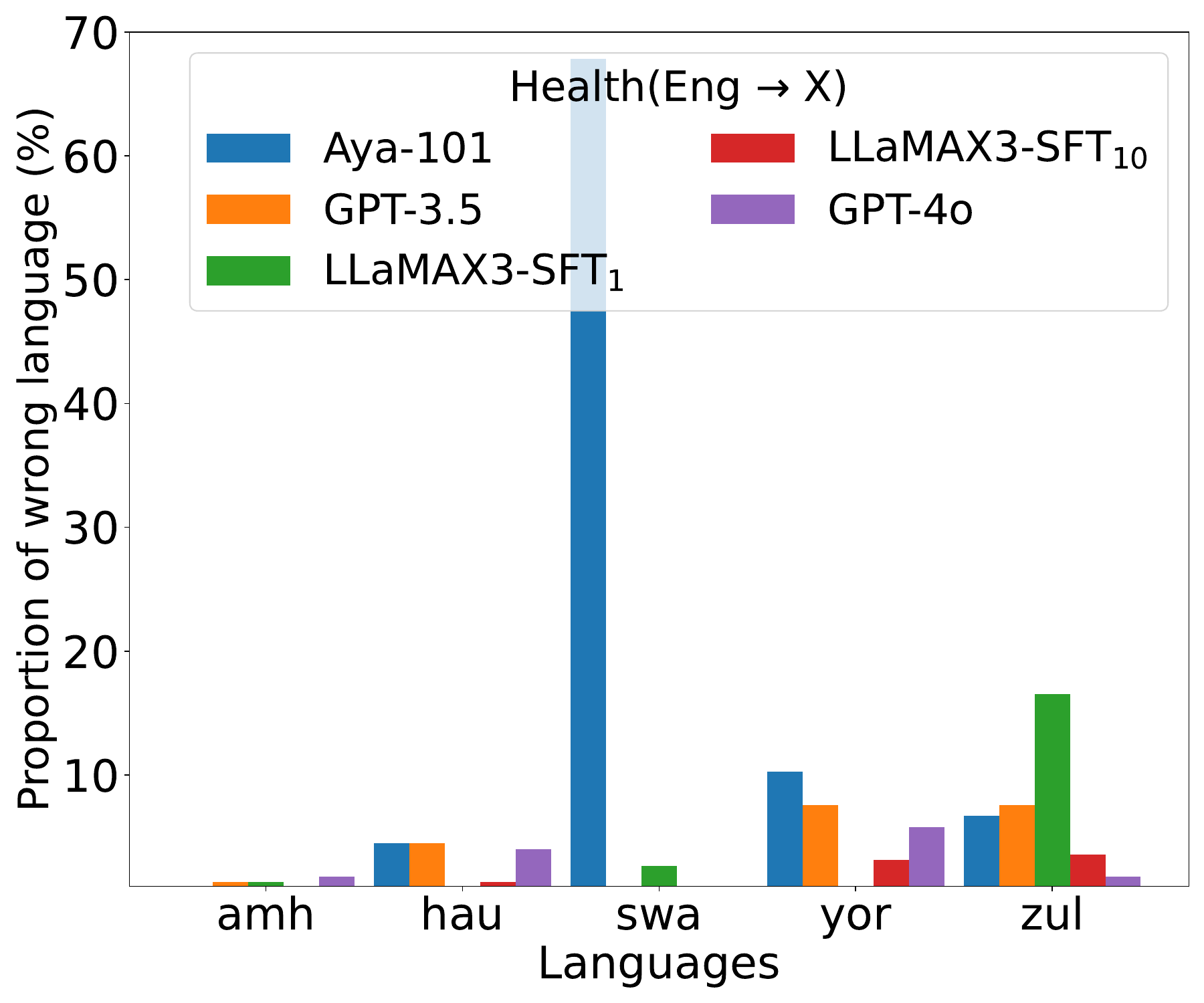}
    \end{subfigure}
    ~
    \begin{subfigure}{0.48\columnwidth}
        \includegraphics[width=\textwidth]{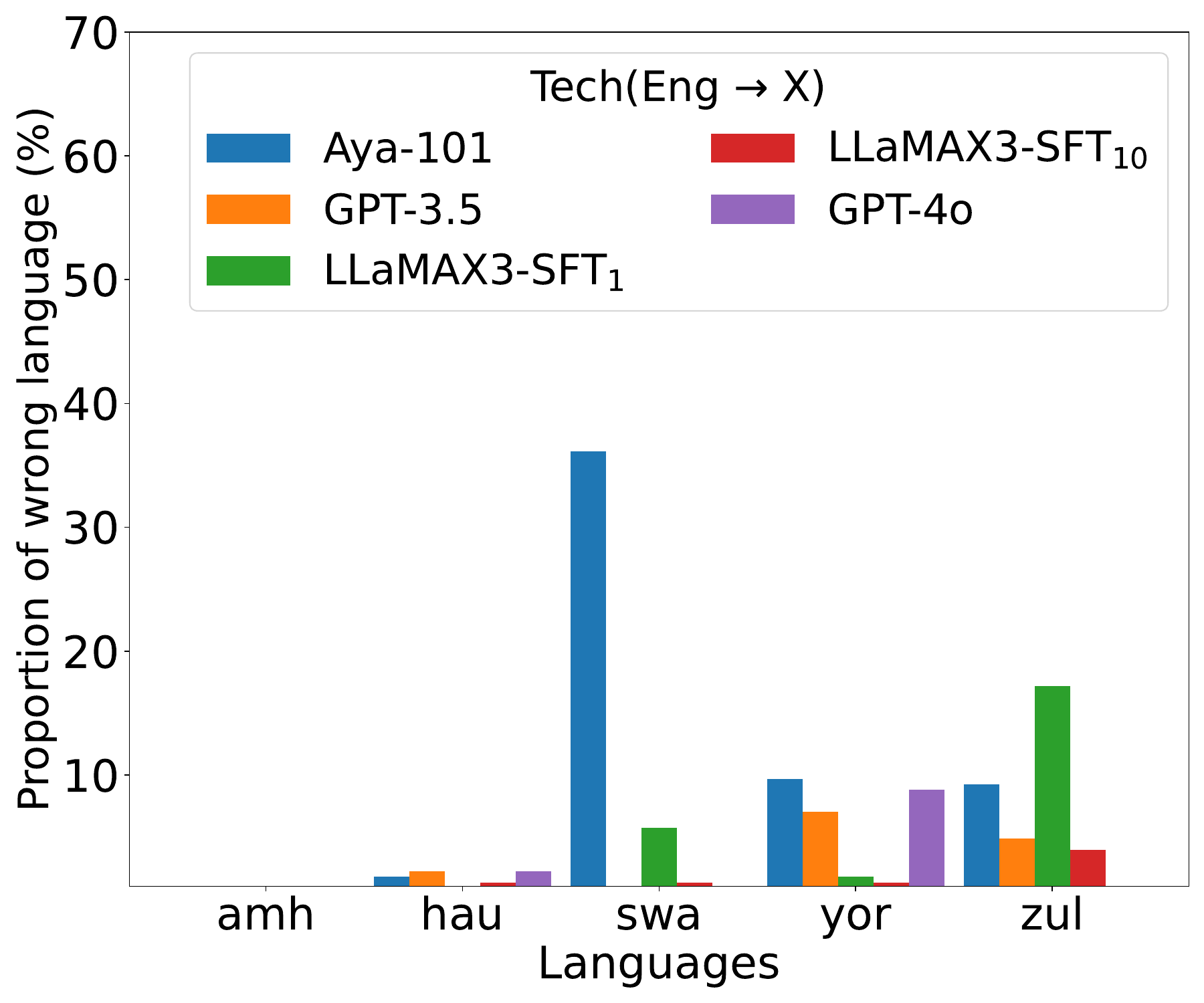}
    \end{subfigure}
\caption{Rate of off-target translation ($k=$10).}
\label{fig:app_mistranslation2}
\end{figure} 

\begin{figure}[h]
\setlength{\belowcaptionskip}{-10pt}
\setlength{\abovecaptionskip}{-1pt}   %
  \centering
    \begin{subfigure}{0.48\columnwidth}
        \includegraphics[width=\textwidth]{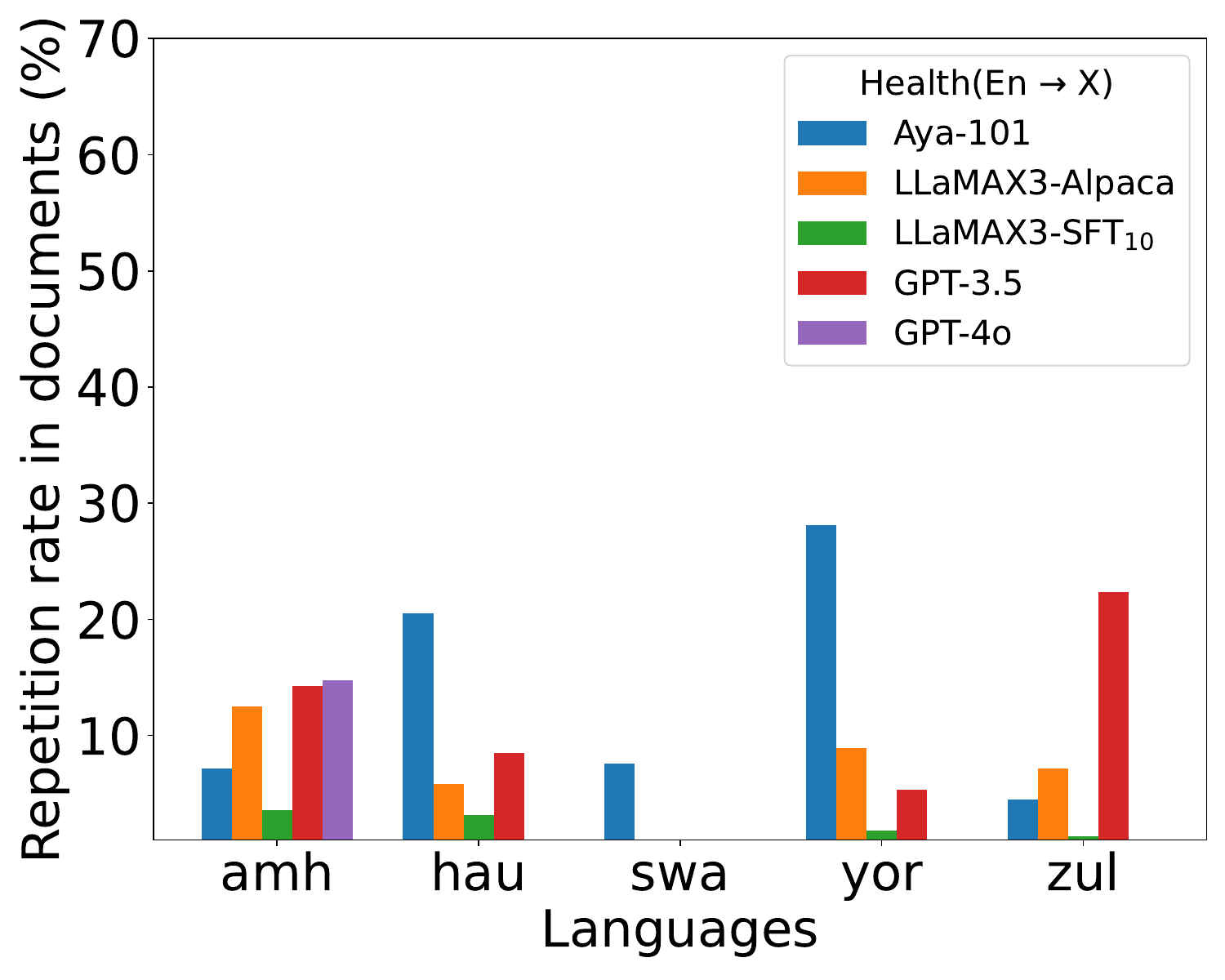}
    \end{subfigure}
    ~
    \begin{subfigure}{0.48\columnwidth}
        \includegraphics[width=\textwidth]{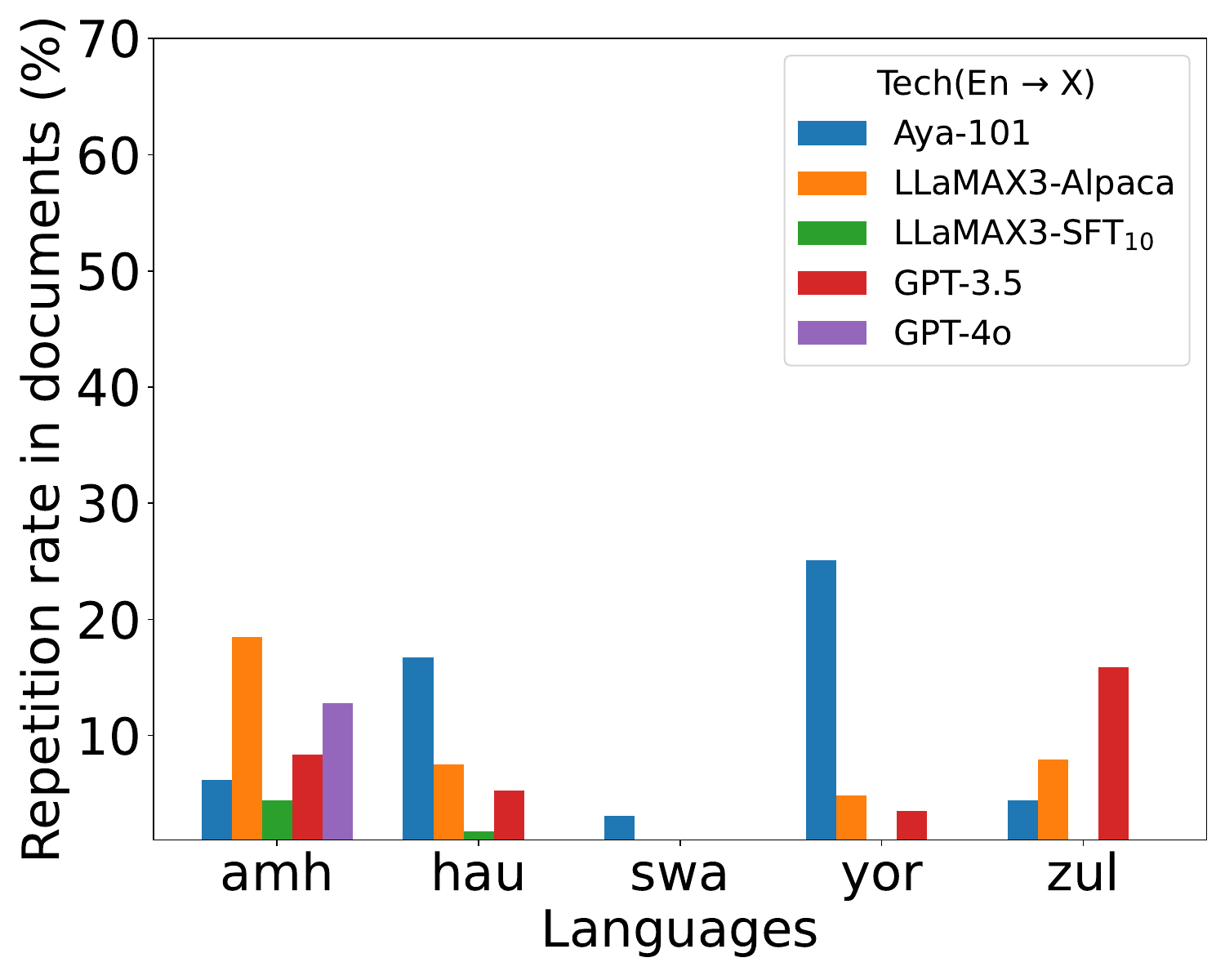}
    \end{subfigure}
\caption{Word repetition rate in the pseudo-document translation ($k=$10).}
\label{fig:app_repition_doc2}
\end{figure}

\begin{figure}[t]
\setlength{\belowcaptionskip}{-2pt}
\setlength{\abovecaptionskip}{-1pt}   %
  \centering
    \begin{subfigure}{0.48\columnwidth}
        \includegraphics[width=\textwidth]{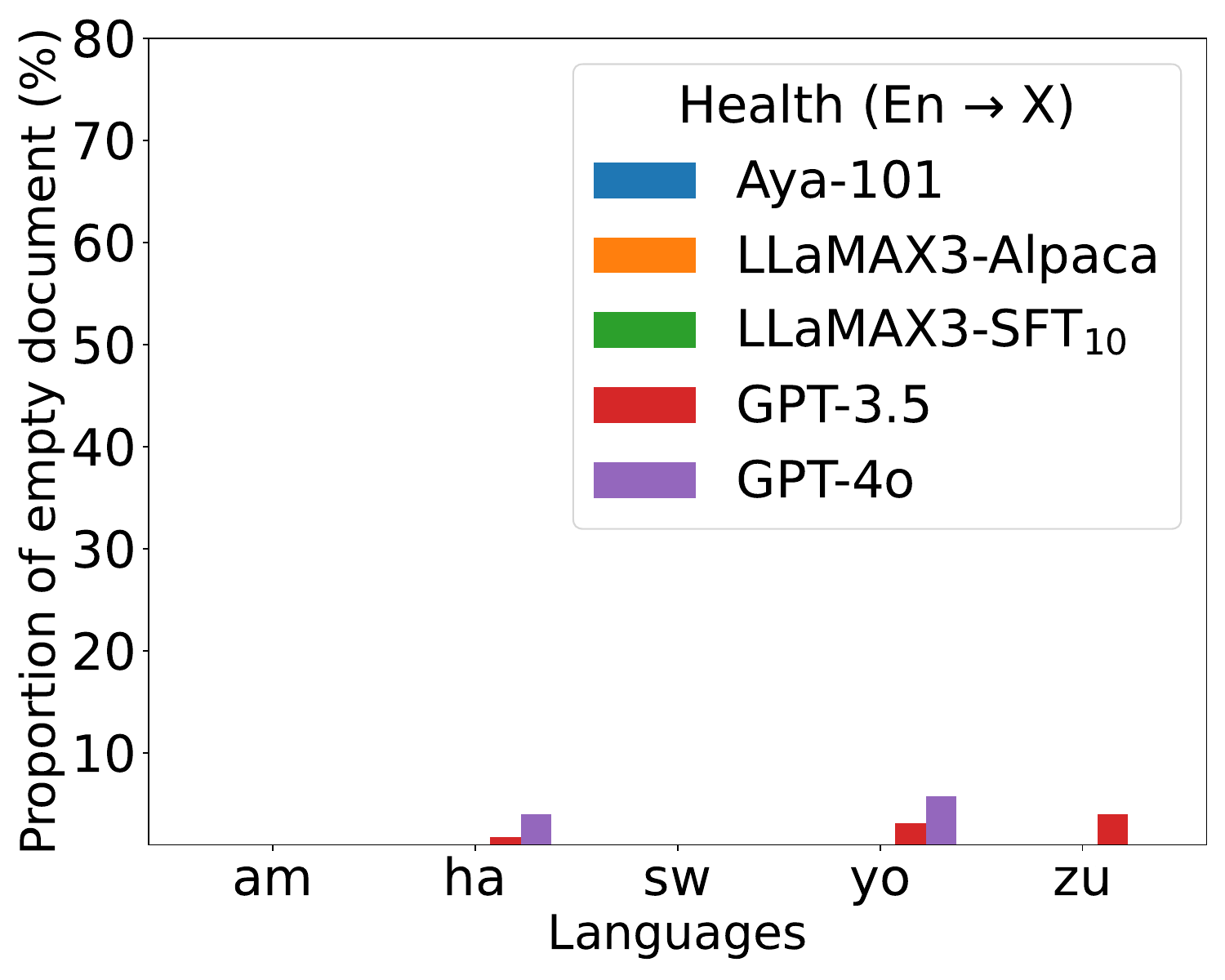}
    \end{subfigure}
    ~
    \begin{subfigure}{0.48\columnwidth}
        \includegraphics[width=\textwidth]{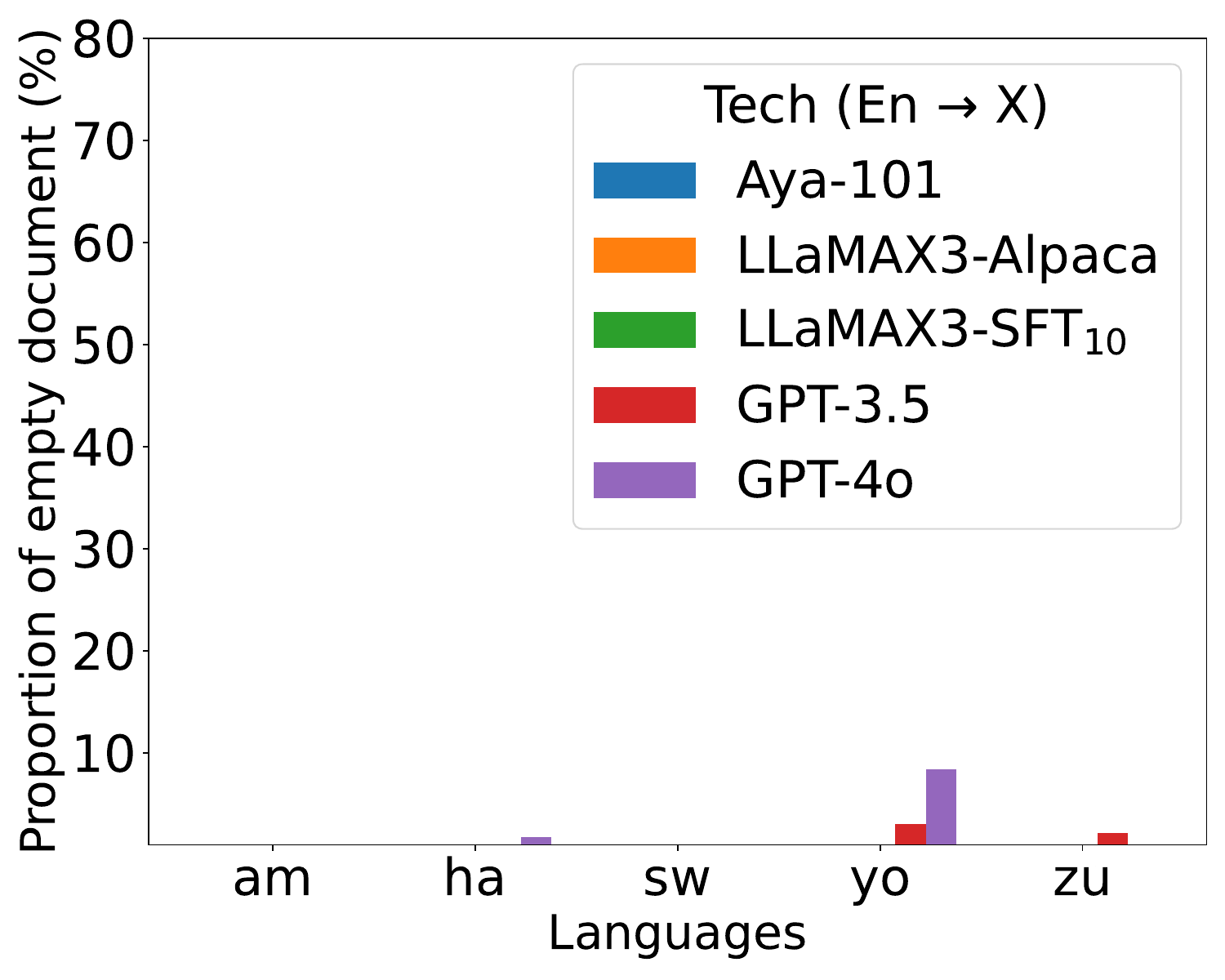}
    \end{subfigure}
\caption{Proportion of empty outputs for pseudo-documents. }
\label{fig:app_empty_doc}
\end{figure} 

\begin{figure*}[t]
\setlength{\belowcaptionskip}{-10pt}
\setlength{\abovecaptionskip}{-1pt}   %
  \centering
    \begin{subfigure}{0.48\columnwidth}
        \includegraphics[width=\textwidth]{images/length_analysis/under_gen_doc10_Health_propen2.pdf}
    \end{subfigure}
    ~
    \begin{subfigure}{0.48\columnwidth}
        \includegraphics[width=\textwidth]{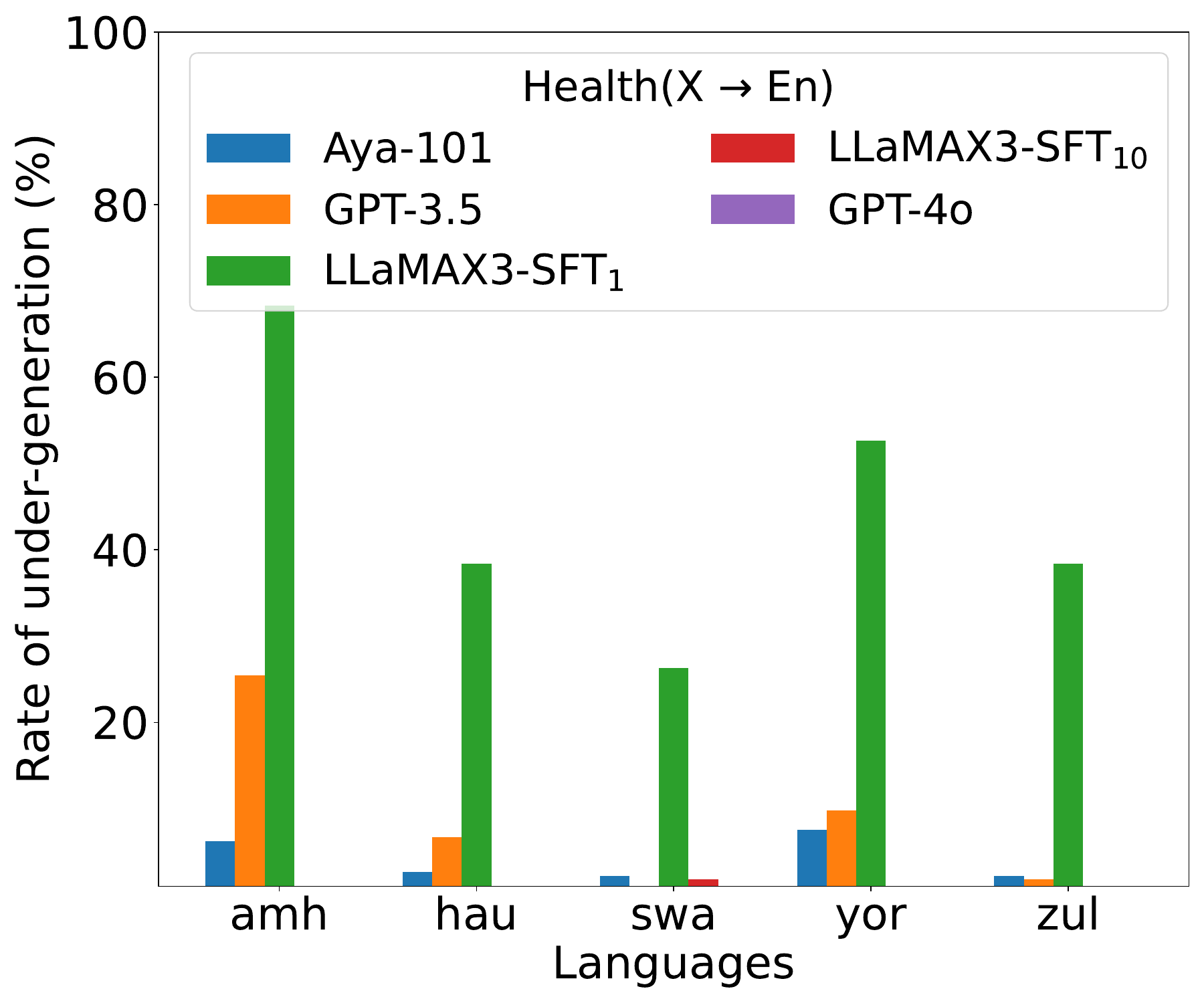}
    \end{subfigure}
    ~
    \begin{subfigure}{0.48\columnwidth}
        \includegraphics[width=\textwidth]{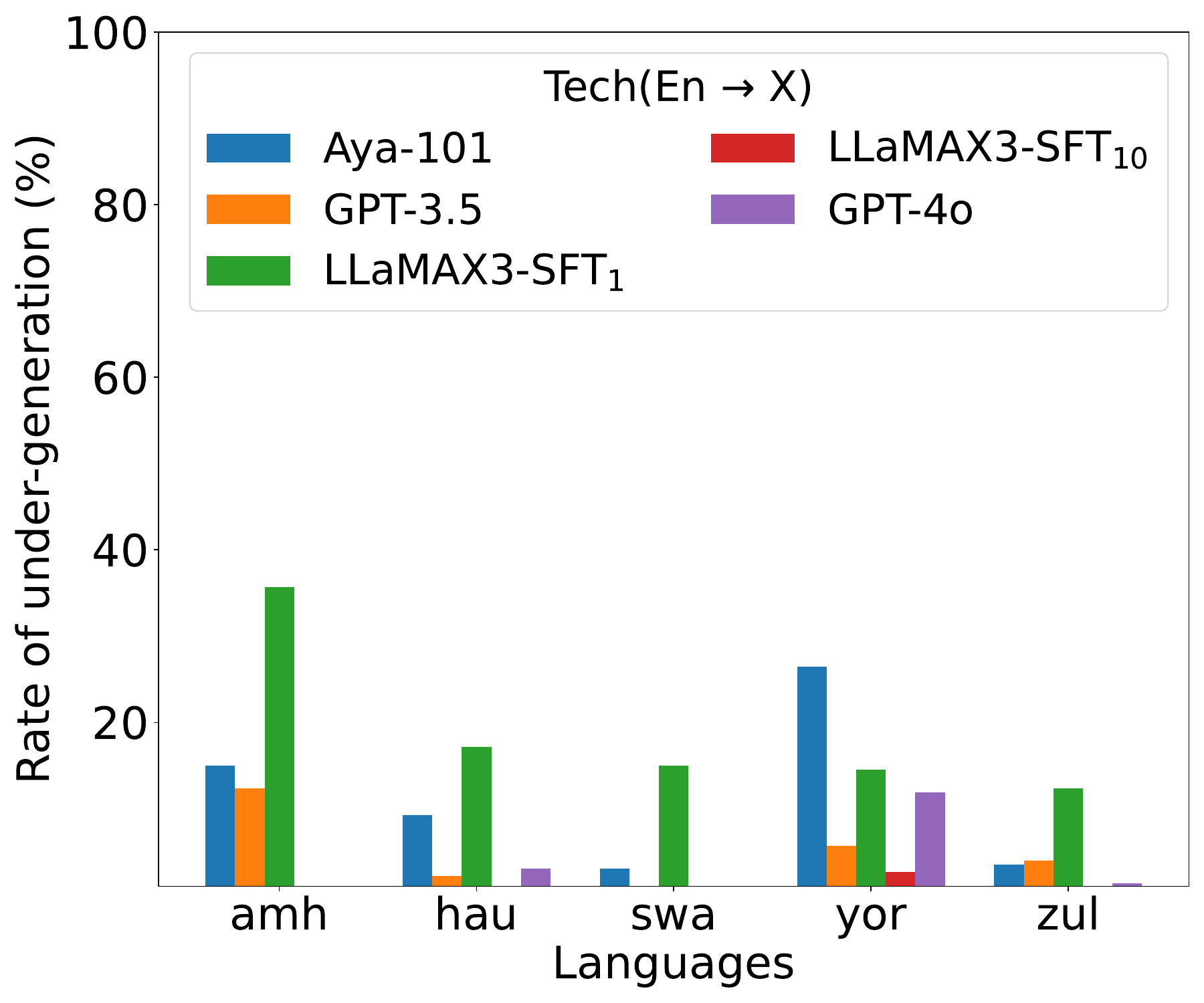}
    \end{subfigure}
    ~
    \begin{subfigure}{0.48\columnwidth}
        \includegraphics[width=\textwidth]{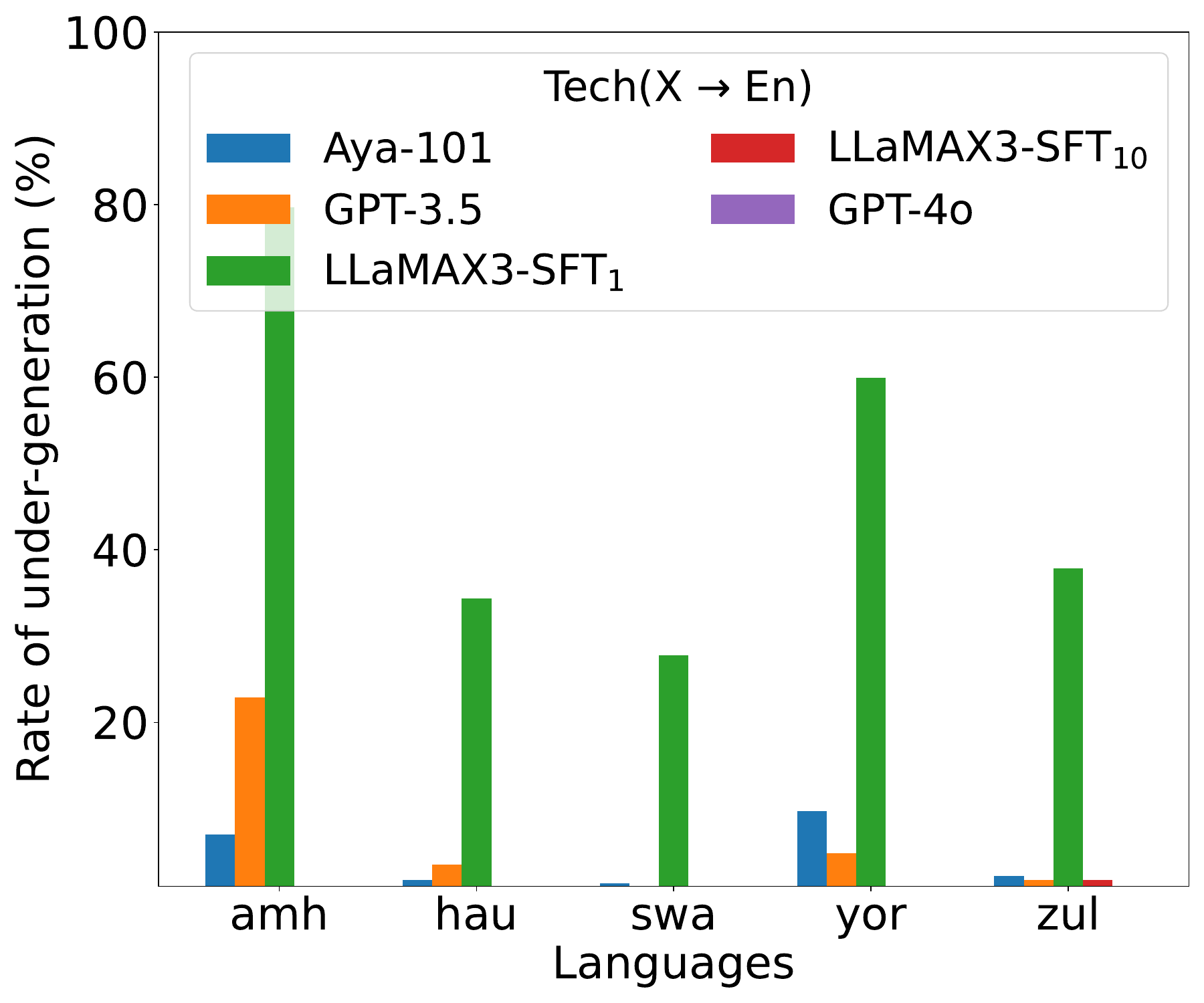}
    \end{subfigure}
\caption{Rate of under-generation in pseudo-document translation ($k=$10)}
\label{fig:app_undergeneration1}
\end{figure*}
\section{More experimental results}
\label{sec:app_extra_result}
\subsection{Sentence-level evaluation}
\begin{figure*}[t]
    \centering
    \includegraphics[width=\textwidth]{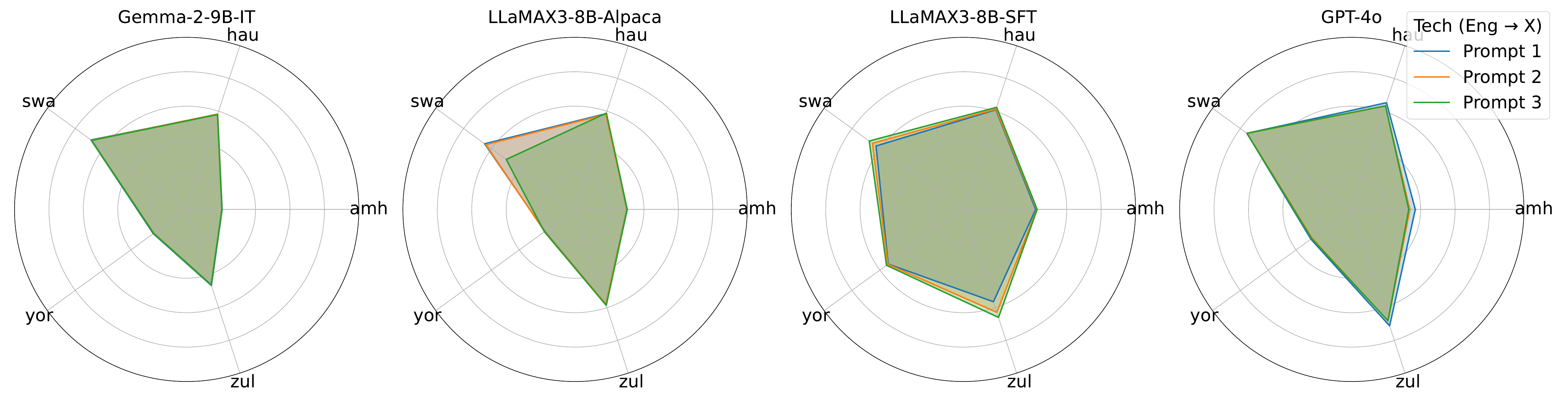}
    \caption{d-chrF scores for some LLMs for sentence-level translation using different prompts when translating into African languages}
    \label{fig:app_tech_afro}
\end{figure*}

\begin{figure*}[t]
    \centering
    \includegraphics[width=\textwidth]{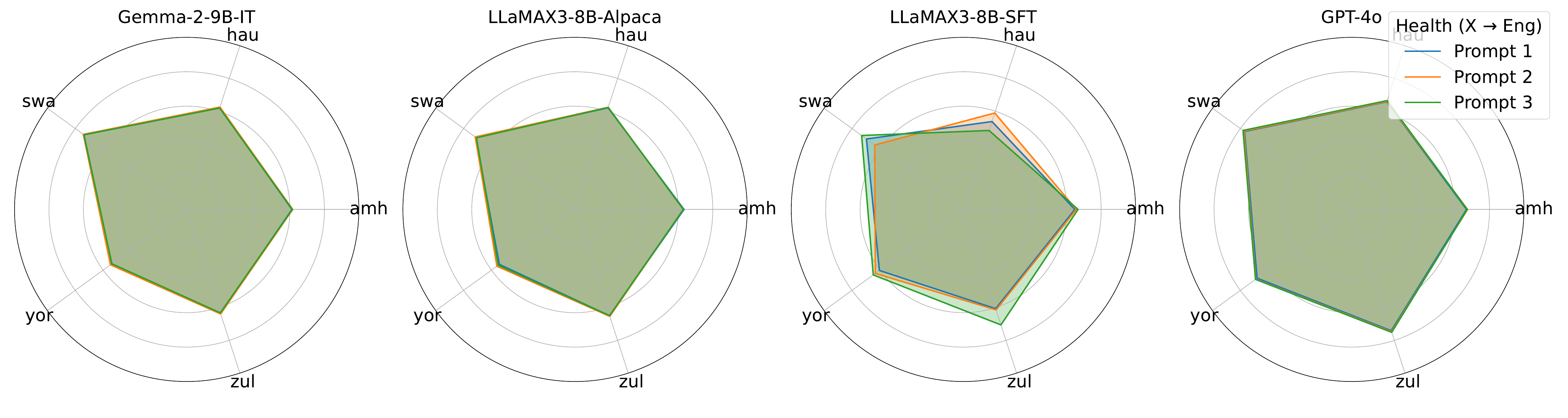}
    \caption{d-chrF scores for some LLMs for sentence-level translation using different prompts when translating into English}
    \label{fig:app_health_eng}
\end{figure*}

\begin{figure*}[t]
    \centering
    \includegraphics[width=\textwidth]{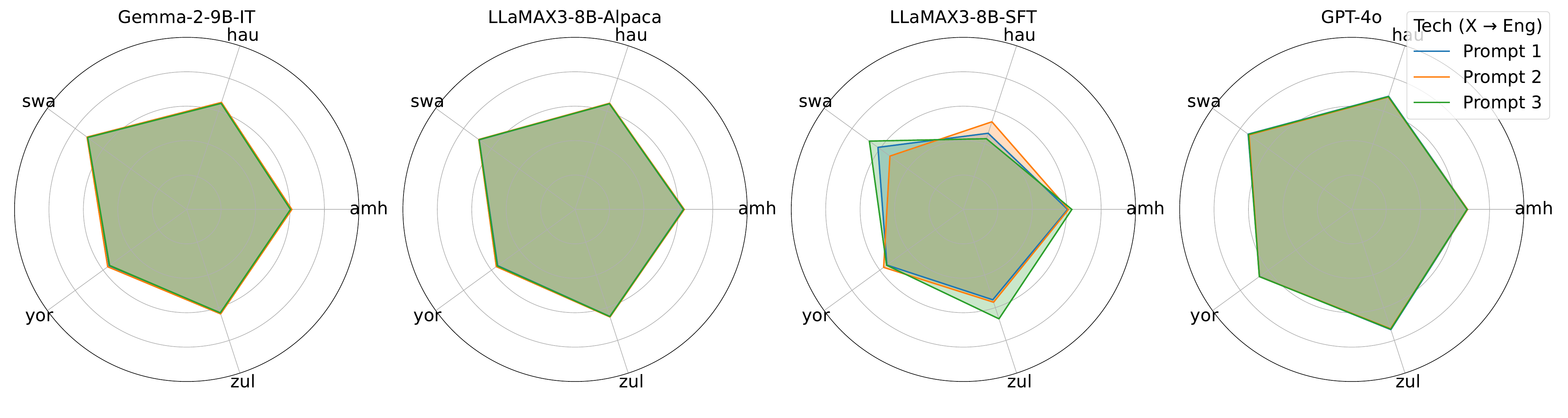}
    \caption{d-chrF scores for some LLMs for sentence-level translation using different prompts when translating into English}
    \label{fig:app_tech_eng}
\end{figure*}

\begin{figure*}[t]
    \centering
    \includegraphics[width=\textwidth]{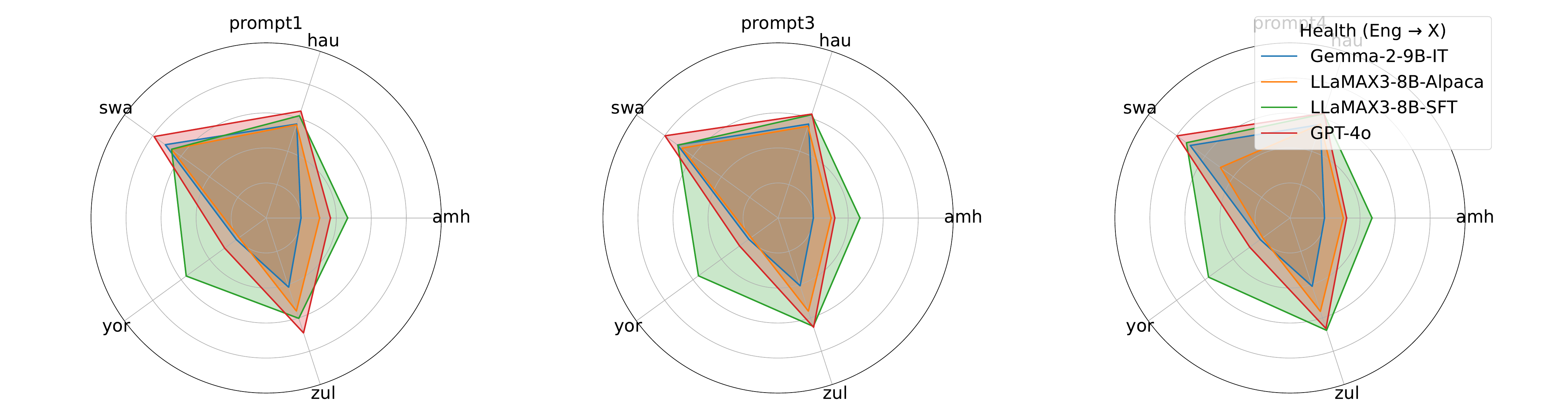}
    \caption{d-chrF scores for some LLMs for sentence-level translation using different prompts when translating into African languages}
    \label{fig:health_prompt_afro}
\end{figure*}

Given that \afridoct is a document-level translation dataset, and due to the limited context length of most translation models and LLMs, which makes it impossible to translate a full document at once, we opted to translate the sentences within the documents and then merge them back to form the complete document. This also serves as a baseline for document-level translation. In the main paper, we present the results for the best prompt for each language pair and model using d-chrF. In this section, we also provide the full results on the merged documents using d-chrF and d-BLEU in \Cref{tab:app_full_sentdoc_health,tab:app_full_sentdoc_tech}. Furthermore, we present results for evaluating just the sentences (without merging them back into documents) using s-BLEU, s-chrF, and s-COMET in \Cref{tab:app_full_sent_health,tab:app_full_sent_tech}. In \Cref{fig:app_tech_afro,fig:app_health_eng,fig:app_tech_eng,fig:health_prompt_afro}, we provide plots that summarize some of the results in the table for a few models. Although the main findings are summarized in the main draft, below are some other points we identify.

\paragraph{M2M-100 is not competitive} Neither version of M2M-100, which was once a state-of-the-art translation model, is competitive with other translation models such as Toucan, NLLB-200, and MADLAD-400, even when compared to models of similar sizes, across all metrics at both the sentence and document levels.

\paragraph{Base LLMs are not translators for African languages.} Base LLMs without instruction tuning and supervised fine-tuning, such as Gemma2 and LLaMAX3,  do not show competitive translation performance either. This can be explained by the fact that they are just language models with limited coverage of African languages. However, LLaMAX3, which was trained on more than 100 languages, including African languages, through continued pre-training, shows improved performance, surpassing LLama3.1-IT.

\paragraph{Amharic and \yoruba are the worst performing language directions.} When translating from English into African languages, our results show that both Amharic and Yoruba perform the least effectively. This may be attributed to specific properties of these languages, such as the use of non-Latin script in Amharic and the use of diacritics in Yoruba, which in turn increase the tokenization rate of these languages by the different model tokenizers.

\begin{figure*}[t]
\setlength{\belowcaptionskip}{-10pt}
\setlength{\abovecaptionskip}{-1pt}   %
  \centering
    \begin{subfigure}{0.48\columnwidth}
        \includegraphics[width=\textwidth]{images/length_analysis/over_gen_doc10_Health_propen2.pdf}
    \end{subfigure}
    ~
    \begin{subfigure}{0.48\columnwidth}
        \includegraphics[width=\textwidth]{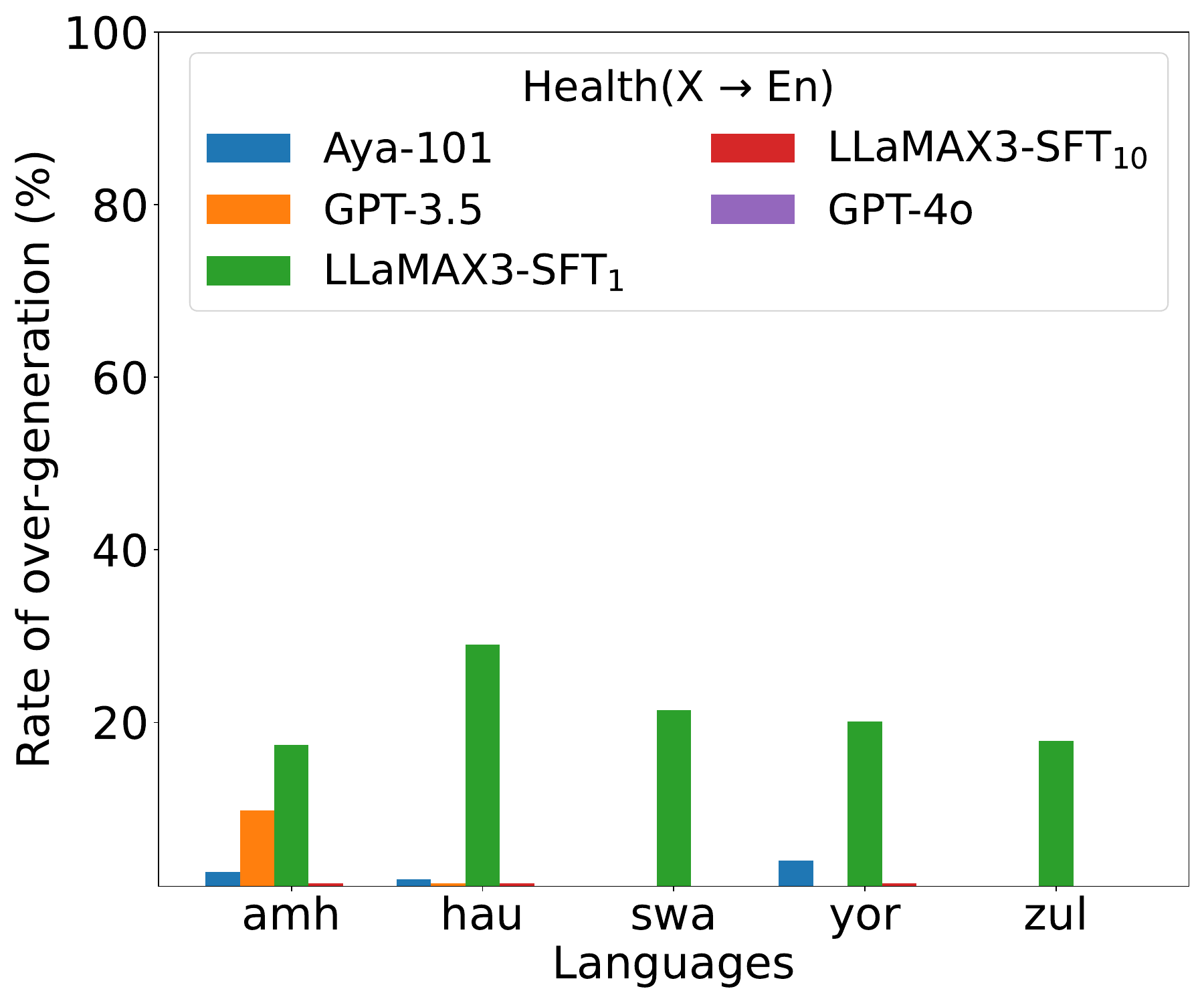}
    \end{subfigure}
    ~
    \begin{subfigure}{0.48\columnwidth}
        \includegraphics[width=\textwidth]{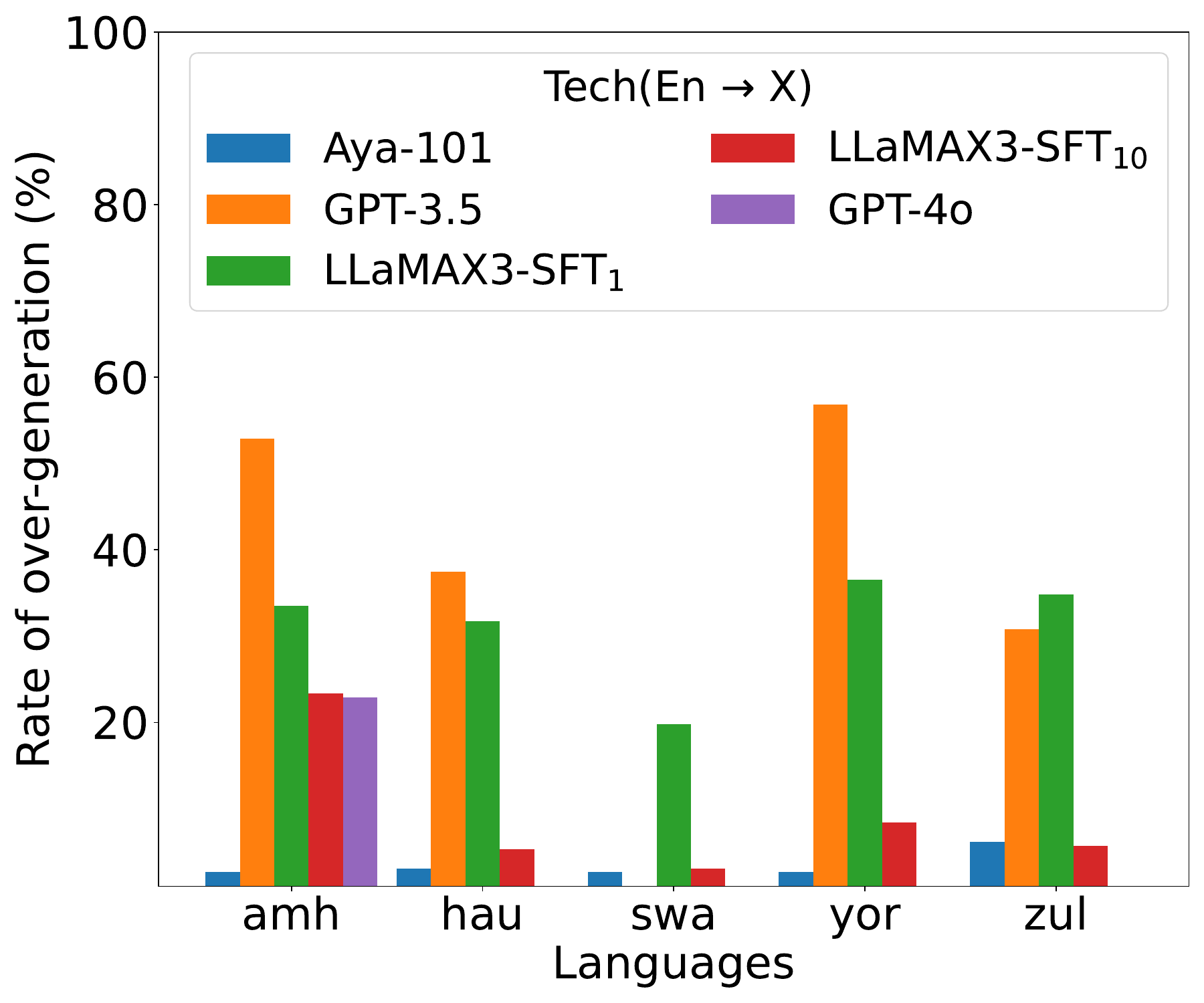}
    \end{subfigure}
    ~
    \begin{subfigure}{0.48\columnwidth}
        \includegraphics[width=\textwidth]{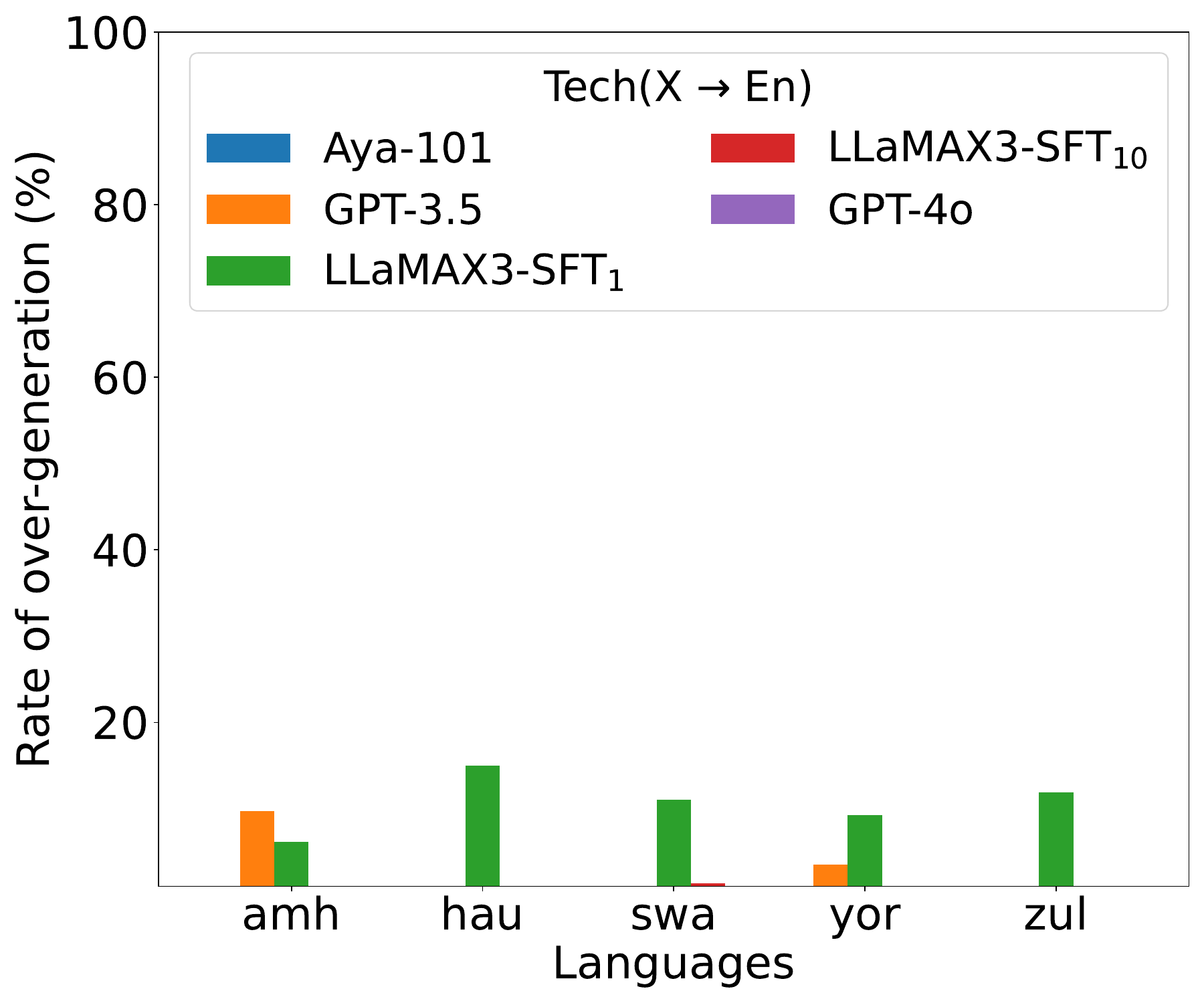}
    \end{subfigure}
\caption{Rate of over-generation in pseudo-document translation ($k=$10)}
\label{fig:app_ovrgeneration1}
\end{figure*}

\begin{figure*}[t]
\setlength{\belowcaptionskip}{-4pt}
  \centering
    \begin{subfigure}{0.235\textwidth}
        \includegraphics[width=\textwidth]{images/document_size/doc_size_health2x.pdf}
    \end{subfigure}
     ~
    \begin{subfigure}{0.235\textwidth}
        \includegraphics[width=\textwidth]{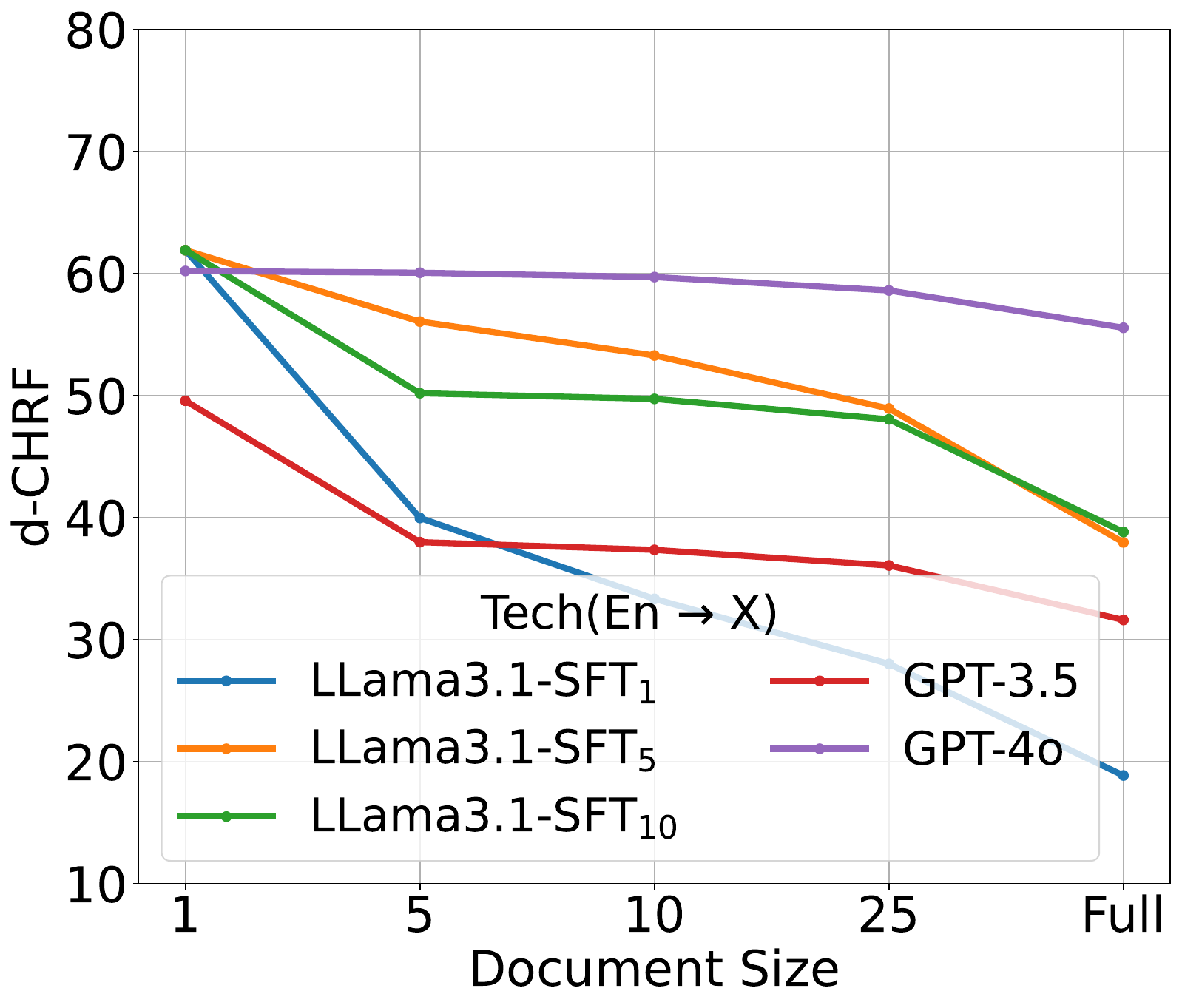}
    \end{subfigure}
     ~     
     \begin{subfigure}{0.235\textwidth}
         \includegraphics[width=\textwidth]{images/document_size/doc_size_health2en.pdf}
     \end{subfigure}
     ~
    \begin{subfigure} {0.235\textwidth} 
        \includegraphics[width=\textwidth]{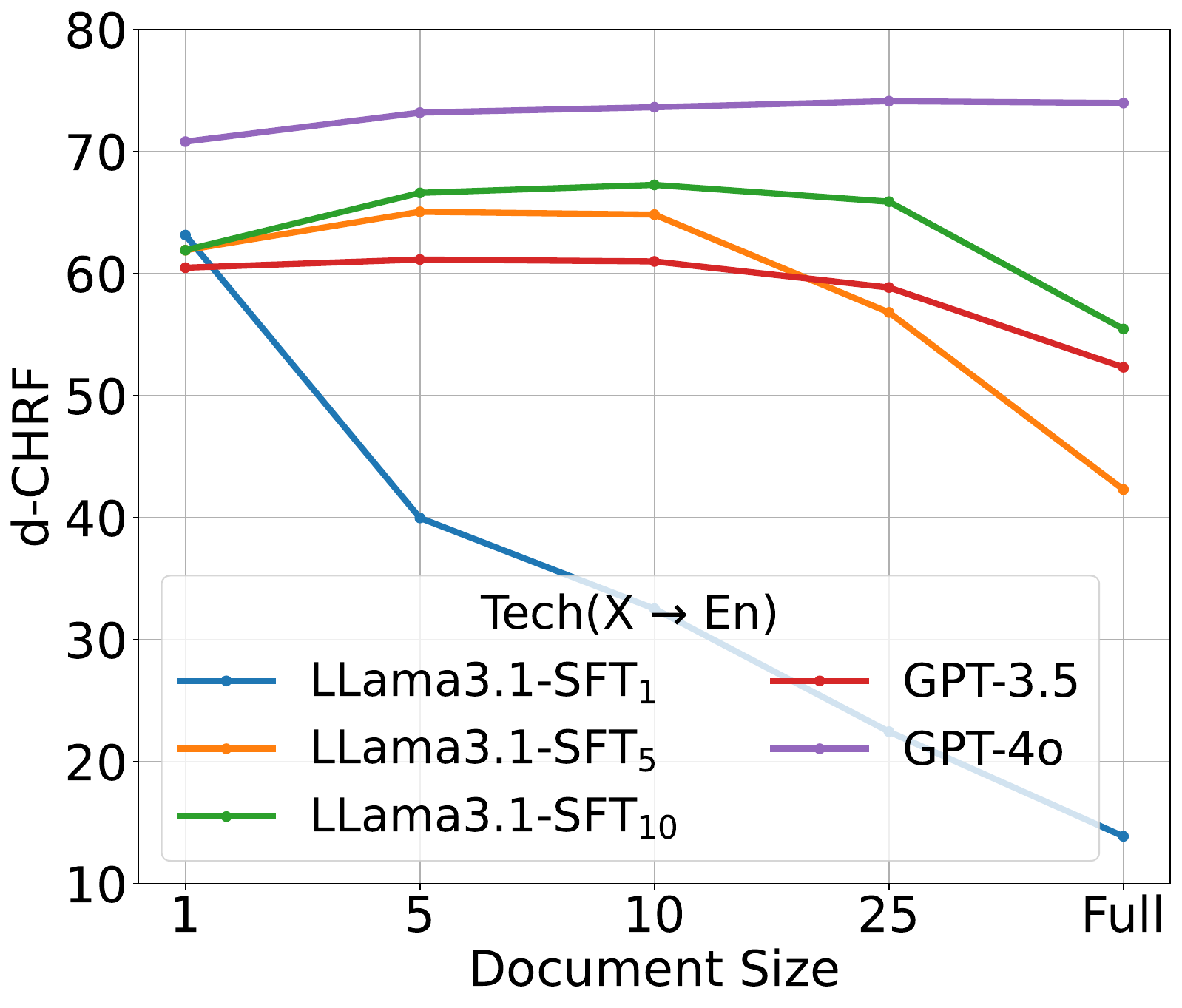}
    \end{subfigure}
\caption{Average chrF score across languages for documents of different sizes.}
\label{fig:doct_trend}
\end{figure*} 

\subsection{Document-level evaluation}

For document-level evaluation, we split the documents into chunks of 10 sentences and translate these chunks using the different models. In \Cref{tab:full_result_health_pt1_25,tab:full_result_tech_pt1_25} we provide the full results on the merged pseudo-documents using d-chrF and d-BLEU. And below are some other relevant points from the results. It is important to note that we also trained and evaluated NLLB-200 for pseudo-document translation. However, due to its 512-token maximum sequence length, it is not competitive. Nevertheless, the results still show the influence of fine-tuning. Below are other findings.

\paragraph{Gemma2-IT shows better translation performance.} Compared to the sentence-level setup, where Gemma2-IT and LLaMAX3-Alpaca achieved similar performance on average, in the pseudo-document setup, Gemma2-IT not only outperforms LLaMAX3-Alpaca but also surpasses GPT-3.5. Although we cannot provide an exact explanation for this performance, we hypothesize that its pre-training setup might be a contributing factor.

\paragraph{Fine-tuning data has an impact on translation quality.} Our results show that both LLama3.1 and LLaMAX3 models, when fine-tuned on sentences, performed significantly worse on pseudo-document evaluations compared to the same models fine-tuned on pseudo-documents for both domains. All these models were trained using a similar setup, with the primary difference being the data used for fine-tuning.

\paragraph{Language-specific performance trends} Overall, no clear trend is observed in MT performance across language family classes. However, Amharic (a non-Latin script language) and \yoruba (a heavily diacriticitized language) result in the lowest chrF scores, while Swahili—the most widely spoken indigenous African language—performs best.

\subsection{Findings from GPT-4o as a judge}
\label{sec:app_gpt_analysis_result}

In \Cref{tab:gpt_full_health,tab:gpt_full_tech} we present the average GPT-4o evaluation results for four models. When translating into African languages, there is no clear pattern: for example, GPT-3.5, despite having the lowest fluency score, also had the fewest content, lexical, and grammatical errors, which is counterintuitive. In contrast, when translating into English, the pattern is clear and consistent: translations of pseudo-documents show better fluency and fewer errors overall. These results suggest that using GPT-4o as a translation judge is not yet well-suited for low-resource languages.

\begin{figure*}[t]
\setlength{\belowcaptionskip}{-4pt}
  \centering
    \begin{subfigure}{0.235\textwidth}
        \includegraphics[width=\textwidth]{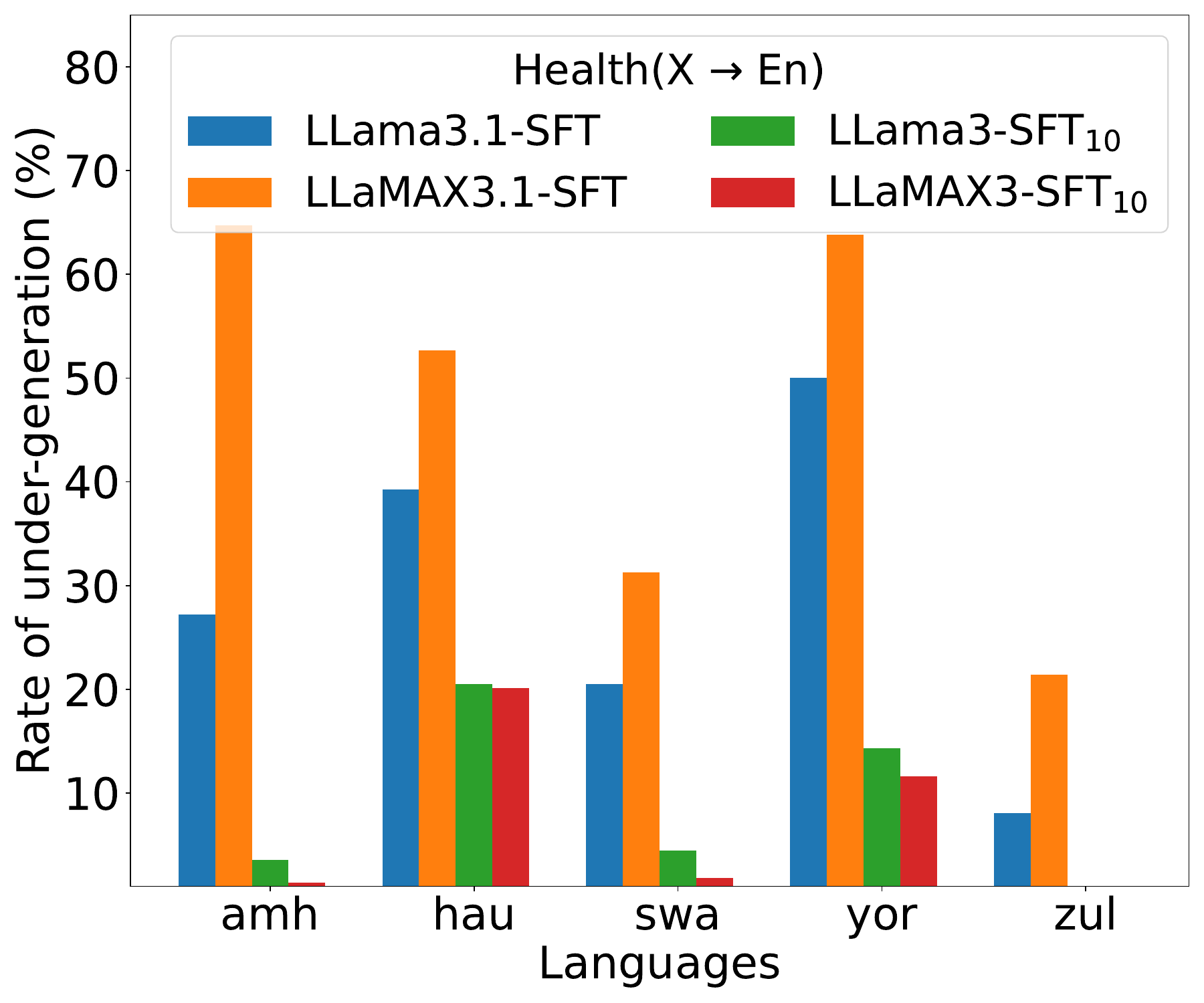}
    \end{subfigure}
     ~
    \begin{subfigure}{0.235\textwidth}
        \includegraphics[width=\textwidth]{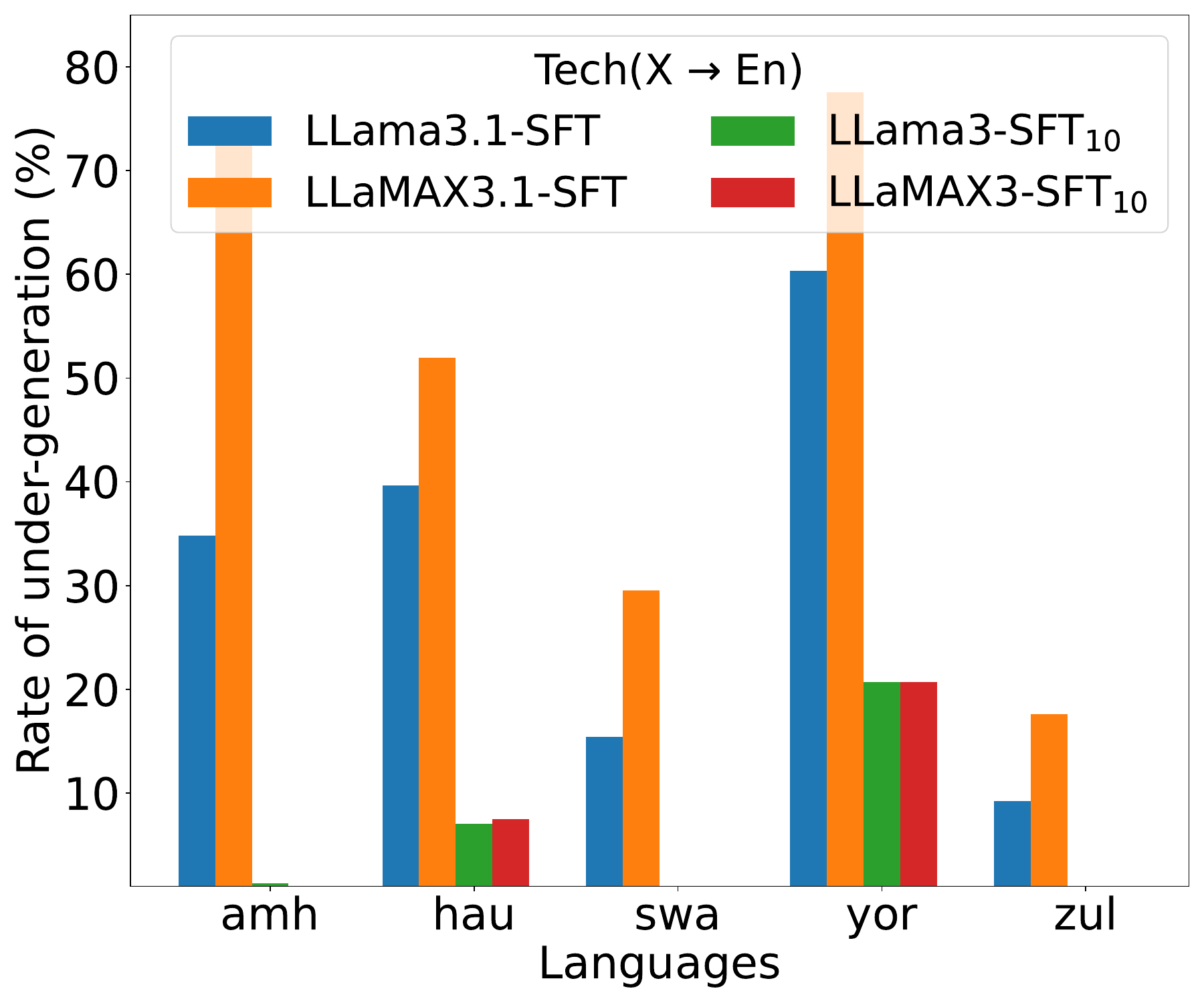}
    \end{subfigure}
     ~     
     \begin{subfigure}{0.235\textwidth}
         \includegraphics[width=\textwidth]{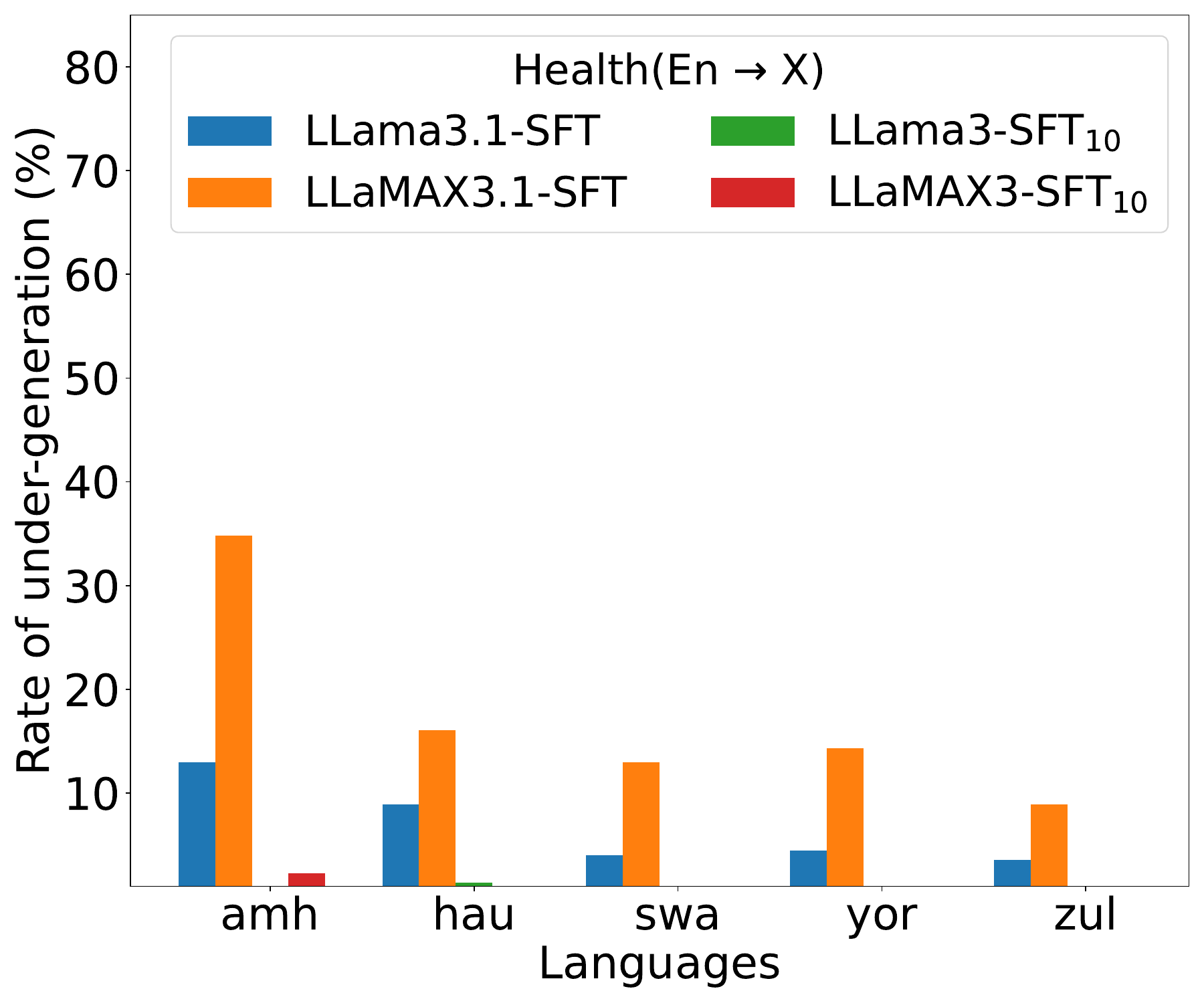}
     \end{subfigure}
     ~
    \begin{subfigure} {0.235\textwidth} 
        \includegraphics[width=\textwidth]{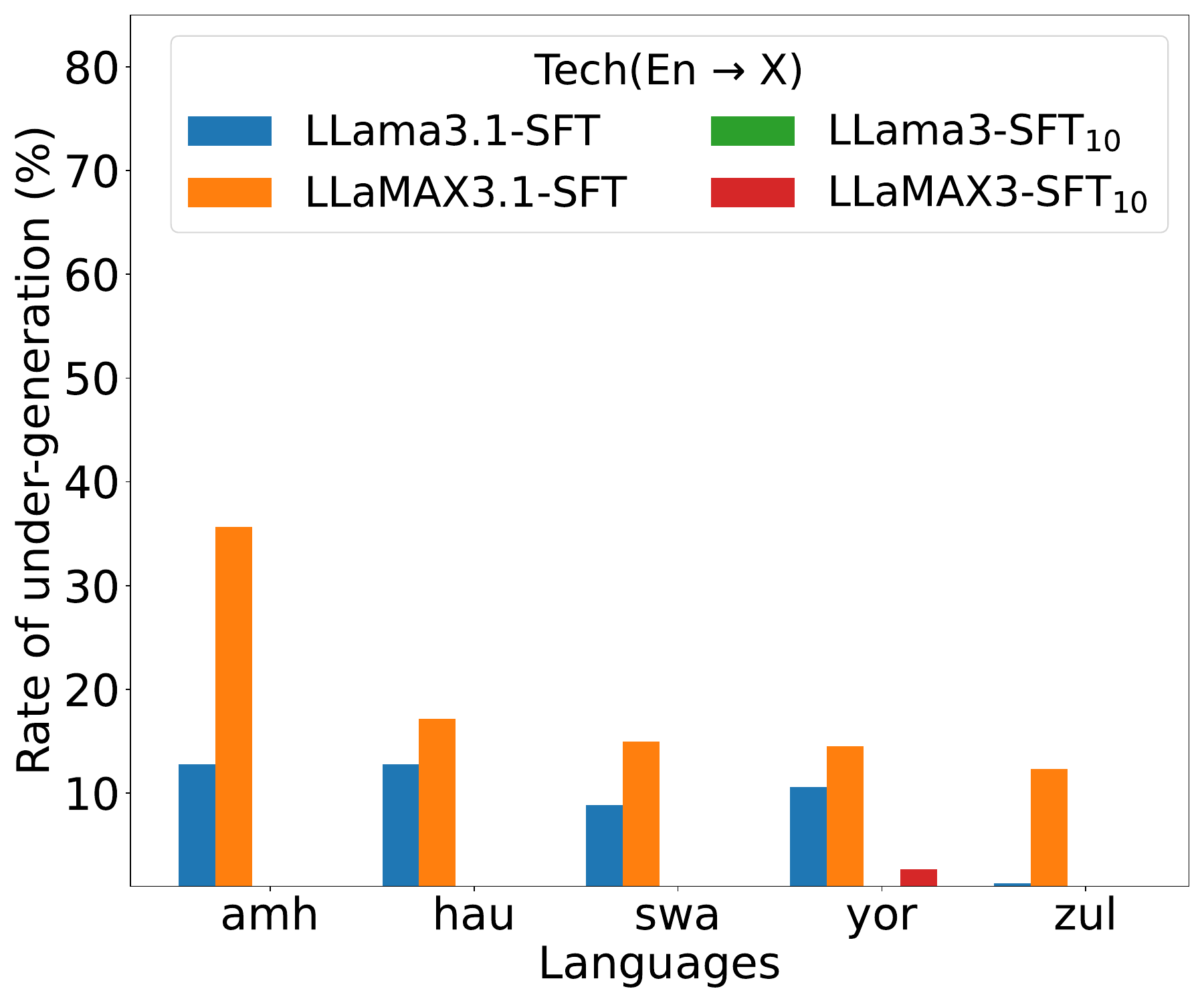}
    \end{subfigure}
\caption{Rate of under-generation in our SFT models.}
\label{fig:app_undergen_our_sft}
\end{figure*}

\begin{figure*}[t]
\setlength{\belowcaptionskip}{-2pt}
  \centering
    \begin{subfigure}{0.235\textwidth}
        \includegraphics[width=\textwidth]{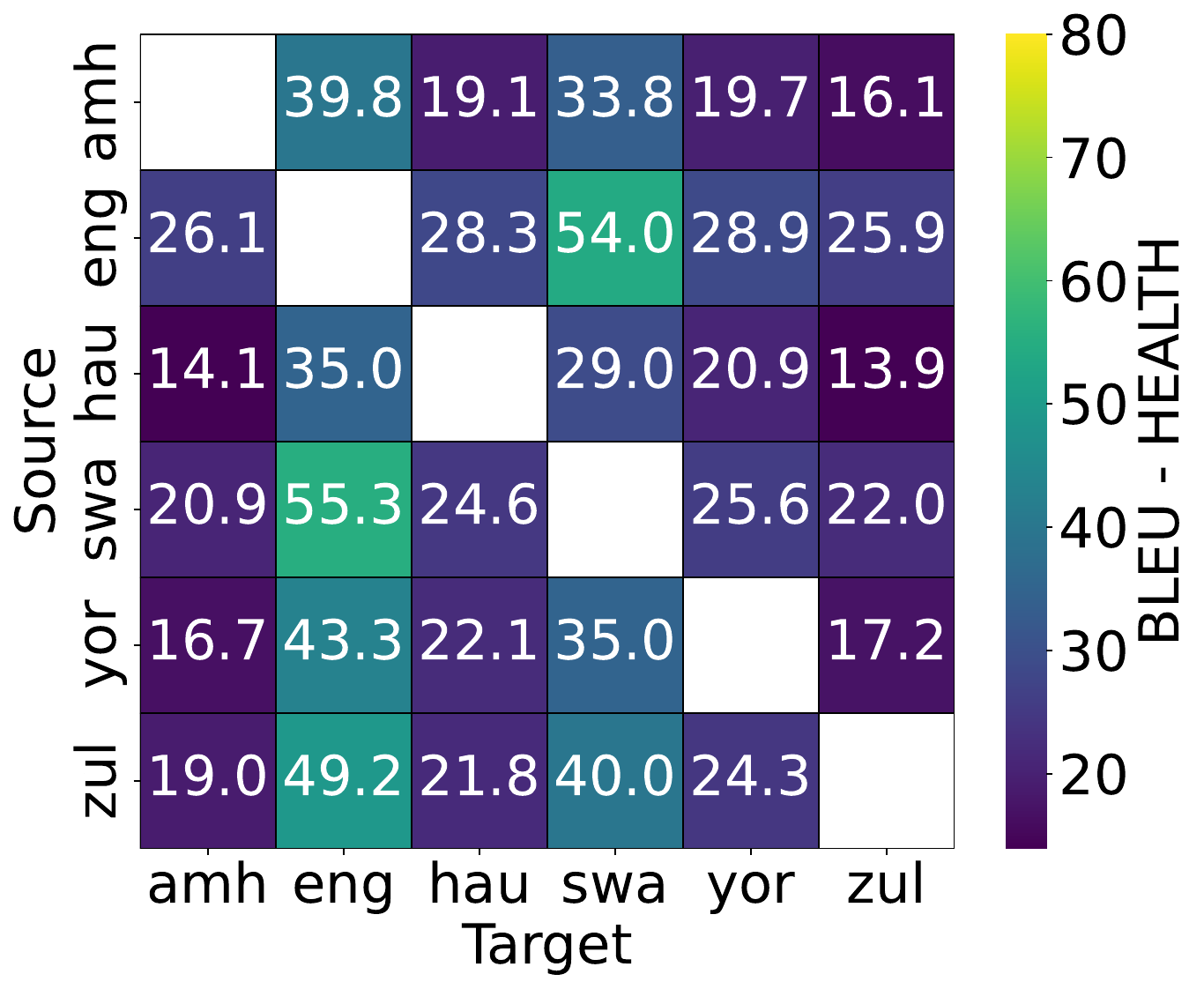}
    \end{subfigure}
    ~
    \begin{subfigure}{0.235\textwidth}
        \includegraphics[width=\textwidth]{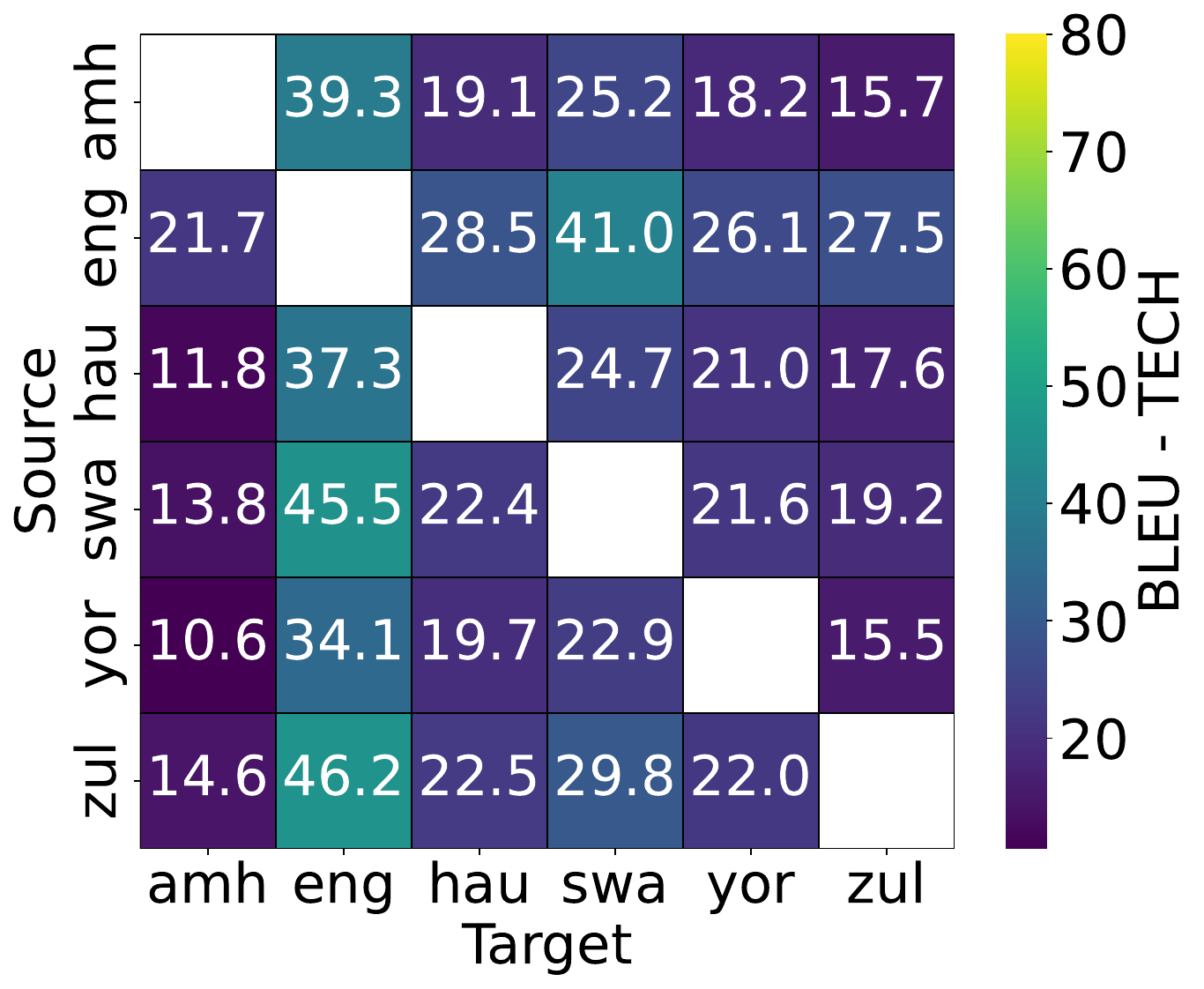}
    \end{subfigure}
    ~
    \begin{subfigure}{0.235\textwidth}
        \includegraphics[width=\textwidth]{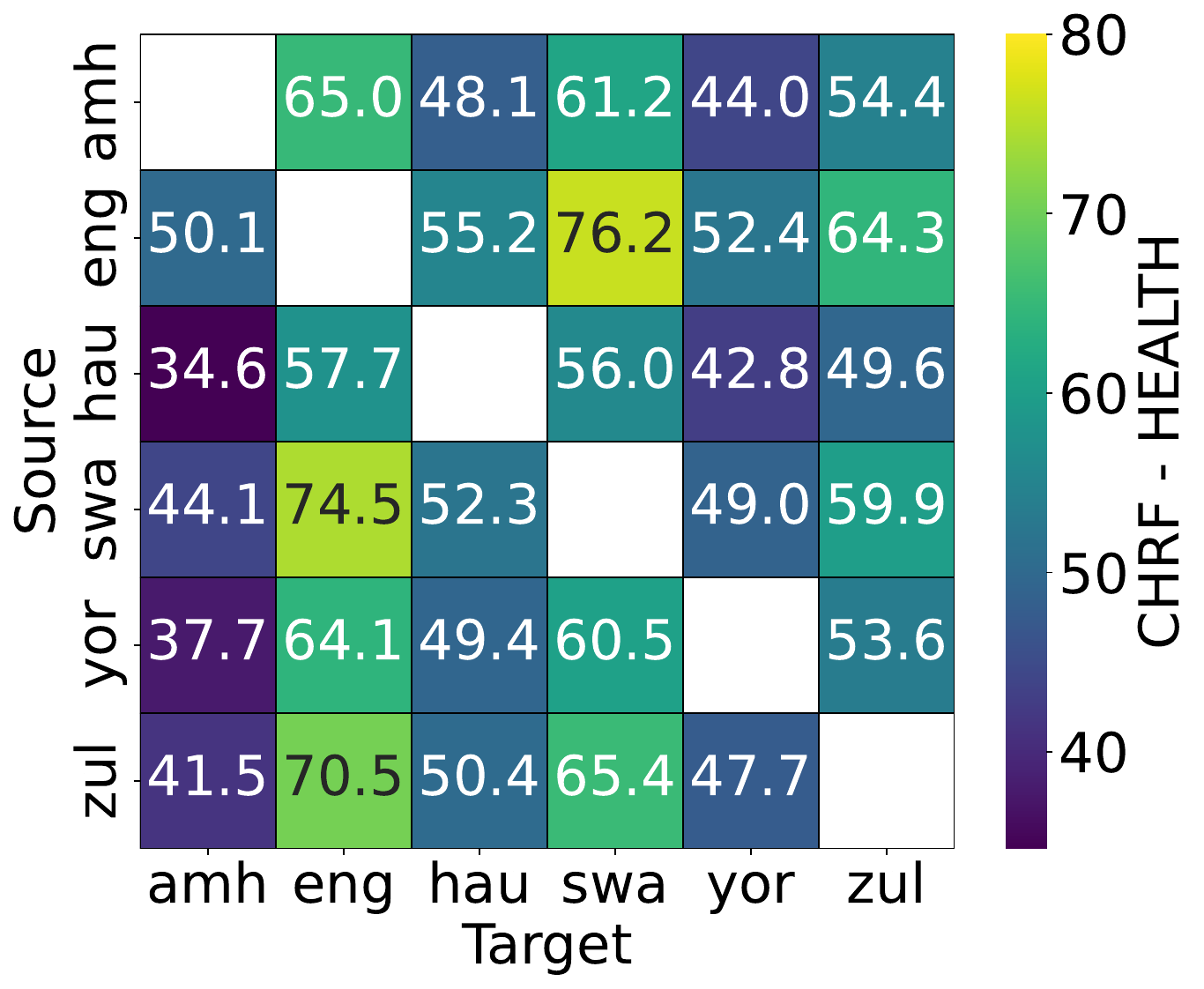}
    \end{subfigure}
    ~
    \begin{subfigure}{0.235\textwidth}
        \includegraphics[width=\textwidth]{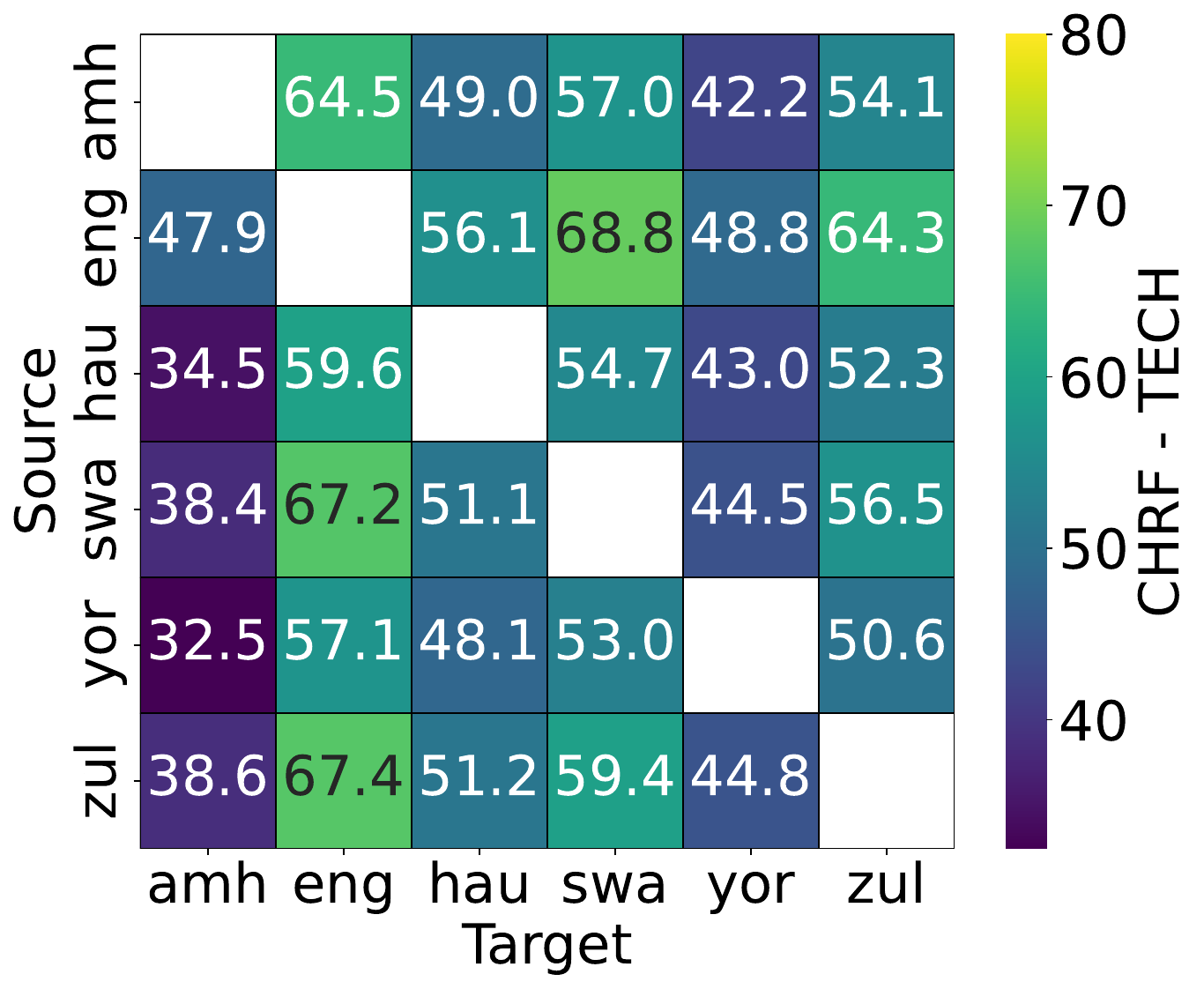}
    \end{subfigure}
\caption{s-BLEU and s-chrF pair-wise comparison of supervised finetuning of NLLB-1.3B on \afridoct}
\label{fig:nllb_perf1}
\end{figure*} 

\begin{table}[t]
\small\centering
\scalebox{0.85}{
  \begin{tabular}{ll|ccccc}
  \toprule
  \textbf{Model} & \textbf{Setup} & \textbf{amh} & \textbf{hau} & \textbf{swh} & \textbf{yor} & \textbf{zul} \\
  \midrule
  \multirow{2}{*}{GPT-3.5} &Sent & 18.3 & 42.1 & \textbf{64.1} & 12.6 & 8.9 \\
& Doc10 & 4.8 & 32.3 & 58.5 & 6.9 & 17.5 \\
\multirow{2}{*}{LLaMAX3-SFT$_{1}$} &Sent & \textbf{58.1} & \textbf{86.0} & 62.1 & \textbf{66.1} & \textbf{52.0} \\ 
&Doc10 & 19.0 & 54.5 & 38.7 & 40.1 & 21.6 \\
LLaMAX3-SFT$_{10}$ & Doc10 & 54.3 & 83.7 & 61.7 & 62.3 & 37.1\\
  \bottomrule
  \end{tabular}
  }
  \vspace{-2mm}
  \caption{Average DA score (scale 0–100) from three human evaluators per language in the \tech domain.}
  \vspace{-4mm}
  \label{tab:human_eval_tech}
\end{table}

\subsection{Findings from human evaluation}
\label{app_human_eval_result}
We were able to obtain DA scores from three annotators for all the languages. For each language, we calculated inter-annotator agreement using Krippendorff’s alpha $\alpha$ over 30 document instances. We obtained $\alpha$ scores of 0.46, 0.57, 0.40, and 0.81, and 0.54 for Amharic, Hausa, Swahili, \yoruba, and Zulu respectively. These are relatively low scores, except for \yoruba. We present the average DA scores in \Cref{tab:human_eval_health,tab:human_eval_tech} for the \health and \tech domains, respectively. The results show that annotators rate documents translated at the sentence-level as higher quality than those translated at the pseudo-document level. Additionally, GPT-3.5 received the lowest ratings among the three models. LLaMAX3-SFT$_{1}$, a model trained on sentence-level data, was rated the best across all languages when evaluated on sentences. However, when evaluated on pseudo-documents, its performance was rated lower than that of LLaMAX3-SFT$_{10}$. These findings are consistent with the d-chrF scores for the models, but they do not align with the evaluations from GPT-4o as a judge.

\begin{figure*}[t]
\setlength{\belowcaptionskip}{-2pt}
  \centering
    \begin{subfigure}{0.235\textwidth}
        \includegraphics[width=\textwidth]{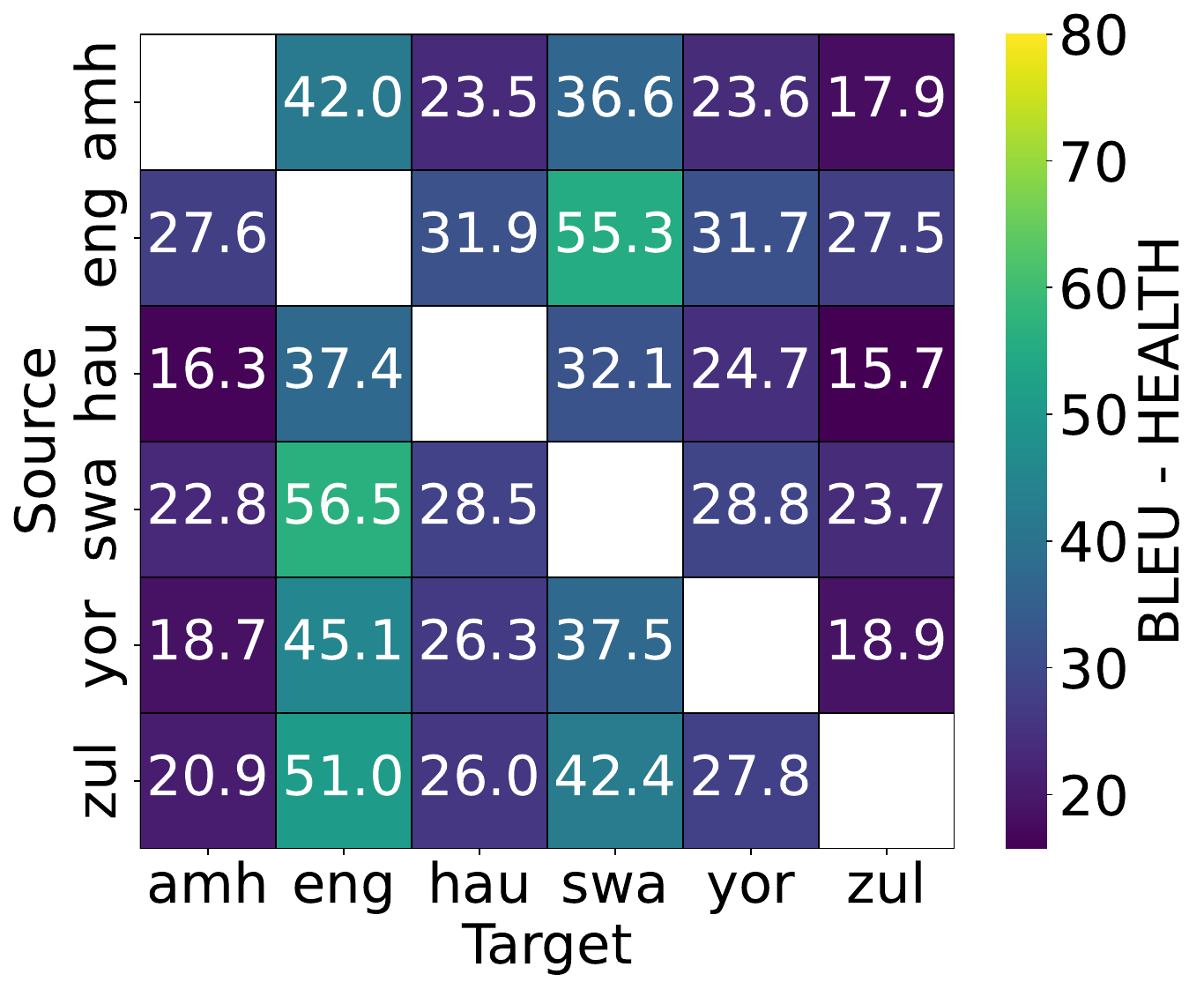}
    \end{subfigure}
    ~
    \begin{subfigure}{0.235\textwidth}
        \includegraphics[width=\textwidth]{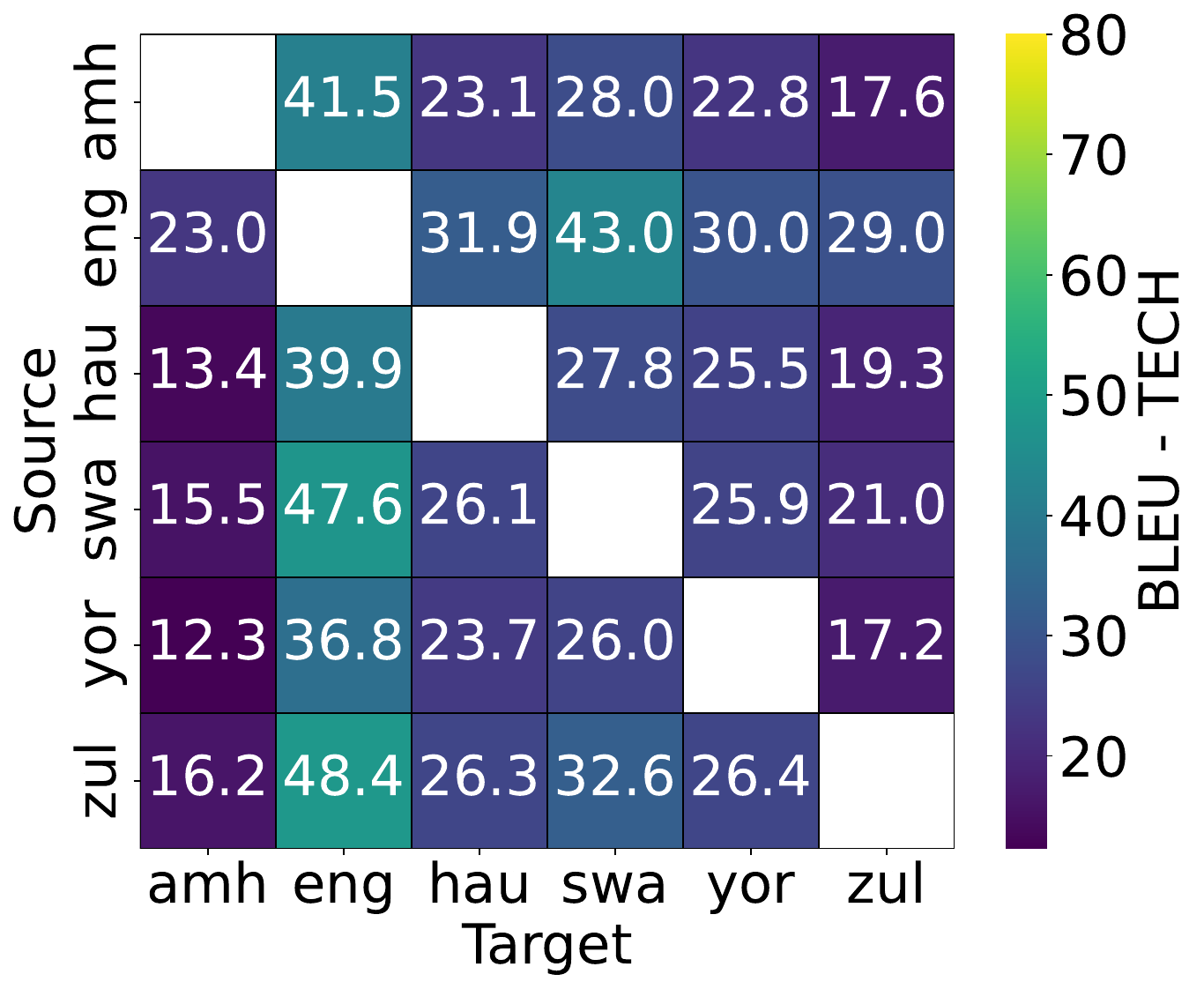}
    \end{subfigure}
    ~
    \begin{subfigure}{0.235\textwidth}
        \includegraphics[width=\textwidth]{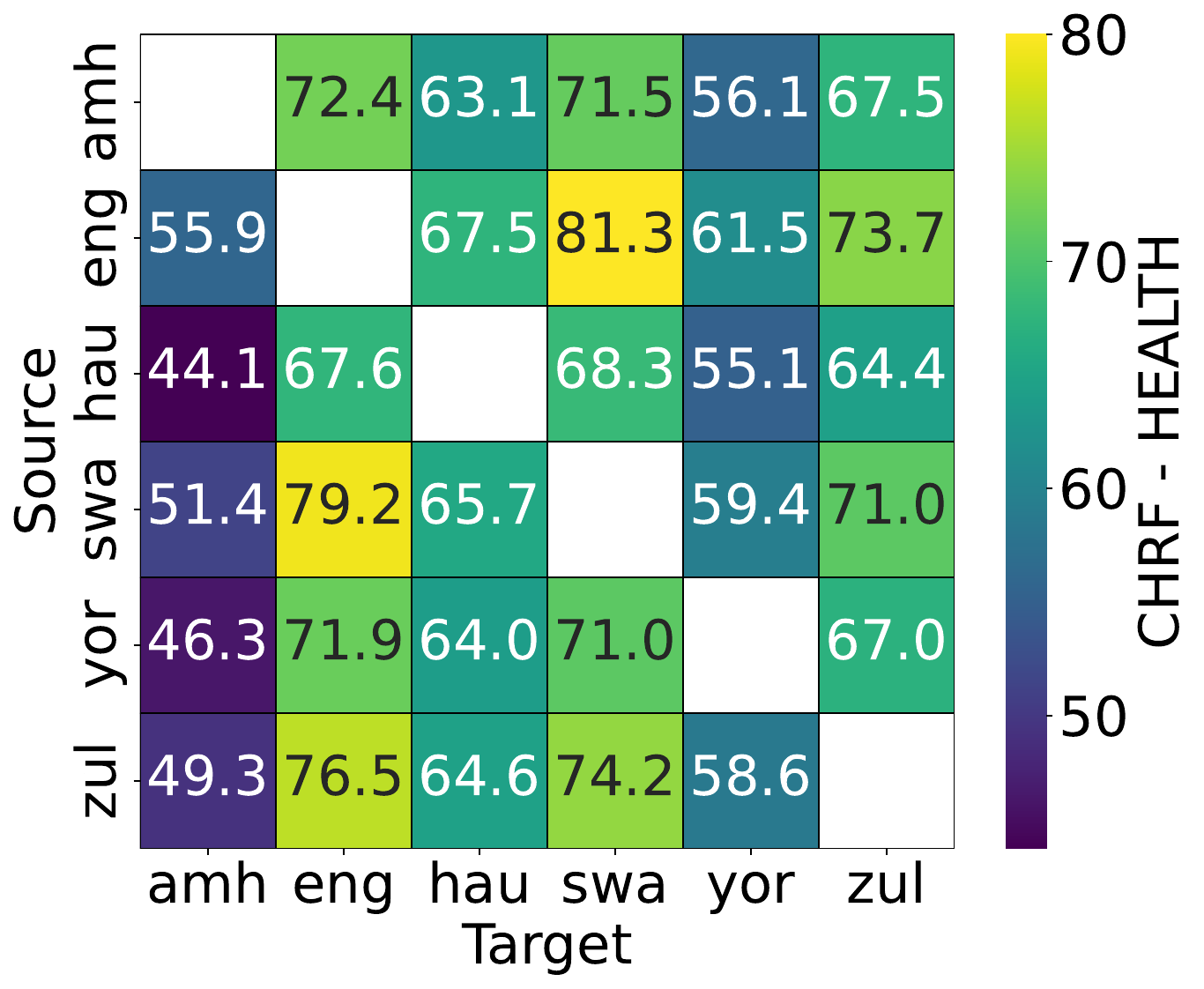}
    \end{subfigure}
    ~
    \begin{subfigure}{0.235\textwidth}
        \includegraphics[width=\textwidth]{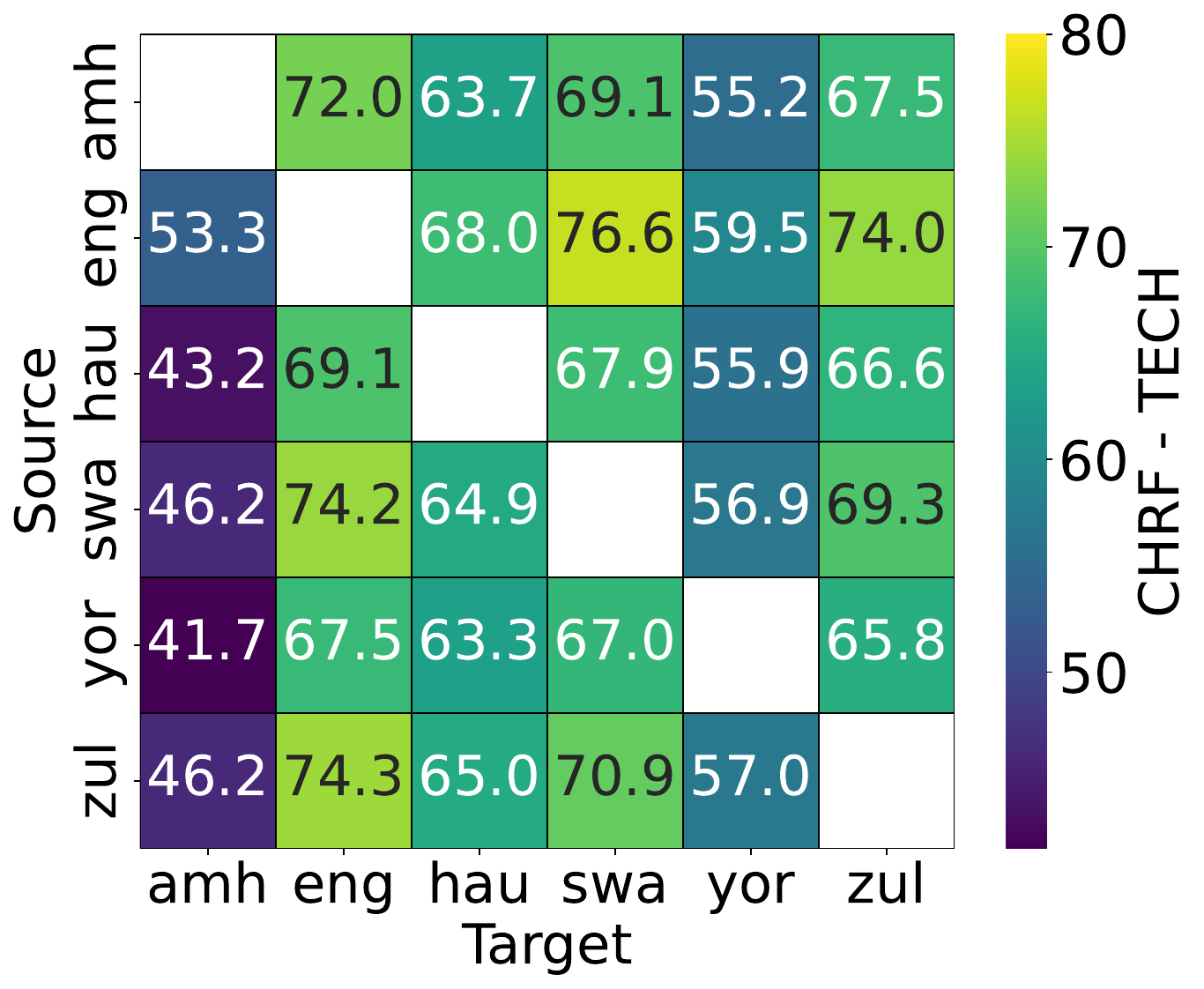}
    \end{subfigure}
\caption{d-BLEU and d-chrF pair-wise comparison of supervised finetuning of NLLB-1.3B on \afridoct}
\label{fig:nllb_perf2}
\end{figure*} 
\begin{figure*}[t]
\setlength{\belowcaptionskip}{-2pt}
  \centering
    \begin{subfigure}{0.235\textwidth}
        \includegraphics[width=\textwidth]{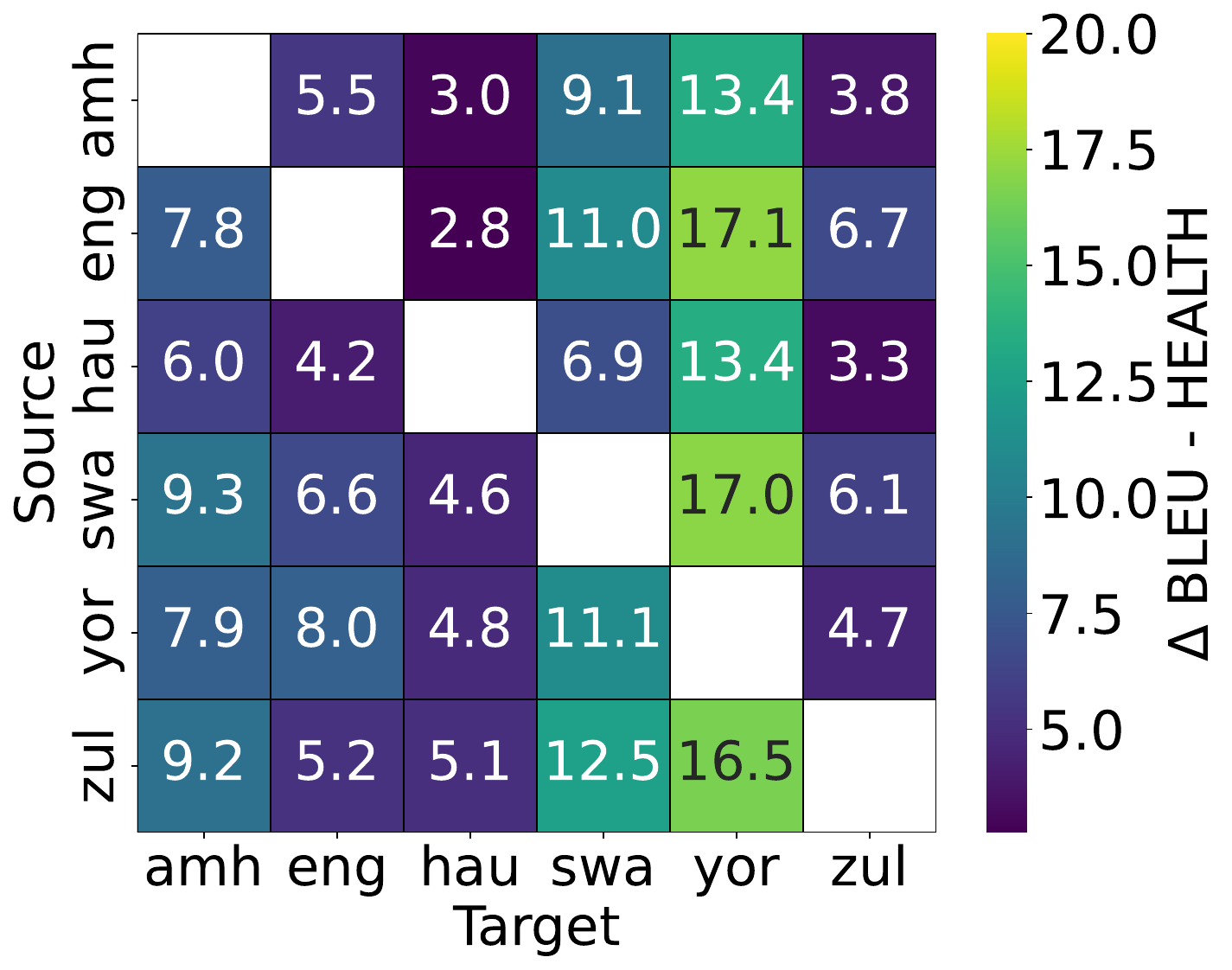}
    \end{subfigure}
    ~
    \begin{subfigure}{0.235\textwidth}
        \includegraphics[width=\textwidth]{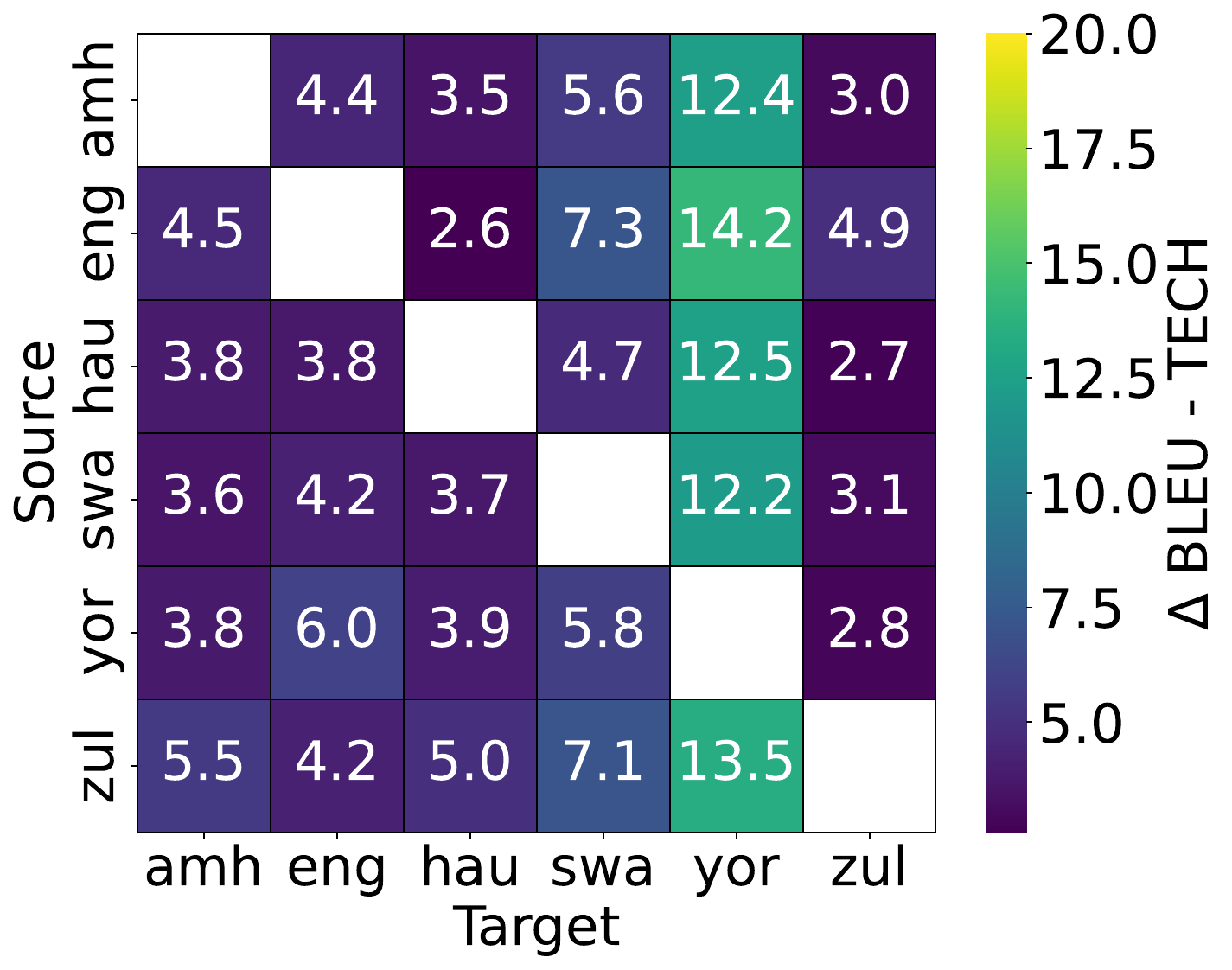}
    \end{subfigure}
    ~
    \begin{subfigure}{0.235\textwidth}
        \includegraphics[width=\textwidth]{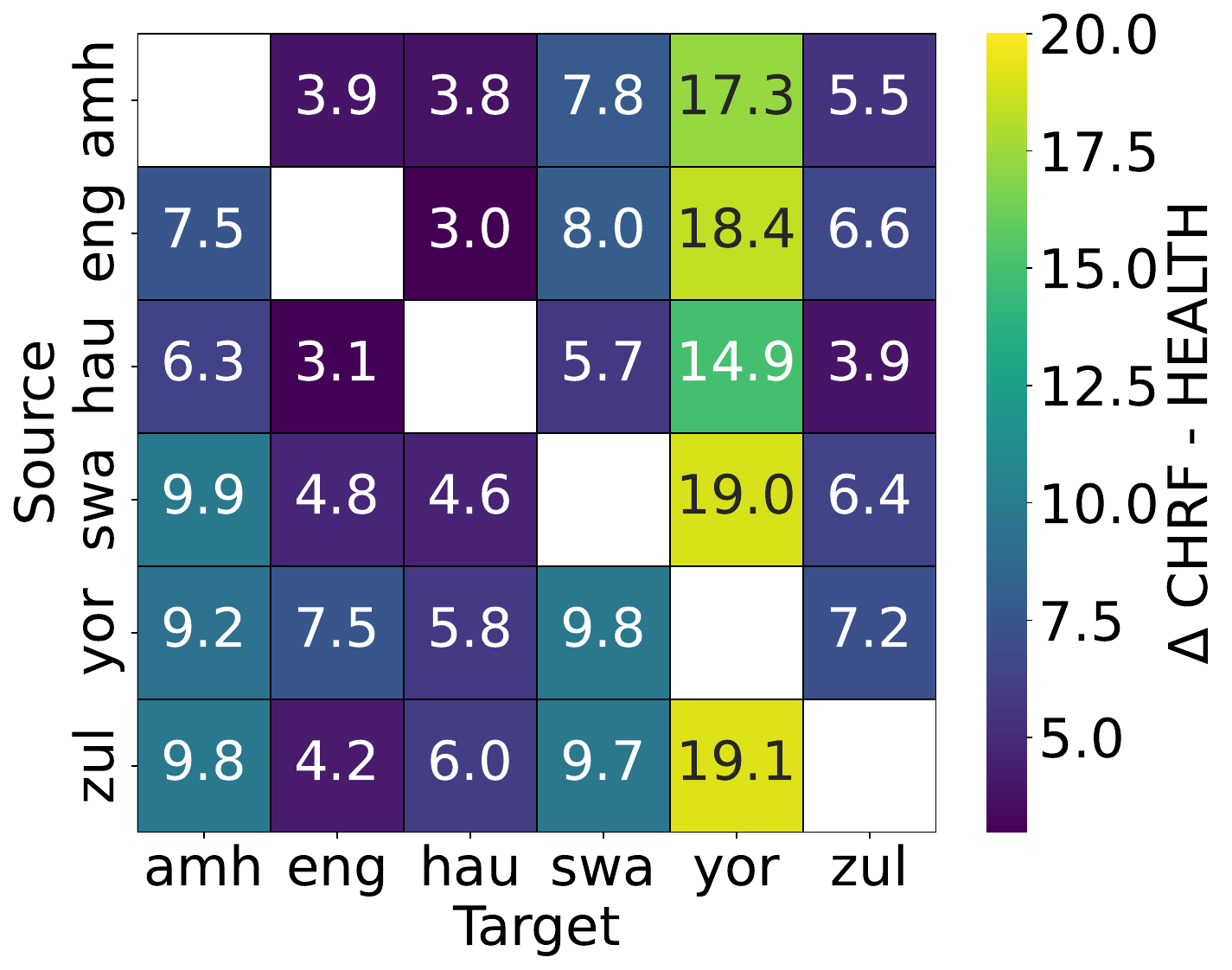}
    \end{subfigure}
    ~
    \begin{subfigure}{0.235\textwidth}
        \includegraphics[width=\textwidth]{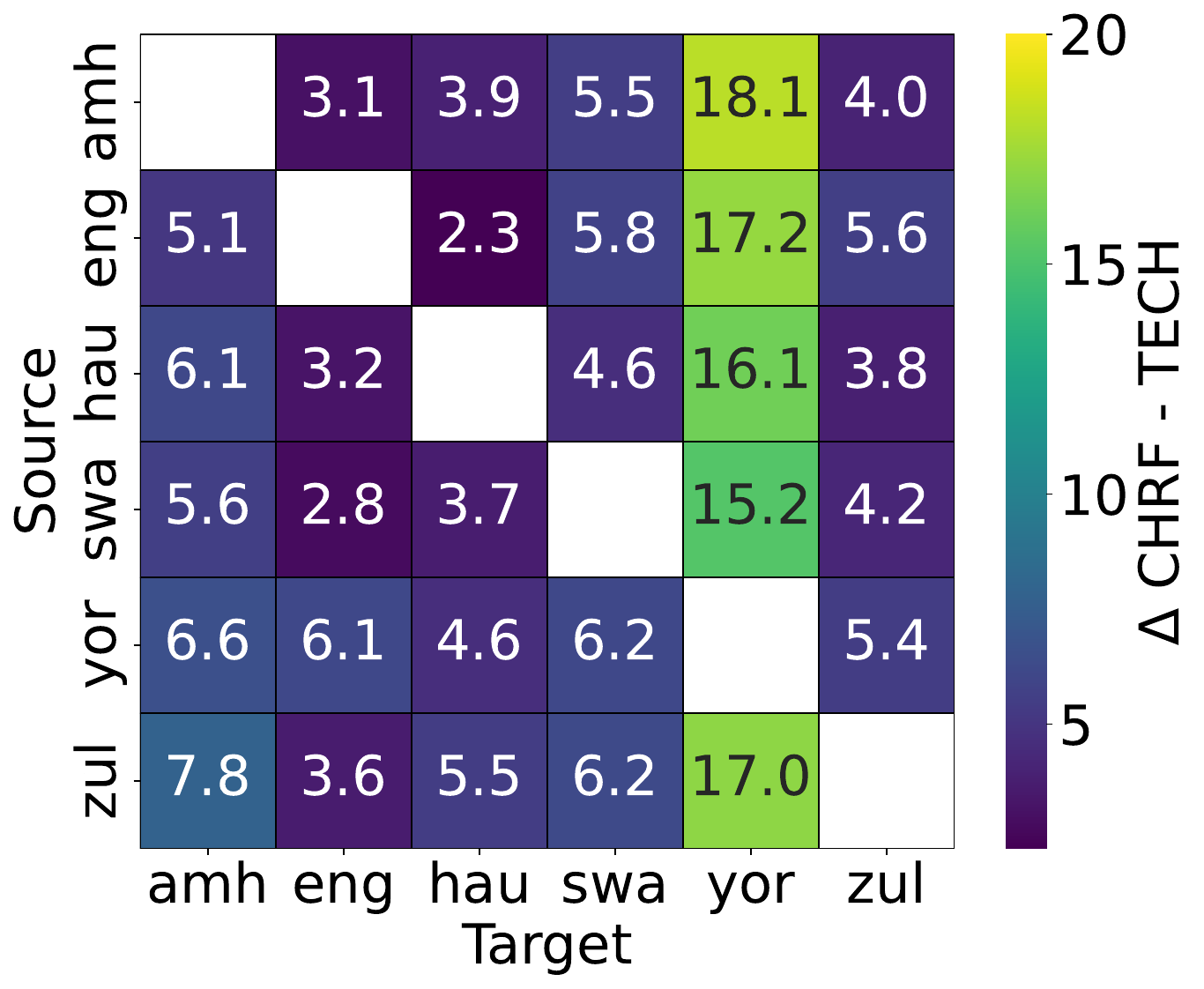}
    \end{subfigure}
\caption{Change ($\Delta$) in s-BLEU and s-chrF for sentence evaluation comparing NLLB1.3B before and after supervised finetuning on \afridoct}
\label{fig:nllb_diff1}
\end{figure*} 
\begin{figure*}[t]
\setlength{\belowcaptionskip}{-2pt}
  \centering
    \begin{subfigure}{0.235\textwidth}
        \includegraphics[width=\textwidth]{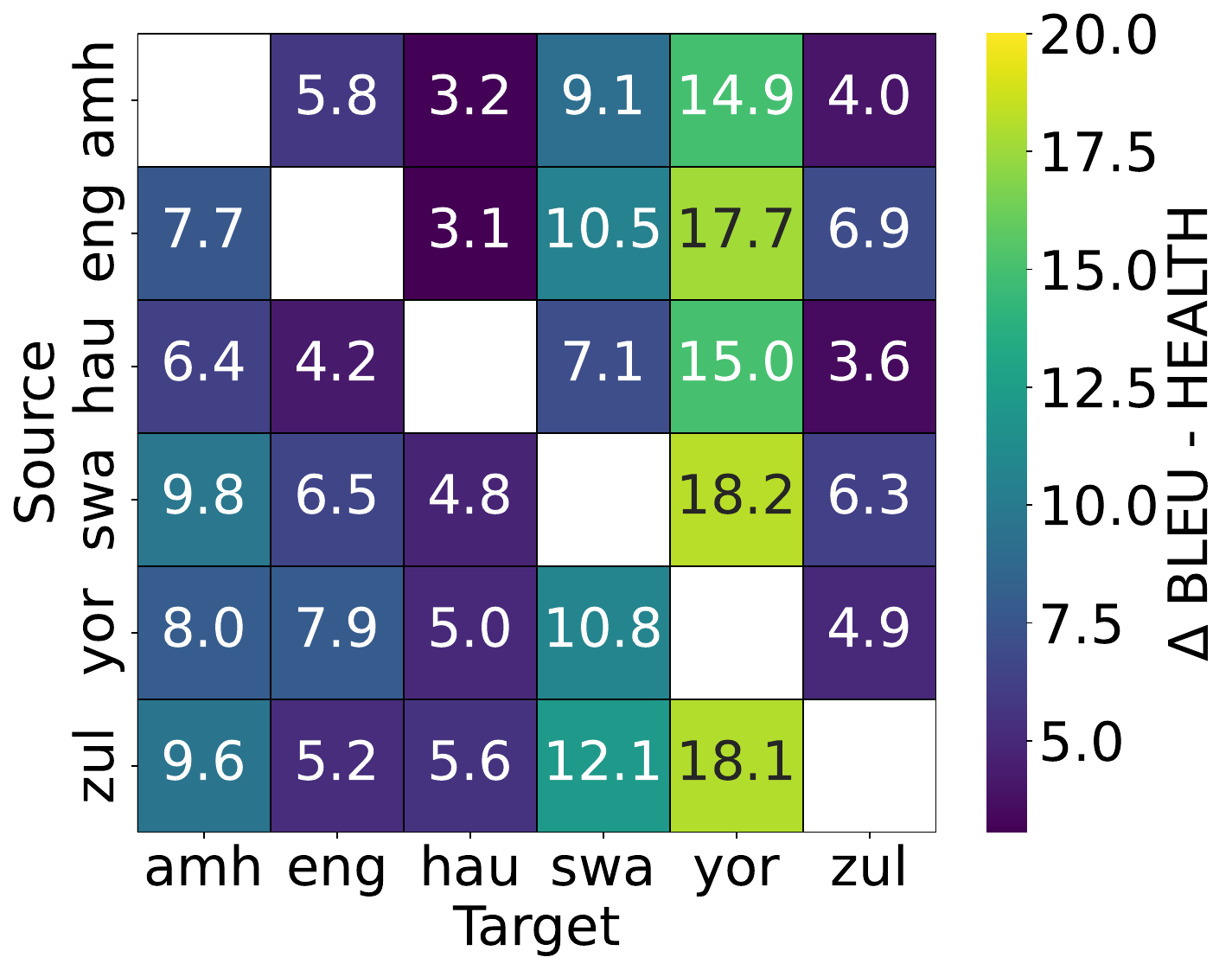}
    \end{subfigure}
    ~
    \begin{subfigure}{0.235\textwidth}
        \includegraphics[width=\textwidth]{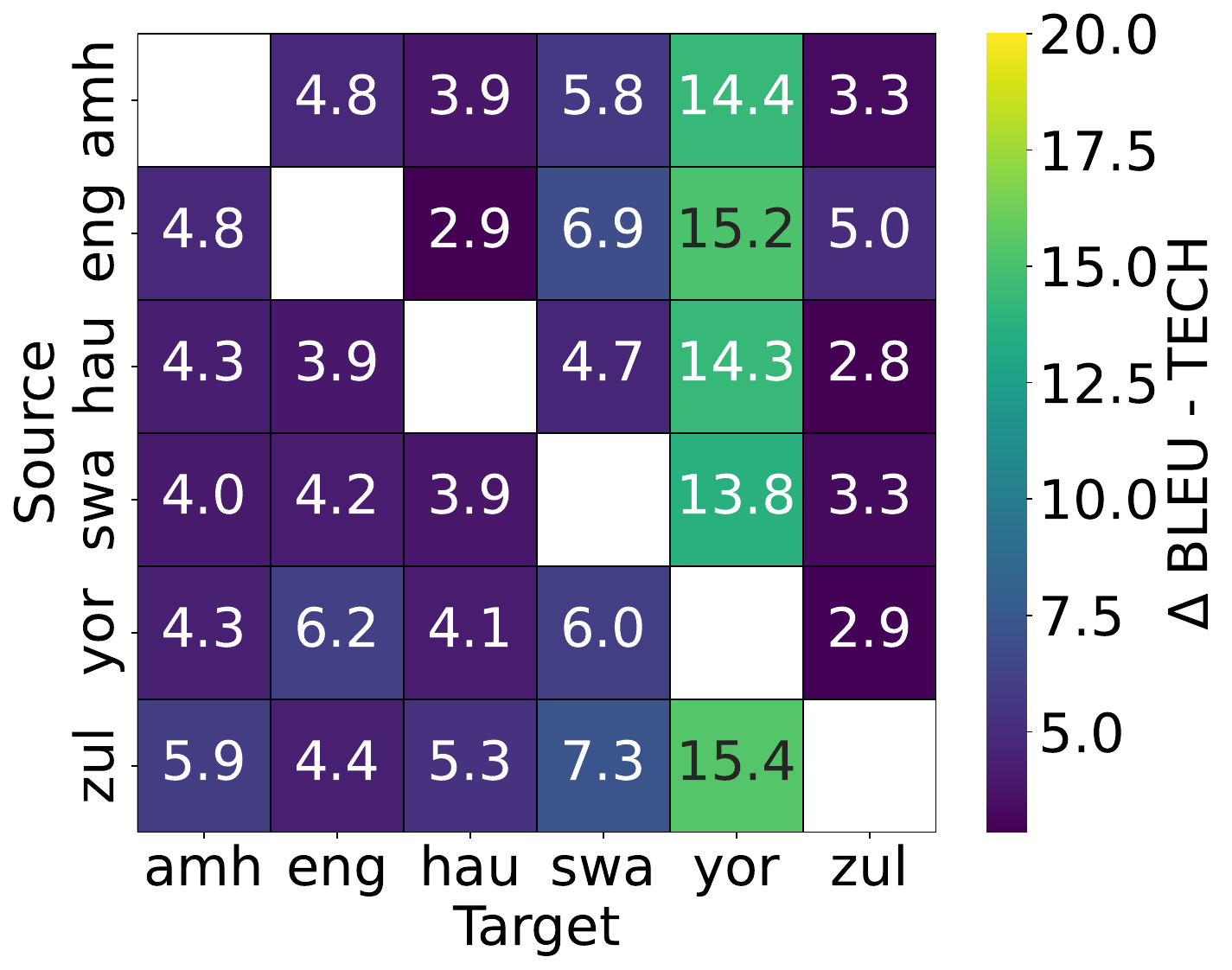}
    \end{subfigure}
    ~
    \begin{subfigure}{0.235\textwidth}
        \includegraphics[width=\textwidth]{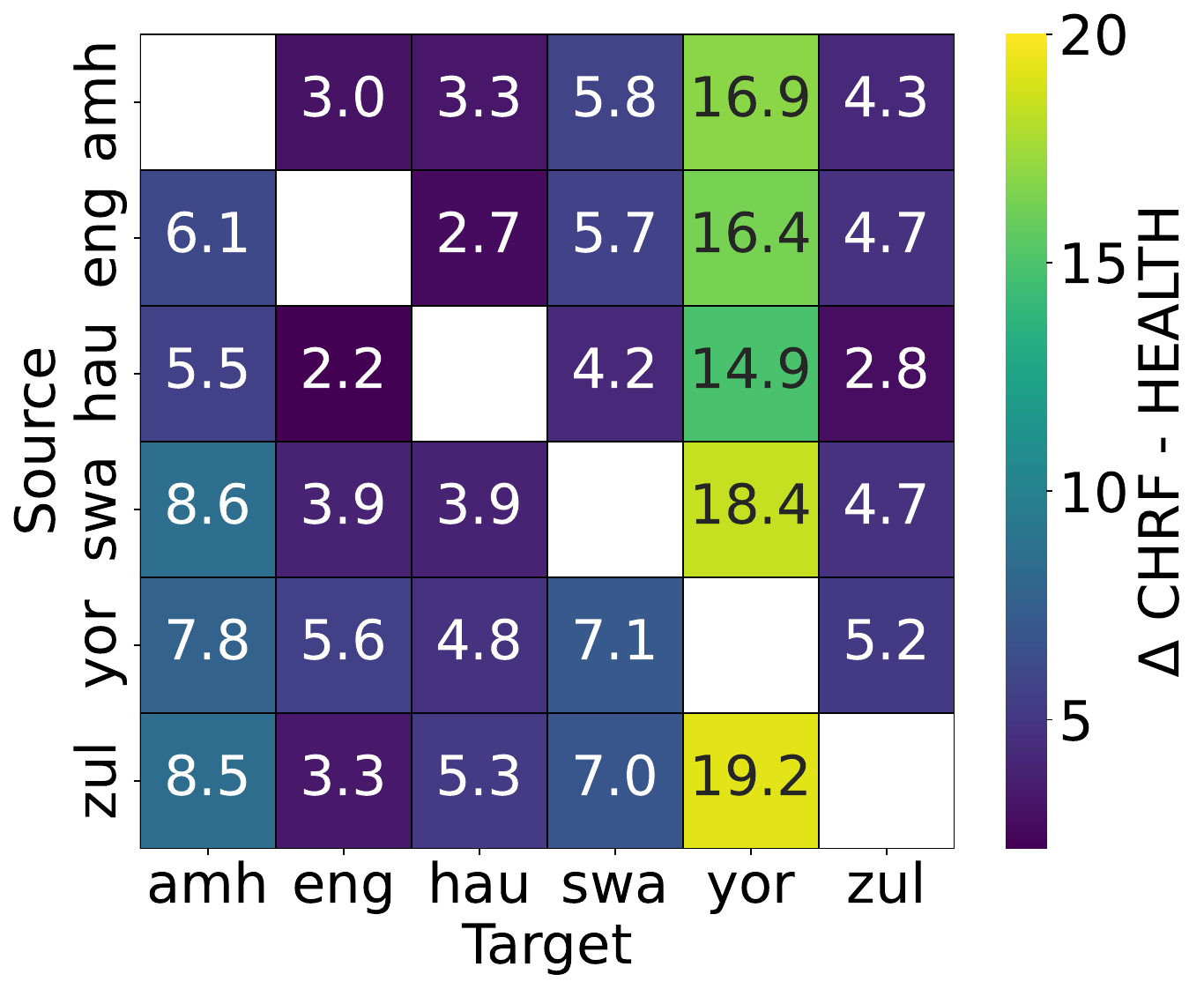}
    \end{subfigure}
    ~
    \begin{subfigure}{0.235\textwidth}
        \includegraphics[width=\textwidth]{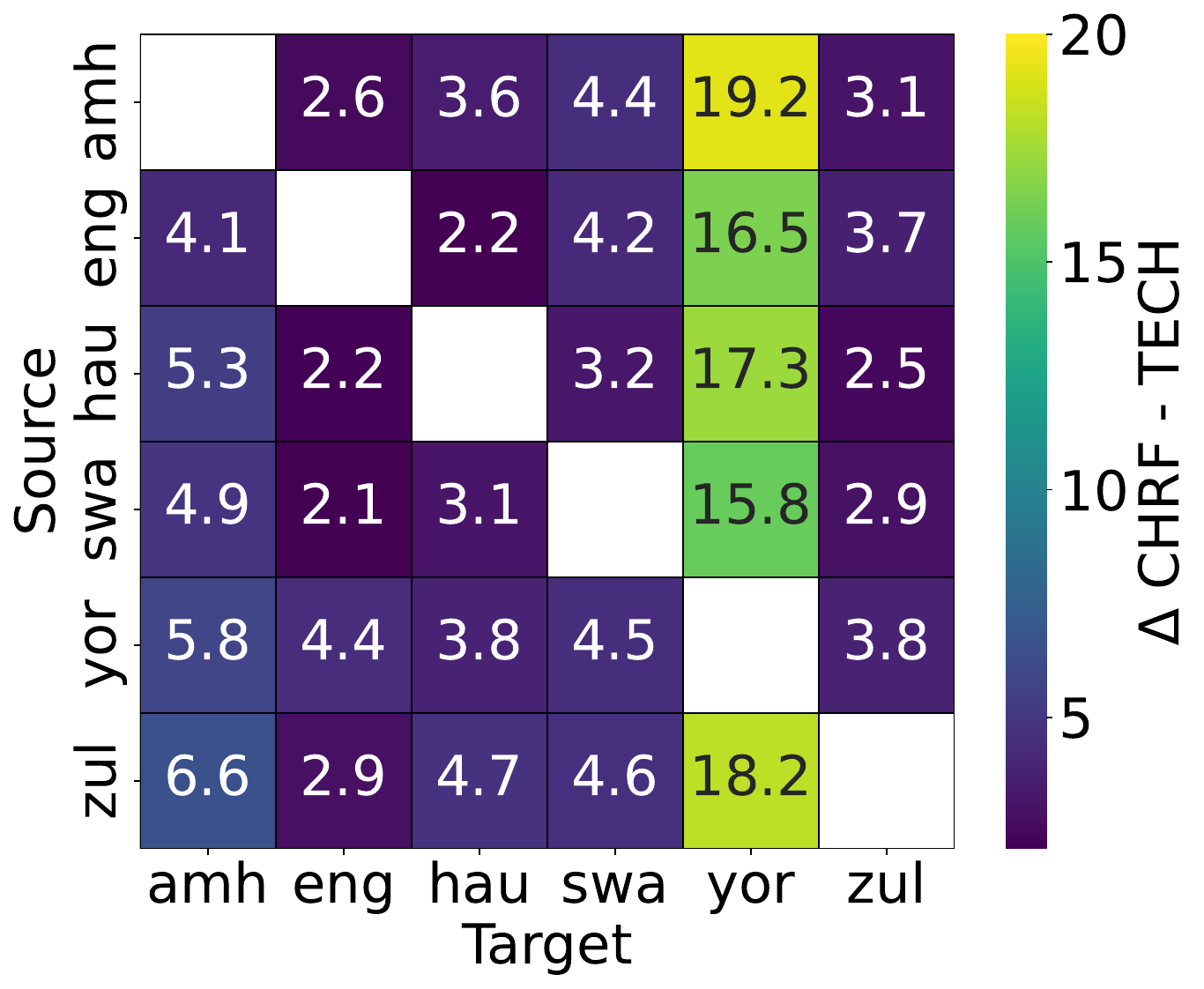}
    \end{subfigure}
\caption{Change ($\Delta$) in d-BLEU and d-chrF for sentence evaluation comparing NLLB1.3B before and after supervised finetuning on \afridoct}
\label{fig:nllb_diff2}
\end{figure*}

\section{More discussion and analysis}
\label{sec:app_extra_analysis}

\paragraph{What language benefits more from supervised finetuning?} 
We focus on the sentence-level task and translated across all 30 directions for which the model was trained, evaluating both NLLB-200 (1.3B) and its fine-tuned version using d-chrF. \Cref{fig:nllb_diff1,fig:nllb_diff2} show performance improvements after supervised fine-tuning of NLLB-200 for both domains. The results shows that translating into \yoruba, which is the direction with the lowest d-chrF score from English among all the languages, benefited the most. One major factor contributing to this is the presence of diacritics. Furthermore, looking at their actual performances and not just the differences, our results show that translations into Swahili and English—both relatively high-resource languages—yield higher BLEU and chrF scores (see Figures~\ref{fig:nllb_perf1} and~\ref{fig:nllb_perf2}), even after supervised finetuning. Hence, there is much to be done to improve translation performance between low-resource language pairs.

\FloatBarrier
\begin{table*}[t]
 \footnotesize
 \begin{center}
 \resizebox{\textwidth}{!}{%
\begin{tabular}{ll|ccccc|ccccc||c}
    \toprule
    \rowcolor{gray!30}
    \textbf{Model} & \textbf{Size} & \multicolumn{5}{c}{\textit{eng {$\rightarrow$} X}} & \multicolumn{5}{c}{\textit{X {$\rightarrow$} eng}} & \textbf{AVG} \\
     \rowcolor{gray!30}
     &  & \textbf{amh} & \textbf{hau} & \textbf{swa} & \textbf{yor} & \textbf{zul} & \textbf{amh} & \textbf{hau} & \textbf{swa} & \textbf{yor} & \textbf{zul} &   \\
\midrule
\multicolumn{13}{c}{\textbf{BLEU}} \\
\midrule
\multicolumn{13}{l}{\textbf{Encoder-Decoder}}\\
M2M-100 & 0.4B & $0.8$ & $0.9$ & $25.6$ & $0.6$ & $3.2$ & $6.7$ & $5.8$ & $32.6$ & $1.7$ & $14.4$ & $9.2$ \\
M2M-100 & 1.2B & $2.4$ & $8.9$ & $37.1$ & $2.4$ & $6.9$ & $15.6$ & $13.7$ & $42.6$ & $4.3$ & $23.7$ & $15.8$ \\
NLLB-200 & 0.6B & $18.4$ & $26.5$ & $42.0$ & $10.9$ & $19.6$ & $33.0$ & $30.4$ & $45.7$ & $32.4$ & $42.2$ & $30.1$ \\
Toucan & 1.2B & $6.6$ & $18.7$ & $37.3$ & $6.4$ & $9.4$ & $17.4$ & $22.4$ & $31.9$ & $18.1$ & $25.2$ & $19.3$ \\
\bottomrule
    
  \end{tabular}
  }
  \vspace{-3mm}
  \caption{Performance results of various models on the sentence-level task for the Health domain, measured using document level metric d-BLEU and d-chrF.}
  \vspace{-4mm}
  \label{tab:app_full_sentdoc_health}
  \end{center}
\end{table*}

\begin{table*}[t]
 \footnotesize
 \begin{center}
 \resizebox{\textwidth}{!}{%
  \begin{tabular}{ll|ccccc|ccccc||c}
    \toprule
    \rowcolor{gray!30}
    \textbf{Model} & \textbf{Size} & \multicolumn{5}{c}{\textit{eng {$\rightarrow$} X}} & \multicolumn{5}{c}{\textit{X {$\rightarrow$} eng}} & \textbf{AVG} \\
     \rowcolor{gray!30}
     &  & \textbf{amh} & \textbf{hau} & \textbf{swa} & \textbf{yor} & \textbf{zul} & \textbf{amh} & \textbf{hau} & \textbf{swa} & \textbf{yor} & \textbf{zul} &   \\
\midrule
\multicolumn{13}{c}{\textbf{BLEU}} \\
\midrule
\multicolumn{13}{l}{\textbf{Encoder-Decoder}}\\
M2M-100 & 0.4B & $1.1$ & $1.7$ & $22.6$ & $1.0$ & $4.7$ & $7.1$ & $10.2$ & $29.0$ & $2.0$ & $15.1$ & $9.5$ \\
M2M-100 & 1.2B & $2.8$ & $13.2$ & $29.7$ & $3.7$ & $9.1$ & $16.0$ & $19.0$ & $36.3$ & $5.3$ & $23.1$ & $15.8$ \\
NLLB-200 & 0.6B & $16.5$ & $27.3$ & $34.5$ & $12.3$ & $23.4$ & $32.6$ & $33.4$ & $40.5$ & $27.3$ & $40.7$ & $28.9$ \\
Toucan & 1.2B & $5.9$ & $20.4$ & $28.0$ & $8.1$ & $12.4$ & $15.8$ & $25.6$ & $30.1$ & $17.7$ & $25.2$ & $18.9$ \\
\bottomrule
    
  \end{tabular}
  }
  \vspace{-3mm}
  \caption{Performance results of various models on the sentence-level task for the Tech domain, measured using document level metric d-BLEU and d-chrF.}
  \vspace{-4mm}
  \label{tab:app_full_sentdoc_tech}
  \end{center}
\end{table*}

\FloatBarrier

\FloatBarrier
\begin{table*}[t]
 \footnotesize
 \begin{center}
 \resizebox{\textwidth}{!}{%
  \begin{tabular}{ll|ccccc|ccccc||c}
    \toprule
    \rowcolor{gray!30}
    \textbf{Model} & \textbf{Size} & \multicolumn{5}{c}{\textit{eng {$\rightarrow$} X}} & \multicolumn{5}{c}{\textit{X {$\rightarrow$} eng}} & \textbf{AVG} \\
     \rowcolor{gray!30}
     &  & \textbf{amh} & \textbf{hau} & \textbf{swa} & \textbf{yor} & \textbf{zul} & \textbf{amh} & \textbf{hau} & \textbf{swa} & \textbf{yor} & \textbf{zul} &   \\
\midrule
\multicolumn{13}{c}{\textbf{BLEU}} \\
\midrule
\multicolumn{13}{l}{\textbf{Encoder-Decoder}}\\
M2M-100 & 0.4B & $0.6$ & $0.7$ & $24.1$ & $0.6$ & $2.7$ & $5.2$ & $4.3$ & $30.8$ & $1.2$ & $12.3$ & $8.3$ \\
M2M-100 & 1.2B & $1.9$ & $7.0$ & $35.6$ & $2.2$ & $6.3$ & $13.3$ & $11.4$ & $41.0$ & $3.1$ & $21.4$ & $14.3$ \\
NLLB-200 & 0.6B & $16.8$ & $23.2$ & $40.2$ & $8.9$ & $18.3$ & $30.9$ & $27.9$ & $44.2$ & $30.2$ & $40.1$ & $28.1$ \\
Toucan & 1.2B & $5.0$ & $15.8$ & $35.4$ & $5.0$ & $8.6$ & $14.7$ & $19.9$ & $30.1$ & $16.0$ & $22.9$ & $17.3$ \\
\bottomrule
    
  \end{tabular}
  }
  \vspace{-3mm}
  \caption{Performance results of various models on the sentence-level task for the Health domain, measured using sentence level metric s-BLEU, s-CHRF, and s-COMET.}
  \vspace{-4mm}
  \label{tab:app_full_sent_health}
  \end{center}
\end{table*}

\begin{table*}[t]
 \footnotesize
 \begin{center}
 \resizebox{\textwidth}{!}{%
  \begin{tabular}{ll|ccccc|ccccc||c}
    \toprule
    \rowcolor{gray!30}
    \textbf{Model} & \textbf{Size} & \multicolumn{5}{c}{\textit{eng {$\rightarrow$} X}} & \multicolumn{5}{c}{\textit{X {$\rightarrow$} eng}} & \textbf{AVG} \\
     \rowcolor{gray!30}
     &  & \textbf{amh} & \textbf{hau} & \textbf{swa} & \textbf{yor} & \textbf{zul} & \textbf{amh} & \textbf{hau} & \textbf{swa} & \textbf{yor} & \textbf{zul} &   \\
\midrule
\multicolumn{13}{c}{\textbf{BLEU}} \\
\midrule
\multicolumn{13}{l}{\textbf{Encoder-Decoder}}\\
M2M-100 & 0.4B & $0.9$ & $1.2$ & $20.9$ & $0.9$ & $4.0$ & $5.5$ & $8.2$ & $26.7$ & $1.4$ & $12.9$ & $8.3$ \\
M2M-100 & 1.2B & $2.6$ & $10.9$ & $27.6$ & $3.4$ & $8.2$ & $13.6$ & $16.3$ & $34.0$ & $4.0$ & $20.7$ & $14.1$ \\
NLLB-200 & 0.6B & $15.3$ & $24.3$ & $32.3$ & $9.7$ & $22.0$ & $30.7$ & $30.7$ & $38.2$ & $24.6$ & $38.5$ & $26.6$ \\
Toucan & 1.2B & $4.8$ & $17.6$ & $25.8$ & $6.1$ & $11.4$ & $13.2$ & $22.8$ & $27.7$ & $15.1$ & $22.8$ & $16.7$ \\
\bottomrule
    
  \end{tabular}
  }
  \vspace{-3mm}
  \caption{Performance results of various models on the sentence-level task for the Tech domain, measured using sentence level metric s-BLEU, s-CHRF, and s-COMET.}
  \vspace{-4mm}
  \label{tab:app_full_sent_tech}
  \end{center}
\end{table*}

\FloatBarrier

\FloatBarrier
\begin{table*}[t]
 \footnotesize
 \begin{center}
 \resizebox{\textwidth}{!}{%
  \begin{tabular}{ll|ccccc|ccccc||c}
    \toprule
    \rowcolor{gray!30}
    \textbf{Model} & \textbf{Size} & \multicolumn{5}{c}{\textit{eng {$\rightarrow$} X}} & \multicolumn{5}{c}{\textit{X {$\rightarrow$} eng}} & \textbf{AVG} \\
     \rowcolor{gray!30}
     &  & \textbf{amh} & \textbf{hau} & \textbf{swa} & \textbf{yor} & \textbf{zul} & \textbf{amh} & \textbf{hau} & \textbf{swa} & \textbf{yor} & \textbf{zul} &   \\
\midrule
\multicolumn{13}{c}{\textbf{BLEU}} \\
\midrule
\multicolumn{13}{l}{\textbf{Encoder-Decoder}}\\
Toucan & 1.2B & $2.6$ & $9.3$ & $17.4$ & $3.2$ & $4.5$ & $8.6$ & $8.0$ & $18.1$ & $8.2$ & $12.4$ & $9.2$ \\
NLLB-200 & 1.3B & $4.7$ & $8.0$ & $13.7$ & $2.7$ & $8.2$ & $6.1$ & $10.7$ & $20.8$ & $9.9$ & $16.1$ & $10.1$ \\
NLLB-200 & 3.3B & $5.2$ & $5.6$ & $14.2$ & $2.3$ & $7.4$ & $12.1$ & $16.0$ & $26.9$ & $12.7$ & $23.7$ & $12.6$ \\
MADLAD-400 & 3B & $5.9$ & $8.0$ & $17.0$ & $1.5$ & $5.7$ & $31.5$ & $30.9$ & $50.8$ & $14.2$ & $38.7$ & $20.4$ \\
\end{tabular}
  }
  \vspace{-3mm}
  \caption{Performance results of various models on the pseudo-document-level task for the Health domain, measured using document level metric d-BLEU and d-CHRF.}
  \vspace{-4mm}
  \label{tab:full_result_health_pt1_25}
  \end{center}
\end{table*}

\begin{table*}[t]
 \footnotesize
 \begin{center}
 \resizebox{\textwidth}{!}{%
  \begin{tabular}{ll|ccccc|ccccc||c}
    \toprule
    \rowcolor{gray!30}
    \textbf{Model} & \textbf{Size} & \multicolumn{5}{c}{\textit{eng {$\rightarrow$} X}} & \multicolumn{5}{c}{\textit{X {$\rightarrow$} eng}} & \textbf{AVG} \\
     \rowcolor{gray!30}
     &  & \textbf{amh} & \textbf{hau} & \textbf{swa} & \textbf{yor} & \textbf{zul} & \textbf{amh} & \textbf{hau} & \textbf{swa} & \textbf{yor} & \textbf{zul} &   \\
\midrule
\multicolumn{13}{c}{\textbf{BLEU}} \\
\midrule
\multicolumn{13}{l}{\textbf{Encoder-Decoder}}\\
Toucan & 1.2B & $2.2$ & $11.2$ & $13.2$ & $4.1$ & $7.4$ & $8.6$ & $15.6$ & $17.9$ & $10.4$ & $14.8$ & $10.5$ \\
NLLB-200 & 1.3B & $5.1$ & $11.2$ & $14.0$ & $2.7$ & $9.8$ & $5.8$ & $9.7$ & $21.9$ & $8.1$ & $16.9$ & $10.5$ \\
NLLB-200 & 3.3B & $5.1$ & $7.2$ & $11.9$ & $2.2$ & $7.4$ & $10.7$ & $12.9$ & $26.5$ & $10.3$ & $20.9$ & $11.5$ \\
MADLAD-400 & 3B & $5.7$ & $6.8$ & $5.4$ & $1.2$ & $6.7$ & $30.6$ & $33.6$ & $39.4$ & $14.9$ & $35.5$ & $18.0$ \\
\end{tabular}
  }
  \vspace{-3mm}
  \caption{Performance results of various models on the pseudo-document-level task for the Tech domain, measured using document level metric d-BLEU and d-CHRF.}
  \vspace{-4mm}
  \label{tab:full_result_tech_pt1_25}
  \end{center}
\end{table*}

\FloatBarrier

\FloatBarrier
\begin{table*}[t]
 \footnotesize
 \begin{center}
 \resizebox{\textwidth}{!}{%
  \begin{tabular}{p{2.5cm}l|ccccc|ccccc}
    \toprule
    \rowcolor{gray!30}
    \textbf{Model} & \textbf{Setup} & \multicolumn{5}{c}{\textit{eng {$\rightarrow$} X}} & \multicolumn{5}{c}{\textit{X {$\rightarrow$} eng}} \\
     \rowcolor{gray!30}
 & &\textbf{d-CHRF}$\uparrow$ &\textbf{Fluency}$\uparrow$ &\textbf{CE}$\downarrow$ &\textbf{LE}$\downarrow$ &\textbf{GE}$\downarrow$ &\textbf{d-CHRF}$\uparrow$ &\textbf{Fluency}$\uparrow$ &\textbf{CE}$\downarrow$ &\textbf{LE}$\downarrow$ &\textbf{GE}$\downarrow$ \\
\midrule 
\multicolumn{12}{l}{Amharic} \\
\multirow{2}{*}{Aya-101} & Sent & 36.6 &  2.8 & 9.0 & 3.8 &  2.5 & 64.6 &  3.0 &  15.8 & 9.6 & 5.1 \\
 & Doc10 & 28.7 &  2.5 & 8.9 & 2.9 &  2.1 & 61.6 &  3.6 &  12.6 & 8.5 & 3.8 \\
\multirow{2}{*}{GPT-3.5} & Sent & 20.4 &  1.1 & 7.2 & 10.8 &  3.1 & 48.3 &  3.0 &  9.5 & 3.7 & 2.4 \\
 & Doc10 & 11.6 &  1.0 & 3.3 & 1.8 &  1.6 & 41.6 &  4.2 &  6.6 & 2.0 & 1.3 \\
\multirow{2}{*}{LLaMAX3-SFT$_{1}$} & Sent & 46.8 &  3.5 & 10.0 & 2.9 &  2.0 & 66.6 &  3.5 &  11.5 & 6.0 & 2.9 \\
\hline
\bottomrule

  \end{tabular}
  }
  \vspace{-3mm}
  \caption{Document-level evaluation in the \health domain, judged by GPT-4o. Compares sentence- vs. document-level outputs on Fluency (1–5 scale), Content Errors (CE), Lexical (LE), and Grammatical Cohesion Errors (GE). Best scores in bold.}
  \vspace{-5mm}
  \label{tab:gpt_full_health}
  \end{center}
\end{table*}

\begin{table*}[t]
 \footnotesize
 \begin{center}
 \resizebox{\textwidth}{!}{%
  \begin{tabular}{p{2.5cm}l|ccccc|ccccc}
    \toprule
    \rowcolor{gray!30}
    \textbf{Model} & \textbf{Setup} & \multicolumn{5}{c}{\textit{eng {$\rightarrow$} X}} & \multicolumn{5}{c}{\textit{X {$\rightarrow$} eng}} \\
     \rowcolor{gray!30}
 & &\textbf{d-CHRF}$\uparrow$ &\textbf{Fluency}$\uparrow$ &\textbf{CE}$\downarrow$ &\textbf{LE}$\downarrow$ &\textbf{GE}$\downarrow$ &\textbf{d-CHRF}$\uparrow$ &\textbf{Fluency}$\uparrow$ &\textbf{CE}$\downarrow$ &\textbf{LE}$\downarrow$ &\textbf{GE}$\downarrow$ \\
\midrule 
\multicolumn{12}{l}{Amharic} \\
\multirow{2}{*}{Aya-101} & Sent & 37.3 &  3.2 & 9.9 & 3.4 &  3.3 & 65.2 &  2.9 &  19.3 & 11.2 & 5.8 \\
 & Doc10 & 30.1 &  2.6 & 8.3 & 2.4 &  1.9 & 62.5 &  3.5 &  13.2 & 7.8 & 3.9 \\
\multirow{2}{*}{GPT-3.5} & Sent & 22.6 &  1.3 & 8.5 & 4.2 &  3.3 & 47.4 &  2.4 &  8.0 & 3.6 & 2.4 \\
 & Doc10 & 13.5 &  1.1 & 4.3 & 2.0 &  1.9 & 38.5 &  3.6 &  7.5 & 2.8 & 1.7 \\
\multirow{2}{*}{LLaMAX3-SFT$_{1}$} & Sent & 42.8 &  3.2 & 11.6 & 3.0 &  2.4 & 63.0 &  3.3 &  14.2 & 7.9 & 3.4 \\
\hline
\bottomrule
    
  \end{tabular}
  }
  \vspace{-3mm}
  \caption{Document-level evaluation in the \tech domain, judged by GPT-4o. Compares sentence- vs. document-level outputs on Fluency (1–5 scale), Content Errors (CE), Lexical (LE), and Grammatical Cohesion Errors (GE). Best scores in bold.}
  \vspace{-5mm}
  \label{tab:gpt_full_tech}
  \end{center}
\end{table*}

\FloatBarrier

\begin{table*}[ht]
    \centering
    \begin{tabular}{ p{0.95\textwidth} }
        \toprule
        \textbf{Prompt 1} \\
        \texttt{\{system\_prompt\}} \\
        \texttt{Translate the following \{source\_language\} text to \{target\_language\}:} \\
        \texttt{Provide only the translation.} \\
        \texttt{\{source\_language\} text: \{\{source\_sentence\}\}} \\
        \texttt{\{target\_sentence\} text:} \\
        
        \vspace{0.1em}
        
        \textbf{Prompt 2} \\
        \texttt{\{system\_prompt\}} \\
\hline
    \end{tabular}
    \caption{The task prompts used for evaluating LLMs are applied to both sentence-level and document-level translation tasks.}
    \label{tab:prompt_examples}
\end{table*}

\begin{figure*}[t]
    \centering
    \includegraphics[width=\textwidth]{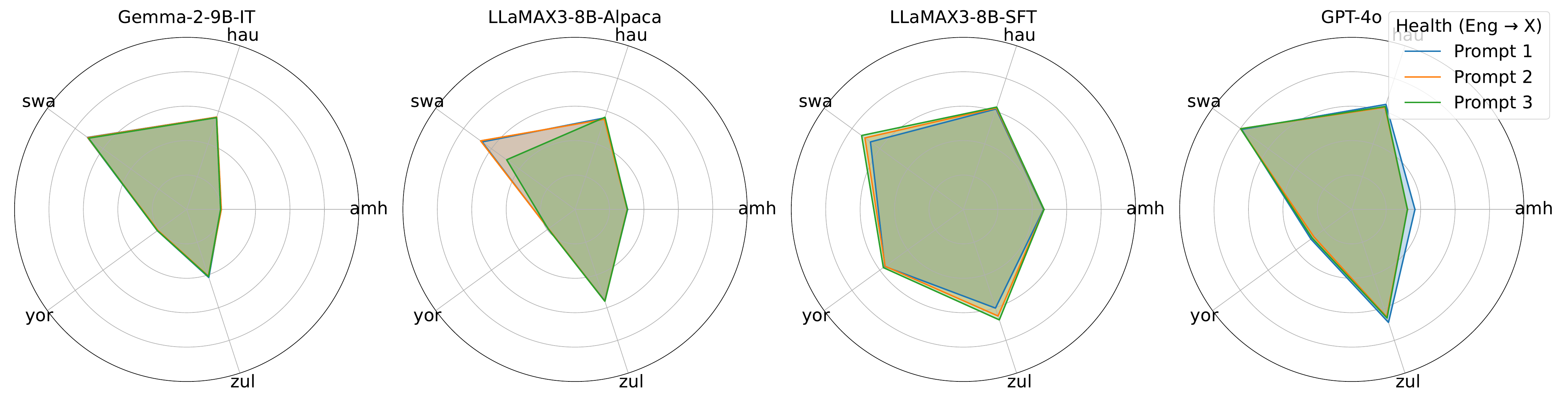}
    \caption{d-chrF scores for some LLMs for sentence-level translation using different prompts when translating into African languages}
    \label{fig:app_health_afro}
\end{figure*}

\begin{figure*}[t]
    \centering
    \includegraphics[width=\textwidth]{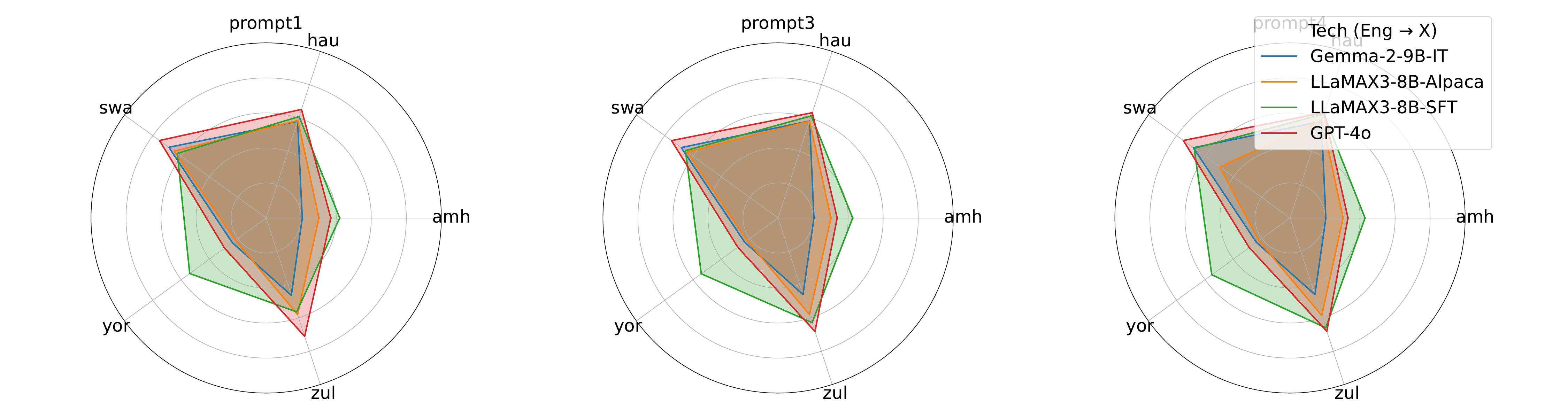}
    \caption{d-chrF scores for some LLMs for sentence-level translation using different prompts when translating into African languages}
    \label{fig:tech_prompt_afro}
\end{figure*}

\begin{figure*}[t]
    \centering
    \includegraphics[width=\textwidth]{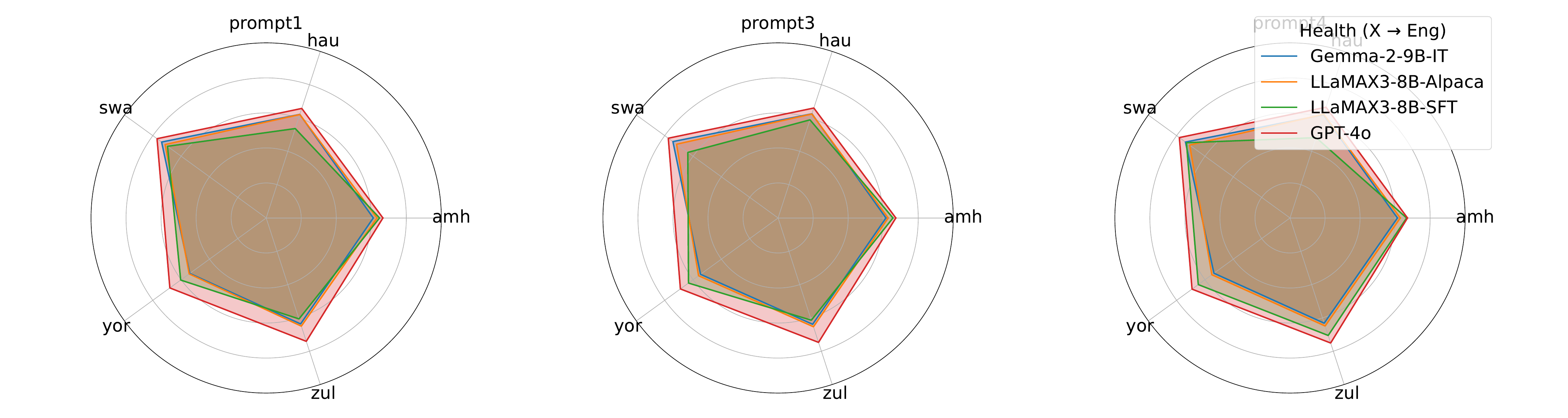}
    \caption{d-chrF scores for some LLMs for sentence-level translation using different prompts when translating into English}
    \label{fig:radar_charts1}
\end{figure*}

\begin{figure*}[t]
    \centering
    \includegraphics[width=\textwidth]{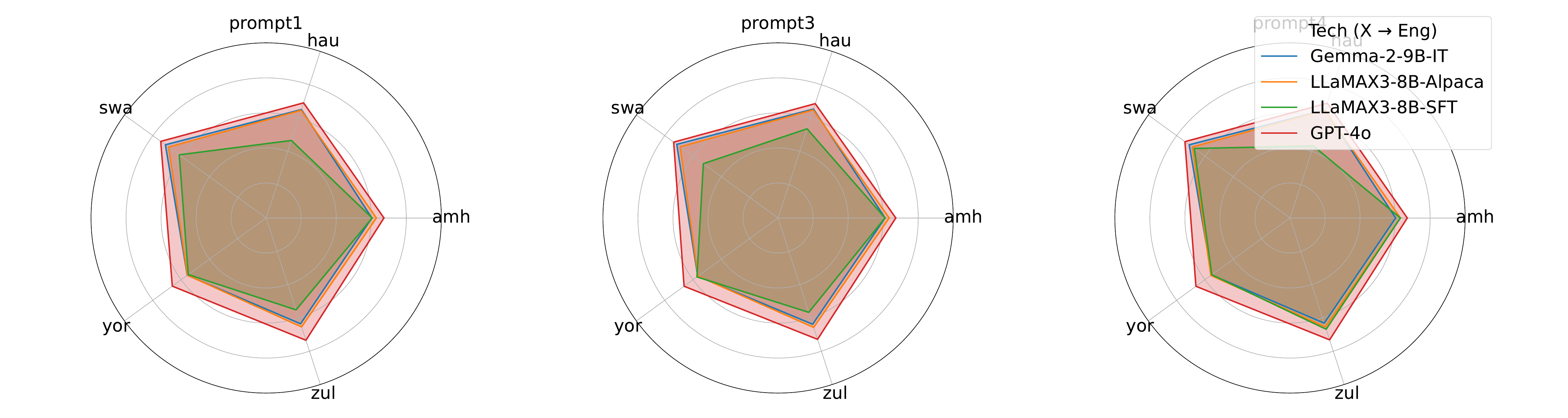}
    \caption{d-chrF scores for some LLMs for sentence-level translation using different prompts when translating into English}
    \label{fig:radar_charts2}
\end{figure*}

\end{document}